\documentclass[fleqn,10pt]{wlscirep}
\usepackage[utf8]{inputenc}
\usepackage[T1]{fontenc}
\usepackage{tikz}
\usetikzlibrary{cd,arrows,calc,shapes,positioning}
\tikzstyle{obs} = [circle,fill=white,draw=black,inner sep=1pt,minimum size=20pt,font=\fontsize{10}{10}\selectfont,node distance=1,thick]
\tikzstyle{latent} = [obs,dotted]
\usepackage{microtype}
\usepackage{booktabs}
\usepackage{bm}
\usepackage{graphicx}
\usepackage{nicefrac}
\usepackage{multirow}
\usepackage[super]{nth}
\usepackage{amsmath}
\usepackage{amssymb}
\usepackage{mathtools}
\usepackage{amsthm}
\usepackage{xcolor}
\usepackage{chngcntr}
\usepackage{appendix}
\usepackage[numbers,square]{natbib}
\usepackage[noabbrev,capitalize]{cleveref}

\newcommand{\edge}[3][]{ %
  \foreach \x in {#2} { %
    \foreach \y in {#3} { %
      \path (\x) edge [->, >={triangle 45}, #1,thick] (\y);%
    };
  };
}

\newcommand{\bR}{\ensuremath \mathbb{R}}

\newcommand{\bS}{\ensuremath \mathbb{S}}

\newcommand{\cL}{\ensuremath \mathcal{L}}

\newcommand{\cN}{\ensuremath \mathcal{N}}
\newcommand{\cD}{\ensuremath \mathcal{D}}

\DeclareMathOperator{\E}{\mathbb{E}}

\DeclareMathOperator{\indep}{\perp\!\!\!\perp}
\DeclareMathOperator{\dep}{\not \perp\!\!\!\perp}
\DeclareMathOperator{\alr}{\mathrm{alr}}
\DeclareMathOperator{\clr}{\mathrm{clr}}
\DeclareMathOperator{\ilr}{\mathrm{ilr}}
\newcommand{\B}[1]{\bm{#1}}
\newcommand{\simp}{\ensuremath \mathbb{S}}
\newcommand{\betaiv}{\ensuremath \hat{\beta}_{\mathrm{iv}}}

\newcommand{\X}{\ensuremath \B{X}}
\newcommand{\y}{\ensuremath \B{y}}
\newcommand{\Z}{\ensuremath \B{Z}}

\newcommand{\err}[2]{$#1$ {\small \color{gray} $\pm #2$}}

\newcommand{\na}{\textcolor{gray}{---}}
\newcommand{\thead}[1]{\multicolumn{1}{c}{\textbf{#1}}}
\newcommand{\ilrilr}{2SLS\textsubscript{ILR}}
\newcommand{\kivilr}{KIV\textsubscript{ILR}}
\newcommand{\ilrlc}{ILR+LC}

\graphicspath{{Fig/}}

\crefname{section}{Supplementary Material}{Supplementary Materials}
\Crefname{section}{Supplementary Material}{Supplementary Materials}

\title{Instrumental variable estimation\\ for compositional treatments}

\author[1,2,3,$\ast$]{Elisabeth Ailer}
\author[1,3,4,5]{Christian~L.~Müller}
\author[1,2,3]{Niki Kilbertus}

\affil[1]{Helmholtz Munich, Ingolstädter Landstraße 1, Neuherberg, 85764, Germany}
\affil[2]{TUM School of Computation, Information and Technology, Technical University of Munich, Boltzmannstraße 3, Garching, 85748, Germany}
\affil[3]{Munich Center for Machine Learning (MCML)}
\affil[4]{Department of Statistics, Ludwig-Maximilian University, Geschwister-Scholl-Platz 1, Munich, 80539, Germany}
\affil[5]{Center for Computational Mathematics, Flatiron Institute, 162 5th Ave, NY 10010, New York, United States}
\affil[$\ast$]{Corresponding author: \href{email:email-id.com}{elisabeth.ailer@helmholtz-munich.de}}


\begin{abstract}
Many scientific datasets are compositional in nature. 
Important biological examples include species abundances in ecology, cell-type 
compositions derived from single-cell sequencing data, and amplicon abundance data
in microbiome research. Here, we provide a causal view on compositional data in an
instrumental variable setting where the composition acts as the cause. First, we
crisply articulate potential pitfalls for practitioners regarding the 
interpretation of compositional causes from the viewpoint of interventions and 
warn against attributing causal meaning to common summary statistics such as diversity indices in microbiome data analysis.
We then advocate for and develop multivariate methods using statistical data transformations and regression techniques that take the special structure of the compositional sample space into account while still yielding scientifically interpretable results.
In a comparative analysis on synthetic and real microbiome data we show the advantages and limitations of our proposal. 
We posit that our analysis provides a useful framework and guidance for valid and informative cause-effect estimation in the context of compositional data.
\end{abstract}
\begin{document}

\flushbottom
\maketitle
\thispagestyle{empty}

\section*{Introduction and Motivation}
The statistical modeling of compositional (or relative abundance) data plays a pivotal role in many areas of science, ranging from the analysis of mineral samples or rock compositions in earth sciences \citep*{aitchison1982statistical} to correlated topic modeling in large text corpora \citep{Blei2005,Blei2007}.
Recent advances in biological high-throughput sequencing techniques, including single-cell RNA-Seq and microbial amplicon sequencing \citep{rozenblatt2017human,turnbaugh2007human}, have triggered renewed interest in compositional data analysis.
Since only a limited total number of transcripts can be captured in a sample by current sequencing technologies, the resulting count data provides relative abundance information about mRNA transcripts or microbial amplicon sequences, respectively \citep{Quinn2018a,gloor2017microbiome}. 

For example, in microbiome sequencing, this stems from the fact that one cannot easily control for the total number of microbes entering the measurement process.
Bacterial microbiome measurements typically come in the form of counts of operational taxonomic units (OTUs)
or amplicon sequencing variants (ASVs) derived from high-throughput sequencing of 16S ribosomal
RNA (rRNA) \citep{johnson2019evaluation} and are summarized as taxonomic compositions, e.g., on the species, genus, or family level. 

One way of dealing with the available relative abundance information is to normalize read counts by their respective totals, resulting in \emph{compositional data}.
Compositional data comprises the proportions of some whole, implying that data points live on the unit simplex $\simp^{p-1} := \{x \in \bR^p_{\ge 0} \mid
\sum_{j=1}^p x_j = 1\}$.

In the microbiome example, assume there are~$p$ different microbial taxa that have been identified in a human gut microbiome experiment.
A specific gut microbiome measurement is then represented by a vector~$x$, where~$x_j$ denotes the relative abundance of taxon~$j$ (under an arbitrary ordering of taxa).
An increase in~$x_1$ within this composition could correspond to an actual increase in the absolute abundance of the first taxon, while the rest remained constant.
However, it could equally result from a decrease of the absolute abundance of the first species with the remaining ones having decreased even more.

Statisticians have recognized the significance of compositional data early on (dating back to Karl Pearson) and tailored models to naturally account for compositionality via simplex arithmetic \citep{aitchison1982statistical}.
Despite these efforts, adjusting predictive statistical and machine learning methods to compositional data remains an active field of research \citep{rivera2018balances,bates2019log, Cammarota2020,Quinn2020a,Oh2020,buettner2021sccoda,park2022kernel,huang2023supervised,taba2023causalIV,xiang2021causalMR}.

This work focuses on estimating the causal effect of a composition on a categorical or continuous outcome.
Only recently have the fundamental challenges in interpreting causal effects of compositions been acknowledged explicitly \citep{arnold2020causal,breskin2020commentary} with little work on how to estimate such effects from observational data.
Our work provides scalable methods that \emph{enable practitioners to answer the simple question: ``What is the causal effect of a composition on some outcome of interest?''}

\subsection*{Pitfalls with Summary Statistics}\label{sec:pitfalls_summary}

First, let us motivate the \textit{compositional} aspect of the question. 
In microbiome research specifically, species diversity became the center of attention to an extent that asking ``what is the causal effect of \emph{the diversity} of a composition $X$ on the outcome $Y$?'' appears more intuitive than asking for the causal effect of individual abundances.
In fact, popular books and research articles alike seem to suggest that (bio-)diversity is indeed an important \emph{causal driver} of ecosystem functioning and human health, even though these claims are largely grounded in observational, non-experimental data \citep{Chapin2000,Blaser2014}.
Similar summary statistics or low-dimensional representations have been proposed in other domains such as in single-cell RNA data \citep{heumos2023best}.
We now explain why, even in situations where summary statistics appear to be useful proxies, no causal conclusions can be drawn from them.

Let us consider $\alpha$-diversity as an example of a one-dimensional summary statistic of a microbiome measurement, e.g. $\alpha_{\text{Simpson}} = -\sum_{j = 1}^p (x_j)^2$ or $\alpha_{\text{Shannon}} = -\sum_{j=1}^p x_j \log x_j$.
The ``causal effect'' of the diversity $\alpha$ on some outcome of interest $Y$ (e.g., health or disease indicator) is usually considered to be the expected value of $Y$ under an \emph{intervention on the diversity}, i.e., externally setting the diversity to a chosen value $\alpha^*$, with all host and environmental factors unchanged.
This causal effect is commonly denoted by $\E[Y \mid do(\alpha=\alpha^*)]$.
When $\E[Y \mid do(\alpha = \alpha_1)] < \E[Y \mid do(\alpha = \alpha_2)]$ for two diversity values $\alpha_1, \alpha_2$ with $\alpha_1 < \alpha_2$, one would then be tempted to conclude that ``increasing diversity $\alpha$ causes an increase in the outcome $Y$'', which is often loosely translated to ``diversity is a causal driver for health''.
We now highlight critical issues with this approach.

(a) When considering the proposed causal effect estimand $\E[Y \mid do(\alpha=\alpha^*)]$ directly, one presupposes the existence of clearly defined interventions on $\alpha$.
However, there are infinitely many ways of changing the diversity of a composition by a fixed amount.
This `many-to-one' nature prevents a consistent conceptualization of external interventions.
In particular, for a given value of $\alpha$, there is a $(p-2)$-dimensional subspace of $\simp^{p-1}$ with that value of $\alpha$.
Hence, an intervention to ``increase the diversity of a given composition'' by some $\Delta \alpha$ is highly ambiguous.
The different ways of achieving this change must be expected to have different implications for the outcome $Y$.
Similarly, most common diversity measures are invariant under permutations of components and the above approach would require us to conclude that all $p!$ permutations of a composition are functionally completely equivalent with regard to the outcome $Y$---an abstruse claim.
\emph{Hence, assigning causal powers to diversity by estimating $\E[Y \mid do(\alpha)]$ is highly ambiguous and does not carry the intended meaning.}
This concern is further exacerbated by the difficulty and ambiguity in measuring $\alpha$-diversity in the first place 
\citep{shade2017diversity,gloor2017microbiome, Willis2019}.

\begin{figure}
    \centering
    \includegraphics[width=0.5\linewidth]{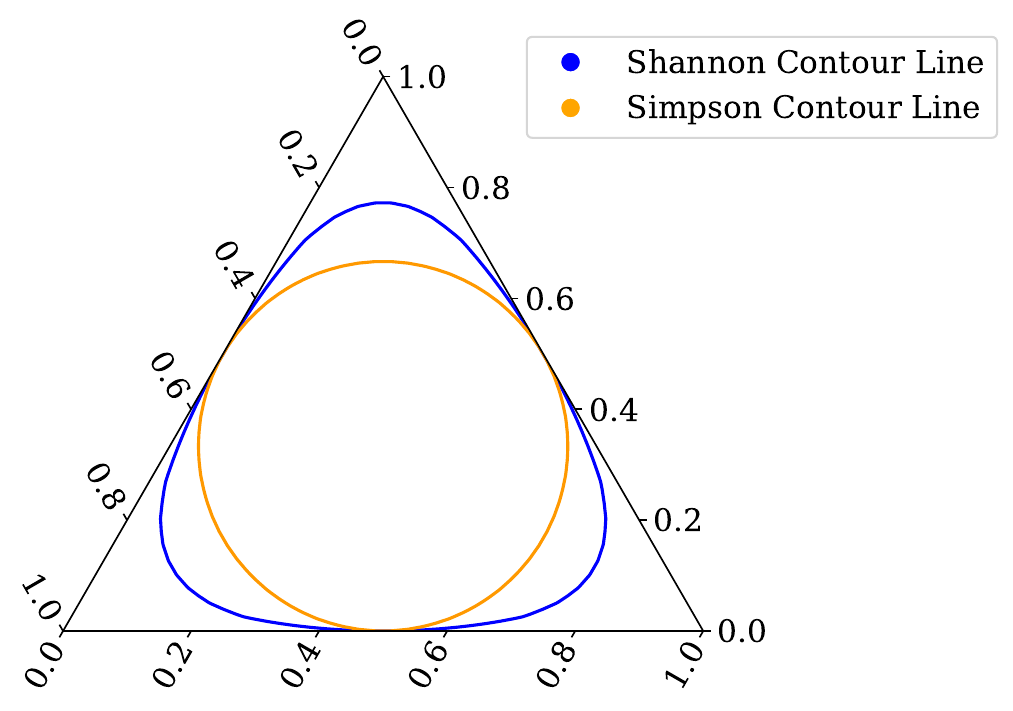}
    \caption{The ternary plot shows an exemplary scenario with $p=3$. The orange contour contains compositions for which the Simpson diversity is constant, while the blue contour shows compositions for which the Shannon diversity is constant. Shannon diversity changes along contours of Simpson diversity and vice versa.}
    \label{fig:shannon_simpson_comparison}
\end{figure}

(b) The definition of $\alpha$-diversity is not unique, which could lead to a potential search for positive results by 
using a different metric \citep{kers2022diversity} or contradictory causal claims. Consider two different one-dimensional 
summary statistics $\alpha_1, \alpha_2$ on $\simp^{p-1}$. These can be defined in terms of their contours, i.e., the 
collection of $(p-2)$-dimensional subspaces of $\simp^{p-1}$ of constant values of $\alpha_1$ and $\alpha_2$ respectively.
Since they are different, there will be a contour line of $\alpha_1$ along which $\alpha_2$ either increases or decreases.
Along this path through compositions, we would have to conclude that the causal effect of one summary statistic is zero, while it is non-zero for the other.
See \cref{fig:shannon_simpson_comparison} for a visualization.
In typical scenarios, there is no ``one correct'' summary statistic, such that reliable claims even about the sign of the causal effect of a summary statistic of a composition become void.

\subsection*{Cause-Effect Estimation with Instrumental Variables}

While researchers continue to develop predictive methods for compositional data \citep{huang2023supervised}, in most scientific contexts causal effects are of greater interest.
For example, the human microbiome co-evolves with its host and the external environment through diet, activity, climate, or geography, etc. leading to plentiful microbiome-host-environment interactions \citep{Vujkovic-Cvijin2020}.
Carefully designed studies may allow us to control for certain environmental factors and specifics of the host.
In fact, several recent works studied the causal mediation effect of the microbiome on health-related outcomes, assuming all relevant covariates are observed and can be controlled for \citep{Sohn2019,carter2020information,Wang2020,xia2021mediation,sohn2022compositional,wang2023mediation,zhang2020mediation}, or vice versa, the effect of environmental factors on the microbiome \citep{sommer2022randomization}.
However, in practice there is little hope of measuring \emph{all} latent factors in these complex interactions.
In such a situation, a purely predictive model will suffer from bias due to the unobserved confounders.
Such unobserved confounders are a major hurdle in cause-effect estimation broadly and also specifically for compositional causes.

Concretely, without further assumptions, the direct causal effect $X \to Y$ is not identified from observational data in the presence of unobserved confounding $X \leftarrow U \to Y$ \citep{pearl2009causality}.
One common way to still identify the causal effect from purely observational data is through so-called instrumental variables (IV) \citep{angrist2008mostly}.
An \emph{instrumental variable} $Z$ is a variable that has an effect on the cause $X$ ($Z \to X$), but is independent of the confounder ($Z \indep U$), and conditionally independent of the outcome given the cause and the confounder ($Z \indep Y \mid \{U, X\}$). 
In practice, it can be hard to find valid instruments for a target effect \citep{hernan2006instruments}, but when they do exist, instrumental variables often render efficient cause-effect estimation possible.

In this work, we develop interpretable methods to estimate the direct causal effect of a \emph{compositional cause}~$X$ on a continuous or categorical outcome~$Y$ within the IV setting.
The question of whether and how cause-effect estimation for compositional treatments under unobserved confounding is possible remains unanswered in the literature, motivating our in-depth analysis of two-stage methods for interpretable cause-effect estimation of individual relative abundances on the outcome.
In the analysis, we focus on a careful selection and combination of existing approaches and a thorough examination of potential pitfalls and mis-usage.
Our extensive empirical evaluations carefully assess assumptions (additive noise, strong instruments) and model misspecification as a potential obstacle to interpretable and reliable effect estimates.
We evaluate the efficacy and robustness of our proposed methods on both synthetic and real data from a mouse experiment, examining how the gut microbiome ($X$) affects body weight ($Y$) instrumented by sub-therapeutic antibiotic treatment (STAT) ($Z$).

The rest of the manuscript proceeds as follows.
First, we introduce the concepts of compositional data and instrumental variables in detail. Following this introduction of our methods, we provide some simulation to study the advantage and potential pitfalls in using high-dimensional compositional data in instrumental variable settings.
Last but not least, we then apply the methods to a real world dataset.


\section*{Methods}
\label{sec:methods}

\begin{figure}[!tpb]
  \centering
  \begin{tikzpicture}
    \node[obs,label={[align=center]below:{$\bR^q$ \\[2mm]{\footnotesize \color{black!50}STAT}}}] (Z) at (0, 0) {$Z$};
    \node[obs,draw=cyan,label={[align=center]below:{$\color{cyan}\simp^{p-1}$ \\[2mm]{\footnotesize \color{cyan} compositional vector}}}] (X) at (2.2, 0) {{\color{cyan}$X$}};
    \node[obs,label={[align=center]below:{\quad $\bR$ or $\{0,1\}$ \\[2mm]{\footnotesize \color{black!50}outcome}}}] (Y) at (4.6, 0) {$Y$};
    \node[latent,label={{\small \color{black!50}unobserved confounder}}] (U) at (3.4, 1.3) {$U$};
    \node (sp) at (3.4,0.2) {\small {\color{orange} sparse}};
    \node (eff) at (3.4,-0.2) {\small {\color{orange} effect}};
    \edge{Z}{X}
    \edge[orange]{X}{Y}
    \edge{U}{X,Y}
  \end{tikzpicture}%
  \caption{Cause-effect estimation of $X \to Y$ via an instrumental variable $Z$ for compositional $X$.}
  \label{fig:setup}
\end{figure}

\subsection*{Instrumental Variables}

We briefly recap the assumptions of the instrumental variable setting as depicted in \cref{fig:setup}.
For an \emph{outcome} (or \emph{effect}) $Y$, a \emph{treatment} (or \emph{cause}) $X$, and potential \emph{unobserved confounders} $U$, we assume access to a discrete or continuous \emph{instrument} $Z \in \bR^q$ satisfying
(i) $Z \indep U$ (the confounder is independent of the instrument),
(ii) $Z \dep X$ (``the instrument influences the cause''), and
(iii) $Z \indep Y \mid \{X, U\}$ (``the instrument influences the outcome only through the cause'').
Our goal is to estimate the direct causal effect of $X$ on $Y$, written as $\E[Y | do(x)]$ in the do-calculus notation \citep{pearl2009causality} or as $\E[Y(x)]$ in the potential outcome framework \citep{imbens2015causal}, where $Y(x)$ denotes the potential outcome for treatment value $x$.
The functional dependencies are $X = g(Z,U)$, $Y = f(X,U)$.
While $Z, X, Y$ denote random variables, we also consider a dataset of $n$ i.i.d.\ samples $\cD = \{(z_i, x_i, y_i)\}_{i=1}^n$ from their joint distribution.
We arrange observations in matrices or vectors denoted by $\X \in \bR^{n\times p}$ or $\X \in (\bS^{p-1})^n$, $\Z \in \bR^{n \times q}$, $\y \in \bR^n$.

Without further restrictions on $f$ and $g$, the causal effect is not identified \citep{pearl1995testability,bonet2001instrumentality,gunsilius2018testability}.
The most common assumption leading to identification is that of \emph{additive noise}, namely $Y = f(X) + U$ with $\E[U] = 0$ but not necessarily $X \indep U$.
Here, we overload the symbols $f$ and $g$ for simplicity.
The implied Fredholm integral equation of first kind $\E[Y \mid Z] = \int f(x)\, \mathrm{d}P(X\mid Z)$ is generally ill-posed.
While the linear case is well understood \citep{angrist2008mostly}, under certain regularity conditions the IV problem can be solved consistently even for non-linear $f$, see e.g., \citep{newey2003instrumental,blundell2007semi} and more recently \citep{singh2019kernel,muandet2019dual,zhang2020maximum,bennett2023minimax}.
However, in this case the problem is typically under-specified in that multiple $f$ are compatible with the observed data and regularization techniques are typically used to obtain a unique solution---typically the smallest compatible $f$ according to some norm.
It is thus difficult to interpret estimates of non-linear causal-effects in a way that aids understanding of the underlying processes.

In the simplest case, where $X \in \bR^p$ and $f, g$ are linear, a standard \emph{instrumental variable estimator} is
\begin{equation}\label{eq:betaiv}
  \betaiv = (\X^T \B{P}_{Z} \X)^{-1} \X^T \B{P}_{Z}\, \y 
\end{equation}
with $\B{P}_{Z} = \Z (\Z^T \Z)^{-1} \Z^T$ \citep{angrist2008mostly}.
For the \emph{just-identified} case $q=p$ as well as the over-identified case $q > p$, this estimator is consistent and asymptotically unbiased, albeit not unbiased.
In the \emph{under-identified} case $q < p$, where there are fewer instruments than treatments, the orthogonality of $Z$ and $U$ does not imply a unique solution. Again, regularization or other objectives such as sparsity assumptions have been proposed to obtain unique a unique solution within the space of compatible $\beta$ \citep{rothenhausler2021anchor,pfister2022identifiability,ailer2023sequential}.
The estimator $\betaiv$ can also be interpreted as the outcome of a \emph{two-stage least squares} (2SLS) procedure consisting of
(1) regressing $\X$ on $\Z$ via OLS $\hat{\delta} = (\Z^T \Z)^{-1} \Z^T \X$, and
(2) regressing $\y$ on the predicted values $\hat{\X} = \Z \hat{\delta}$ via OLS, again resulting in $\betaiv$.
Practitioners are typically discouraged from using the manual two-stage approach, because the OLS standard errors of the second stage are wrong---a correction is needed \citep{angrist2008mostly}.
However, we note that the point estimator obtained by the manual two-stage procedure is equivalent to \cref{eq:betaiv}.

Moreover, the two-stage description suggests that the two-stages are independent problems and thereby seems to invite us to mix and match different regression methods as we see fit.
The authors in \cite{angrist2008mostly} highlight that the asymptotic properties of $\betaiv$ rely on the fact that for OLS the residuals of the first stage are uncorrelated with the instruments $\Z$.
Hence, for OLS we achieve consistency of $\betaiv$ \emph{even when the first stage is misspecified}.
For a non-linear first stage regression we may only hope to achieve uncorrelated residuals asymptotically when the model is correctly specified.
Replacing the OLS first stage with a non-linear model is known as the ``forbidden regression'', a term commonly attributed to Prof.~Jerry Hausmann.
Angrist and Pischke acknowledge that the practical relevance of the forbidden regression is not well understood.
When also the second stage is assumed to be non-linear, one would require independence of the first stage residuals from $Z$.
Starting with 
\cite{kelejian1971two} there is now a rich literature on the circumstances under which ``manual 2SLS'' with non-linear first (and/or second) stage can yield consistent causal estimators.
Primarily interested in \emph{compositional} treatments $X$, we cannot directly use OLS for either stage.
Since there is no theoretical guidance for this case, we assess our options empirically, paying great attention to potential issues due to the ``forbidden regression'' and misspecification in our proposed methods.

\subsection*{Compositional Data}
\label{sec:ivcomp}

\subsubsection*{Simplex geometry}
The authors in 
\cite{aitchison1982statistical}  introduced the \emph{perturbation} and \emph{power transformation} as the simplex $\simp^{p-1}$ counterparts to addition and scalar multiplication of Euclidean vectors in $\bR^p$:

\medskip
\noindent\begin{minipage}{0.45\linewidth}
\centering
Perturbation
\begin{align*}
    \qquad \qquad \oplus &: \simp^{p-1} \times \simp^{p-1} \to \simp^{p-1} \\
    x \oplus &w = C(x_1 w_1, \ldots, x_p w_p)
\end{align*}
\end{minipage}%
\hfill\vline\hfill
\begin{minipage}{0.45\linewidth}
\centering
Power transformation
\begin{align*}
    \qquad \qquad \odot &: \bR \times \simp^{p-1} \to \simp^{p-1} \\
    a \odot &x := C(x_1^a, x_2^a, \ldots, x_p^a)
\end{align*}
\end{minipage}
\bigskip

Here, the \emph{closure operator} $C: \bR^p_{\ge 0} \to \simp^{p-1}$ normalizes a $p$-dimensional, non-negative vector to the simplex $C(x) := x / \sum_{i=1}^p x_i$.
Together with the dot-product 
\begin{equation}
  \langle x, w \rangle := \frac{1}{2 p} \sum_{i,j=1}^p \log \Bigl(\frac{x_i}{x_j}\Bigr) \log \Bigl(\frac{w_i}{w_j}\Bigr)
\end{equation}
the tuple $(\simp^{p-1}, \oplus, \odot, \langle \cdot, \cdot \rangle)$ forms a finite-dimensional real Hilbert space \citep{pawlowsky2001geometric} allowing to transfer usual geometric notions such as lines and circles from Euclidean space to the simplex.

\subsubsection*{Coordinate representations}
The $p$ entries of a composition remain dependent via the unit sum constraint, leading to $\simp^{p-1}$ having dimension $p-1$.
To deal with this fact, different invertible log-based transformations have been proposed, for example the additive log ratio, centered log ratio \citep{aitchison1982statistical}, and isometric log ratio \citep{egozcue2003isometric} transformations
\begin{equation}
 \alr(x)= V_{\mathrm{a}} \log(x) \in \bR^{p-1},\quad
 \clr(x)= V_{\mathrm{c}} \log(x) \in \bR^{p},\quad
 \ilr(x)= V_{\mathrm{i}} \log(x) \in \bR^{p-1},
\end{equation}
where the logarithm is applied element-wise and the matrices $V_{\mathrm{a}}, V_{\mathrm{i}} \in \bR^{(p-1)\times p}$ and $V_{\mathrm{c}} \in \bR^{p\times p}$ are defined in \cref{supp:compositional_transformations}.
While $\alr$ is a vector space isomorphism that preserves a one-to-one correspondence between all components except for one, which is chosen as a fixed reference point to reduce the dimensionality (we choose $x_p$, but any other component works), it is not an isometry, i.e., it does not preserve distances or scalar products.
Both $\clr$ and $\ilr$ are also isometries, but $\clr$ only maps onto a subspace of $\bR^p$, which often renders measure theoretic objects such as distributions degenerate.
As an isometry between $\simp^{p-1}$ and $\bR^{p-1}$, $\ilr$ allows for an orthonormal coordinate representation of compositions.
However, it is hard to assign meaning to the individual components of $\ilr(x)$, which all entangle a different subset of relative abundances in $x$ leading to challenges for interpretability \citep{greenacre2019isometric}.
Therefore, $\alr$ remains a useful tool in statistical analyses where interpretability is required despite the lack of the isometric property.

\subsubsection*{Log-contrast estimation}
The key advantage of such coordinate transformations is that they allow us to use regular multivariate data analysis methods (typically tailored to Euclidean space) for compositional data.
For example, we can directly fit a linear model $y = \beta_0 + \beta^T \ilr(x) + \epsilon$ on the $\ilr$ coordinates via ordinary least squares (OLS) regression.
However, in real-world datasets,~$p$ is often a large number capturing ``all possible components in a measurement'', leading to $p \gg n$ with each of the $n$ measurements being sparse, i.e., a substantial fraction of~$x$ being zero.
Moreover, in many (especially high-dimensional) situations only few components exert direct causal influence on the outcome.
Both overparameterization $p\gg n$ as well as assuming sparse effects call for regularization.
The problem with enforcing sparsity in a ``linear-in-$\ilr{}$'' model is that a zero entry in~$\beta$ does not correspond directly to a zero effect of the relative abundance of any single component.
This motivates \emph{log-contrast} estimation \citep{Aitchison1984} with a sparsity penalty \citep{Lin2014c,Shi2016, Combettes2021}
\begin{equation}\label{eq:logcontrast}
  \min_{\beta} \sum_{i=1}^n \cL(x_i, y_i, \beta) + \lambda \| \beta \|_1 \quad \text{s.t.}\: \sum_{i=1}^p \beta_i = 0 \:.
\end{equation}
In our examples, we focus mostly on continuous $y \in \bR$ and the squared loss $\cL(x, y, \beta) = (y - \beta^T \log(x))^2$.
However, our framework  
also supports the Huber loss for robust log-contrast regression as well as an optional joint concomitant scale estimation for both losses \citep{Combettes2020a,Combettes2021}.
Moreover, for classification tasks with $y \in \{0, 1\}$, we can directly use the squared Hinge loss (or a ``Huberized'' version thereof) for $\cL$, see \cref{supp:method_training} for details.
These flexible estimation formulations respect the compositional nature of $x$ while retaining the association between the entry $\beta_i$ and the relative abundance of the individual component $x_i$.
Even though, due to the additional sum constraint, individual components of $\beta$ are still not---and can never be---entirely disentangled.

\subsubsection*{Logs and zeros}
In the previous paragraphs, we introduced multiple log-based coordinate representations for compositions and at the same time claimed that measurements are often sparse in relevant settings.
Since the logarithm is undefined for zero entries, a simple strategy is to add a small constant to all absolute counts, so called \emph{pseudo-counts} \citep{kaul2017analysis,lin2020analysis}. These pseudo-counts are particularly popular in the microbiome and single-cell RNA literature where there are many more possible taxa/genes (up to tens of thousands) that occur in any given sample. Despite the simplicity of adding a constant pseudo-count, for example $0.5$, recent work gives theoretical and empirical evidence for this approach \citep{shi2022high}, which we also use here.

\subsubsection*{Summary statistics}
Traditionally, interpretability issues around compositions have been circumvented by focusing on summary statistics instead of individual relative abundances.
One of the key measures to describe ecological populations is \emph{diversity}.
Diversity captures the variation within a composition and is in this context often called $\alpha$-diversity.
There is no unique definition of $\alpha$-diversity.
Among the most common ones in the literature are \emph{richness}, i.e. the number of non-zero entries denoted as $\|x\|_{0}$, \emph{Shannon diversity} $-\sum_{j=1}^p x_j \log(x_j)$ and \emph{Simpson diversity} $-\sum_{j=1}^p x_j^2$.
Especially in the microbial context, there exist entire families of diversity measures taking into account species, functional, or phylogenetic similarities between taxa and tracing out continuous parametric profiles for varying sensitivity to highly-abundant taxa.
See for example \citep{Cobbold2012,chao2014unifying,daly2018ecological} for an overview of the possibilities and choices of estimating $\alpha$-diversity in a specific application.
While the popularity of $\alpha$-diversity for assessing the impact and health of microbial compositions \citep{Bello2018} seemingly renders it a natural choice for causal queries, we argue that such claims are misleading and void of a solid foundation. 

\subsection*{Methods for Higher Dimensional Causes}\label{sec:composition}

In this section we develop methods to reason about the effects of hypothetical interventions on the relative abundance of individual components from observational data.

\begin{itemize}
  \item \emph{2SLS:} As the first baseline, we run 2SLS from \cref{eq:betaiv} directly on $X \in \simp^{p-1}$ ignoring its compositional nature.
  \item \emph{Only LC}
    For completeness, as the second baseline, we run log-contrast (LC) estimation for the second stage only, thereby entirely ignoring confounding.
  \item \emph{\ilrilr{}:}
    2SLS with $\ilr(X) \in \bR^{p-1}$ as the treatment;
    since OLS minima do not depend on the chosen basis, parameter estimates for different log-transformations of $X$ are related via fixed linear transformations.
    Hence, as long as no sparsity penalty is added, $\ilr$ and $\alr$ regression yield equivalent results.
    The isometric $\ilr$ coordinates are useful due to the consistency guarantees of 2SLS given that $\Z^T \ilr(\X)$ has full rank.
    For interpretability, $\alr$ coordinates can be beneficial as individual coordinates correspond to individual components (given a reference).
    The respective coordinate transformations are given in \cref{supp:compositional_transformations}.
  \item \emph{\kivilr{}:}
    Following 
    \cite{singh2019kernel} 
    we replace OLS in \ilrilr{} with kernel ridge regression in both stages to allow for non-linearities.
    Like \ilrilr{}, \kivilr{} cannot enforce sparsity in an interpretable fashion.
  \item \emph{\ilrlc{}:}
    To account for sparsity, we use sparse log-contrast estimation (see \cref{eq:logcontrast}) for the second stage, while retaining OLS to $\ilr$ coordinates for the first stage.
    Log-contrast estimation conserves interpretability in that the estimated parameters correspond directly to the effects of individual relative abundances.
  \item \emph{DIR+LC:}
    Finally, we circumvent log-transformations entirely and deploy regression methods that naturally work on compositional data in both stages.
    For the first stage, we use a Dirichlet distribution---a common choice for modeling compositional data---where $X\mid Z \sim \mathrm{Dirichlet}(\alpha_1(Z), \ldots, \alpha_p(Z))$ with density $B(\alpha_1, \ldots, \alpha_p)^{-1} \prod_{j=1}^{p}x_j^{\alpha_j - 1}$ where we drop the dependence of $\alpha = (\alpha_1,\ldots ,\alpha_p) \in \bR^p$ on $Z$ for simplicity.
    With the mean of the Dirichlet distribution given by $\nicefrac{\alpha}{\sum_{j=1}^p \alpha_j}$, we account for the $Z$-dependence via $\log(\alpha_j(Z_i)) = \omega_{0,j} + \omega_j^T Z_j$.
    We then estimate the newly introduced parameters $\omega_{0,j} \in \bR$ and $\omega_j \in \bR^q$ via maximum likelihood estimation with $\ell_1$ regularization.
    For the second stage we again resort to sparse log-contrast estimation.
    If the non-linear first stage is misspecified, the ``forbidden regression'' bias may distort effect estimates of this approach.
    This is contrasted by Dirichlet regression potentially resulting in a better fit of the data than linearly modeling log-transformations.
\end{itemize}

We highlight that only \ilrlc{} and DIR+LC accommodate all relevant requirements: (i) unobserved confounding, (ii) compositional treatments, (iii) sparse effects, and (iv) interpretable estimates.

\section*{Simulation Studies} 
\label{sec:simulation}

\subsection*{Data Generation}
\label{sec:datageneration}

\begin{figure}
  \centering
  \newlength{\myheight}
  \setlength{\myheight}{2.8cm} 

  \fboxsep=1mm 
  \fboxrule=0.0mm 
  \fcolorbox{gray}{white}{
    \begin{minipage}[t][\myheight][c]{.55\textwidth}
      \centering
      \textbf{Instrument Strength}
      \vfill
      \includegraphics[width=.4\textwidth]{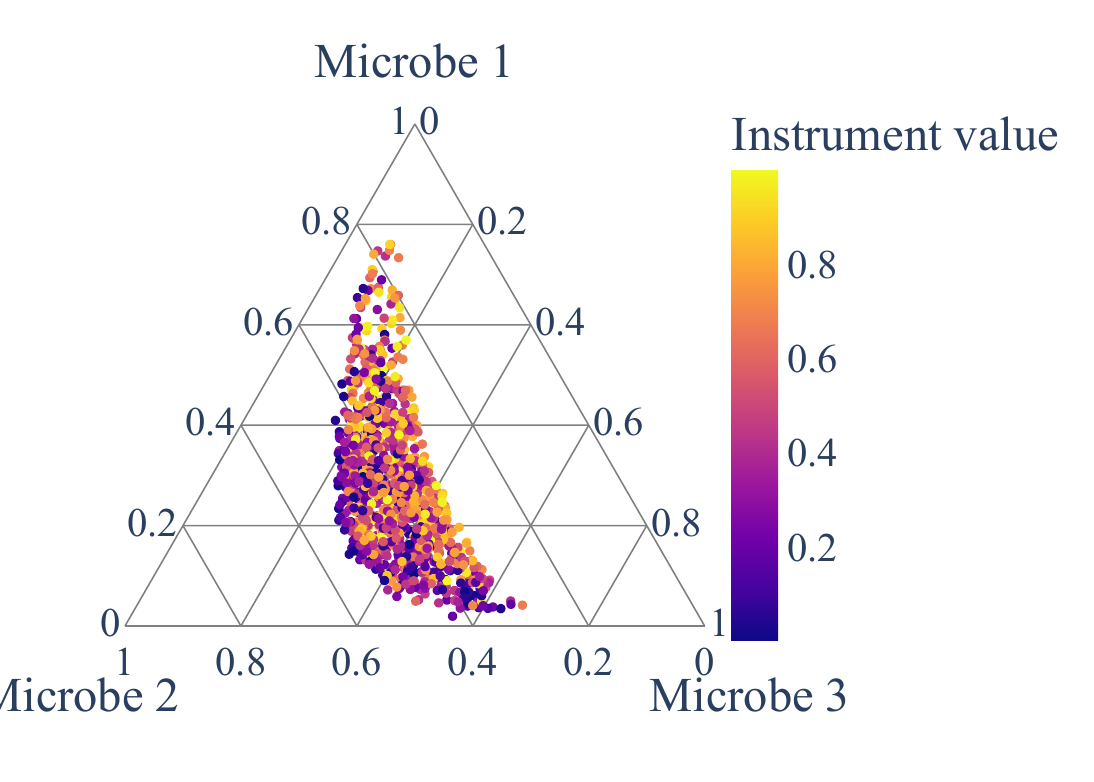}
      \includegraphics[width=.4\textwidth]{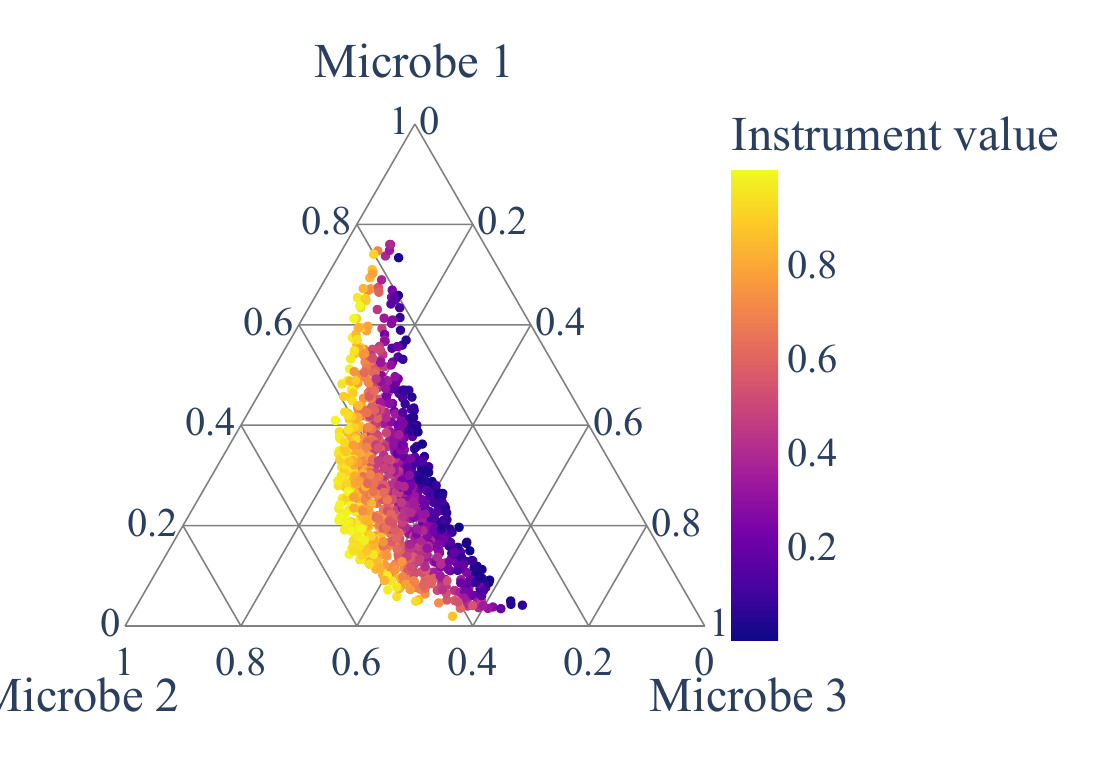}
    \end{minipage}
  }
  \fcolorbox{gray}{white}{
    \begin{minipage}[t][\myheight][c]{.35\textwidth}
      \centering
      \textbf{Confounding Strength}
      \vfill
      \includegraphics[trim={25cm 0 0 0},clip,width=0.9\textwidth]{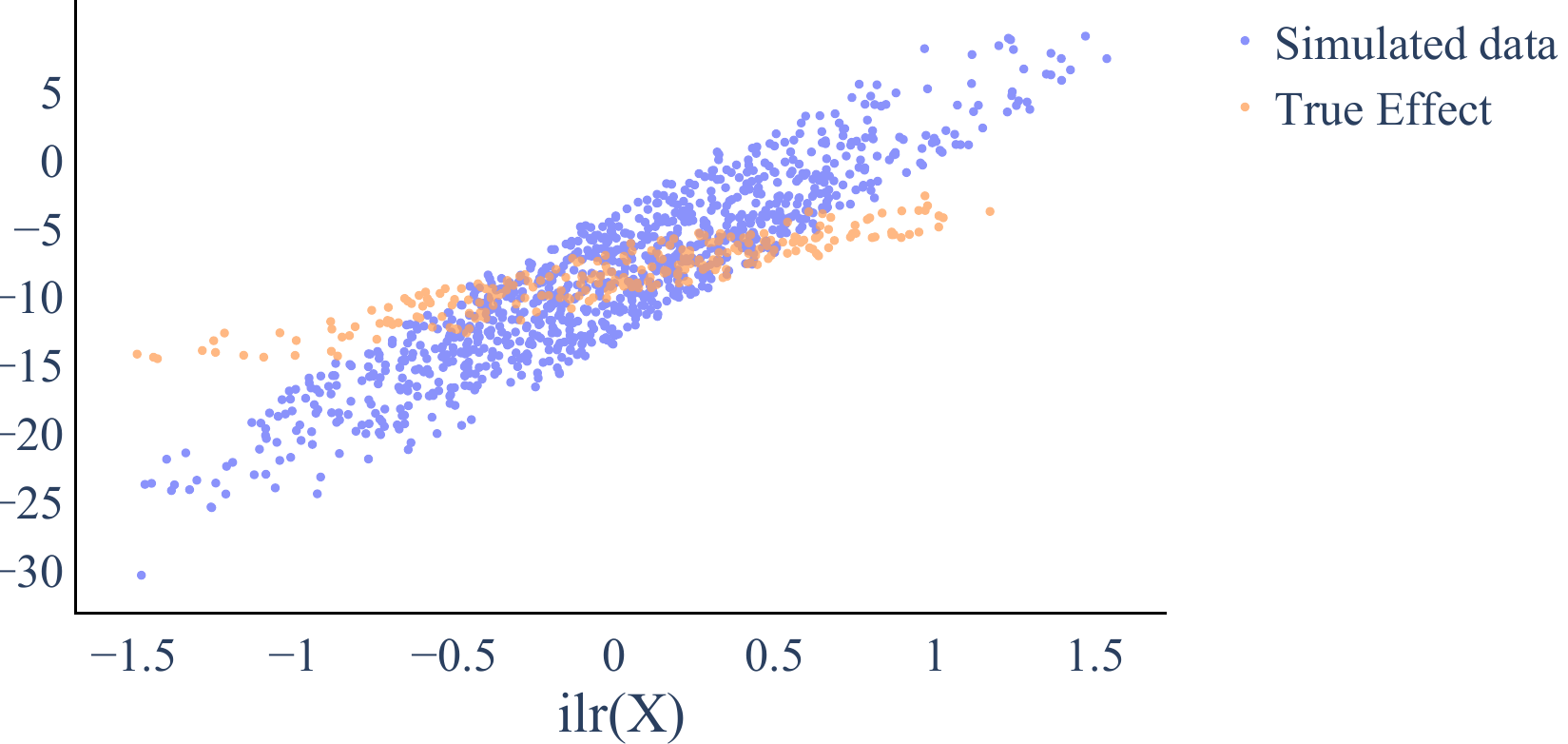}
    \end{minipage}
  }

  \caption{Visualization of a Setting A ($p=3, q=2$): The left panel shows both a weak (left) and a strong (right) instrument. The right panel shows a discrepancy between the true causal effect and the observed effect which stems from a confounding factor.}
  \label{fig:datagen_example}
\end{figure}

For the evaluation of our methods we require the ground truth causal effect to be known.
Since unobserved confounders (and thus counterfactuals) are never observed in practice (by definition), this can only be achieved via synthetic data. We simulate data (in two different settings) to maintain control over ground truth effects, confounding strength, potential misspecification, and the strength of instruments (see Fig.~\ref{fig:datagen_example}). 

\begin{itemize}
  \item \emph{Setting A:}
    The first setting is
    \begin{gather}\label{eq:syntheticdata}
        Z_j \sim \mathrm{Unif}(0, 1),\qquad U \sim \cN(\mu_c, 1), \\
        \ilr(X) = \alpha_0 + \alpha^T Z + U c_X,\quad Y = \beta_0 + \beta^T \ilr(X)  + U c_Y,\nonumber
    \end{gather}
    where we model $\ilr(X) \in \bR^{p-1}$ directly and $\mu_c, c_Y \in \bR$, $\alpha_0, c_X \in \bR^{p-1}$, $\alpha \in \bR^{q \times (p-1)}$ are fixed up front.
    Our goal is to estimate the causal parameters $\beta \in \bR^{p-1}$ and the intercept $\beta_0 \in \bR$.
    This setting satisfies the standard 2SLS assumptions (linear, additive noise) and all our linear methods are thus \emph{wellspecified}.
    To explore effects of \emph{misspecification}, we also consider the same setting only replacing (using $\B{1} = (1, \ldots, 1)$)
    \begin{equation}\label{eq:syntheticdatamis}
    Y = \beta_0 + \frac{1}{100} \B{1}^T(\ilr(X) + 1)^2 + 10 \cdot \B{1}^T \sin(\ilr(X)) + c_Y U.
    \end{equation}
  \item \emph{Setting B:}
    We consider a sparse effect model for $X \in \simp^{p-1}$ which is more realistic for higher-dimensional compositions.
    Note that some parameter dimensions are different, i.e., the same symbols have different meanings in the settings A and B.
    With $\mu = \alpha_0 + \alpha^T Z$ for fixed $\alpha_0 \in \bR^p$, $\alpha \in \bR^{q \times p}$ we use
    \begin{gather}
        Z_j \sim \mathrm{Unif}({Z_{\min}}, {Z_{\max}} ),\qquad U \sim \mathrm{Unif}({U_{\min}}, {U_{\max}}), \nonumber \\
        X  \sim C \bigl(\mathrm{ZINB}(\mu, \Sigma, \theta, \eta ) \bigr) \oplus (U \odot \Omega_C), \label{eq:semisyntheticdata} \\
        Y = \beta_0 + \beta^T \log(X) + c_Y^T \log(U \odot \Omega_C). \nonumber
    \end{gather}
    The treatment $X$ is assumed to follow a zero-inflated negative binomial (ZINB) distribution \citep{greene1994accounting}, commonly used for modelling count data with excess zeros \citep{xu2015assessment}.
    Here, $\eta \in (0, 1)$ is the probability of zero entries, $\Sigma \in \bR^{p\times p}$ is the covariance matrix, and $\theta \in \bR$ the shape parameter.
    The confounder $U \in [U_{\min}, U_{\max}]$ perturbs this base composition in the direction of another fixed composition $\Omega_C \in \simp^{p-1}$ scaled by $U$.
    In simplex geometry $x_0 \oplus (U \odot x_1)$ corresponds to a line starting at $x_0$ and moving along $x_1$ by a fraction $U$.
    A linear combination of the log-transformed perturbation enters $Y$ additively with weights $c_Y \in \bR^p$ controlling confounding strength.
    All other parameter choices are given in \cref{supp:data_generation}.
    This setting is linear in how $Z$ enters $\mu$ and how $U$ enters $X$ and $Y$ in the simplex geometry.
    All our two-stage models are (intentionally) misspecified in the first stage for setting B.
\end{itemize}
The precise choices of all parameters for the different empirical evaluations are described in the appendix (\cref{supp:data_generation}). All relevant code is available at \url{https://github.com/EAiler/causal-compositions}.
    
\subsection*{Metrics and Evaluation}

Appropriate evaluation metrics are key for cause-effect estimation tasks.
We aim at capturing the average causal effect (under interventions) and the causal parameters when warranted by modeling assumptions.
When the true effect is linear in $\log(X)$, we can compare the estimated causal parameters $\hat{\beta}$ from \ilrilr{}, \ilrlc{}, and DIR+LC with the ground truth $\beta$ directly.
In these linear settings, we report causal effects of individual relative abundances $X_j$ on the outcome $Y$ via the mean squared difference (\textbf{$\beta$-MSE}) between the true and estimated parameters $\beta$ and $\hat{\beta}$.
Moreover, we also report the number of falsely predicted non-zero entries (\textbf{FNZ}) and falsely predicted zero entries (\textbf{FZ}), which are most informative in sparse settings and metrics of key interest to biostatisticians.

In the general case, where a measure for identification of the interventional distribution $P(Y \mid do(X))$ is not straightforward to evaluate, we focus on the \emph{out of sample error} (\textbf{OOS MSE}):
For the true causal effect we first draw an i.i.d.\ sample $\{x_i\}_{i=1}^{m}$ from the data generating distribution (that are not in the training set, i.e., out of sample) and compute $\E_{U}[f(x_i, U)]$ for the known $f(X, U)$, the expected $Y$ under intervention $do(x_i)$.
We use $m=250$ for all experiments.
OOS MSE is then the mean square difference to our second-stage predictions $\hat{f}(x_i)$ on these out of sample $x_i$.
Because in real observational data we do not have access to $P(Y \mid do(X))$ (but only the conditional distribution $P(Y \mid X)$, we can not evaluate OOS MSE in real-world observational data.

We run each method for 50 random seeds in setting A (\cref{eq:syntheticdata}), and 20 random seeds in setting B (\cref{eq:semisyntheticdata}).
In result tables, we report mean and standard error over these runs.
The sample size is $n=1000$ in the low-dimensional case ($p=3$) and $n=10,\!000$ in the higher-dimensional cases ($p=30$, $p=250$). 
Additionally, we report results for an overparameterized setting with $n=100$ and $p=250$.
\Cref{supp:data_generation,supp:method_results} contain further explanations and more detailed results.
Note, that since $\alr$ coordinates for $X$ yield equivalent optimization minima as \ilrlc{}, we only report results from \ilrlc{}. All numbers match precisely for ALR+LC in our empirical evaluation.

\begin{table*}
\centering
{
\begin{tabular}{p{0.08\textwidth}crrrr}
\toprule
  & & \multicolumn{4}{c}{\textbf{Setting A, \cref{eq:syntheticdata}}} \\
  \cmidrule(lr){3-6}
  \thead{Dim.} & \thead{Method} & \textbf{OOS MSE} & \thead{$\beta$-MSE} & \thead{FZ} & \thead{FNZ} \\ 
  \midrule
  
  \multirow{5}{*}{\shortstack{$p=3$ \\$q=2$}}
  & DIR+LC    & \err{0.58}{0.08}  & \err{1.6}{0.17}   & 0.0 & 0.0 \\ 
  & \ilrlc{}$^{\dagger}$    & \err{\bm{0.37}}{0.07}  & \err{\bf{1.1}}{0.15}  & 0.0 & 0.0  \\ 
  & \kivilr{}   & \err{\bm{0.37}}{0.07} & \na{} & \na{} & \na{}  \\
  & Only LC   & \err{15.03}{0.20} & \err{32.6}{0.14} & 0.0 & 0.0  \\ 
  & 2SLS   & $>200$  & $>5$k & 0.0 & 0.0  \\ 
  \midrule
  
  \multirow{3}{*}{\shortstack{$p=30$ \\$q=10$}} 
  & \ilrlc{}    & \err{\bm{0.42}}{0.08}  & \err{\bm{0.22}}{0.01} & 0.0 & 12.0 \\
  & \kivilr{}    & \err{240.6}{35.7}  & \na{} & \na{}& \na{}\\ 
  & Only LC   & \err{24.4}{0.37} & \err{1.9}{0.00} & 0.0 & 12.3  \\ 
  \midrule
  
  \multirow{3}{*}{\shortstack{$p=250$ \\$q=10$}}
  & \ilrlc{}    & \err{\bm{0.67}}{0.14} & \err{\bm{0.22}}{0.02}    & 0.0 & 0.0  \\
  & \kivilr{}    & \err{5060.5}{1196.2}  & \na{}  & \na{} & \na{}\\
  & Only LC   & \err{30.8}{0.48}  & \err{143.3}{0.27} & 3.0 & 1.0  \\ 
  \bottomrule
\end{tabular}}\\

{\footnotesize {}$^{\dagger}$ Identical to \ilrilr{} in low-dimensional setting without sparsity.}
\caption{Results for setting A (fully linear in $\ilr(X)$).} \label{tab:res_all_settinga}
\end{table*}

\subsection*{Results for Low-Dimensional Compositions}

We first consider settings A and B with $p=3$ and $q=2$.
The top section of \cref{tab:res_all_settinga,tab:res_all_settingb} shows our metrics for all methods.
First, effect estimates are far off when ignoring the compositional nature (2SLS) or the confounding (Only LC) as expected.
Also, recent non-linear IV methods such as 
\cite{hartford2017deep,bennett2019deep,zhang2020maximum} 
could not overcome the issues of 2SLS in this setting.
Without sparsity in the second stage, \ilrilr{} and \ilrlc{} yield equivalent estimates in this low-dimensional linear setting---we only report \ilrlc{}.
\ilrlc{} (and equivalent methods) succeed in cause-effect estimation under unobserved confounding: they recover the true causal parameters with high precision on average (low $\beta$-MSE) and thus achieve low OOS MSE.
While DIR+LC performs reasonably well in setting A, setting B surfaces that despite being a seemingly plausible approach with powerful regression techniques, DIR+LC suffers substantially under a misspecified first-stage.

\begin{table*}
\centering
{
\begin{tabular}{p{0.08\textwidth}crrrrr}
\toprule
  & & \multicolumn{4}{c}{\textbf{Setting B, \cref{eq:semisyntheticdata}}} \\
  \cmidrule(lr){3-6}
  \thead{Dim.} & \thead{Method} & \textbf{OOS MSE} & \thead{$\beta$-MSE} & \thead{FZ} & \thead{FNZ} \\ 
  \midrule
  
  \multirow{5}{*}{\shortstack{$p=3$ \\$q=2$}}
  & DIR+LC    & $>10$k & $>2$k & 0.0 & 0.0 \\ 
  & \ilrlc{}$^{\dagger}$    & \err{\bm{20.3}}{4.85}  & \err{\bm{9.9}}{3.3}  & 0.0 & 0.0 \\ 
  & \kivilr{}  & \err{\bm{19.2}}{4.42}  & \na{} & \na{} & \na{} \\
  & Only LC   &\err{269.0}{6.85} & \err{129.8}{2.13} & 0.0 & 0.0 \\ 
  & 2SLS   & $>15$k & $>300$k  & 0.0 & 0.0 \\ 
  \midrule
  
  \multirow{3}{*}{\shortstack{$p=30$ \\$q=10$}} 
  & \ilrlc{}    &  \err{\bm{120.4}}{25.1}   & \err{\bm{37.0}}{15.1}  & 0.0 & 13.3\\
  & \kivilr{}    & \err{287.2}{19.1}   & \na{} & \na{} & \na{}\\
  & Only LC   & \err{3863.8}{166.3} & \err{458.8}{12.2} & 3.4 & 15.8 \\ 
  \midrule
  
  \multirow{3}{*}{\shortstack{$p=250$ \\$q=10$}}
  & \ilrlc{}    & \err{\bm{99.1}}{7.9}   & \err{\bm{24.4}}{4.30}  & 0.13 & 0.39 \\
  & \kivilr{}    &  \err{622.8}{30.1}   & \na{}  & \na{} & \na{} \\
  & Only LC   &  \err{3366.0}{166.3} & \err{498.3}{17.2} & 6.9  & 1.9 \\ 
  \bottomrule
\end{tabular}}\\
{\footnotesize {}$^{\dagger}$ Identical to \ilrilr{} in low-dimensional setting without sparsity.}
\caption{Results for setting B (first stage ZINB with sparse effects in higher dimensions), where all our two-stage methods are (intentionally) misspecified in the first stage.} \label{tab:res_all_settingb}
\end{table*}

\begin{figure}[ht]
  \centering
  \includegraphics[width=0.3\linewidth]{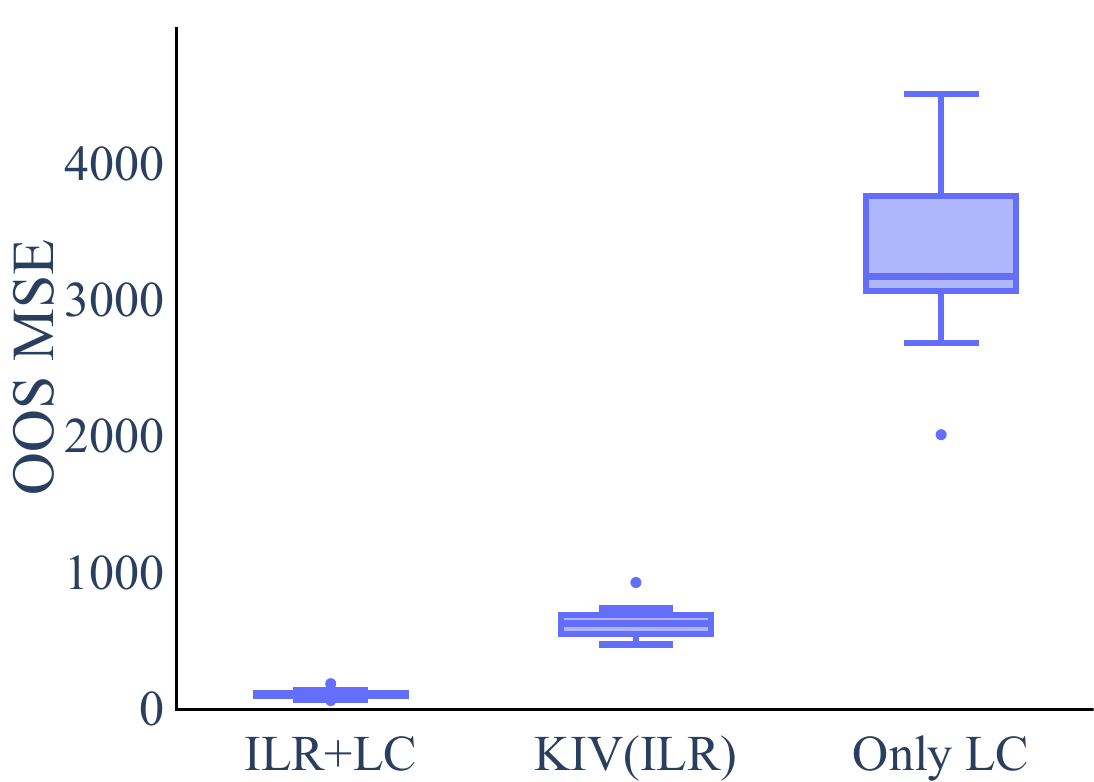}
  \hfill
  \includegraphics[width=0.3\linewidth]{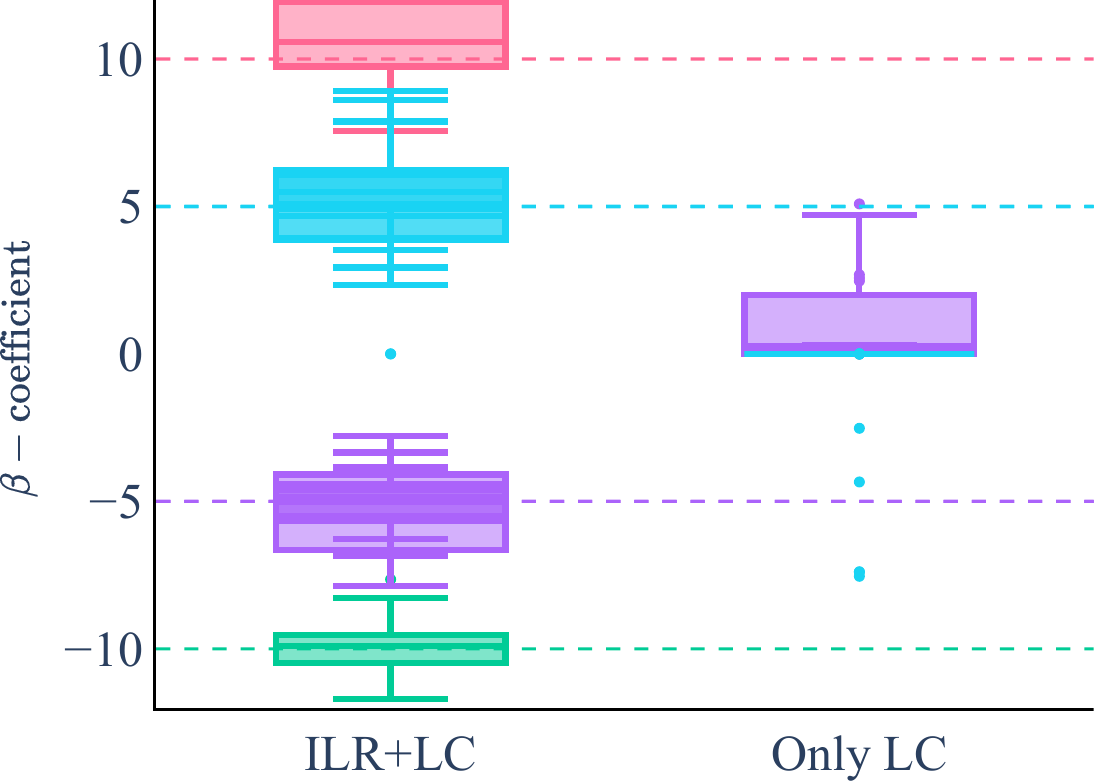}
  \hfill
  \includegraphics[width=0.3\linewidth]{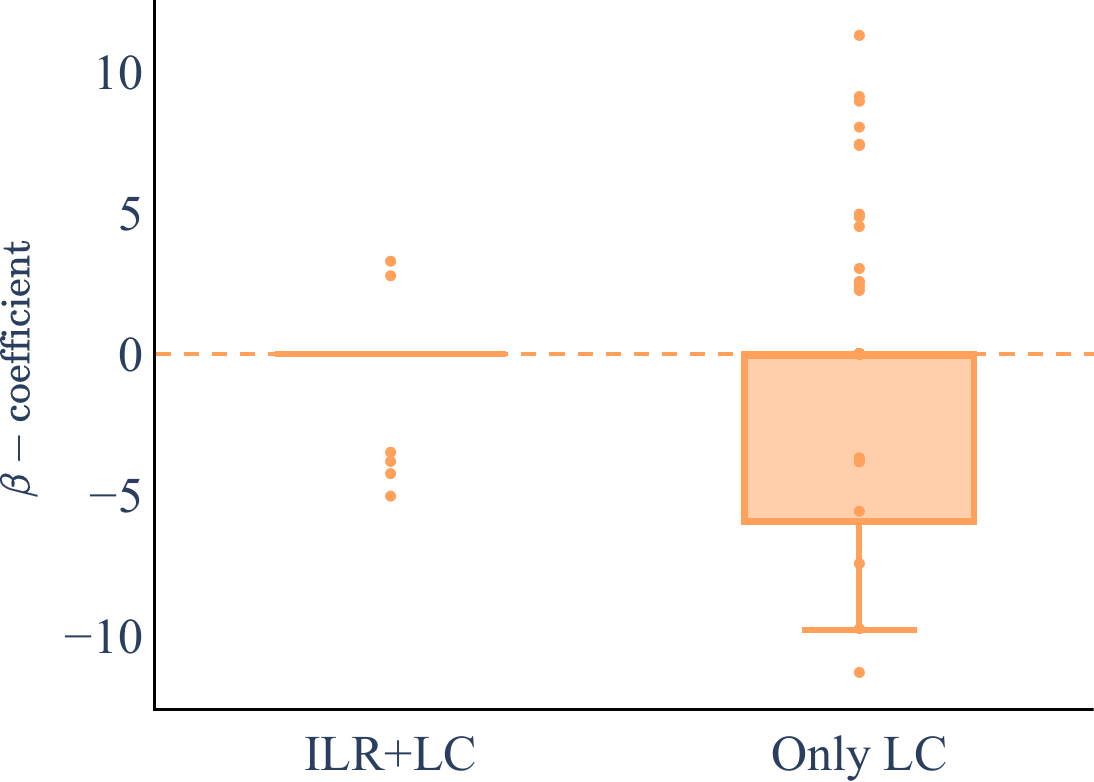}
\caption{Boxplots of the results for setting B in \cref{tab:res_all_settingb} with $p=250, q=10$.
We show OOS MSE (left), recovery of non-zero $\beta$ coefficients (middle), and recovery of zero $\beta$ coefficients.}
\label{fig:res_negbinom250}
\end{figure}

\subsection*{Results for High-Dimensional Compositions}
We now consider the challenging cases $p=30$ and $p=250$ with $q=10$ and sparse ground truth $\beta$ for settings A and B (8 non-zeros: 3 times $-5$ and $5$ and once $-10$ and $10$) in the bottom sections of \cref{tab:res_all_settinga,tab:res_all_settingb}.
\ilrlc{} deals well with sparsity: unlike Only LC, it identifies non-zero parameters perfectly ($\mathrm{FZ}=0$) and rarely predicts false non-zeros.
It also identifies the true $\beta$ and accordingly predicts interventional effects (OOS MSE) well.
DIR+LC and \ilrilr{} fail entirely in these settings because the optimization does not converge.
While we could get \kivilr{} to return a solution, tuning the kernel hyperparameters for high-dimensional $\ilr$ coordinates becomes increasingly challenging, which is reflected in poor OOS MSE.
In \cref{fig:res_negbinom250} we show detailed results for the most challenging setting (setting B with $p=250$ and $q=10$) including the OOS MSE (left), recovery of individual non-zero coefficients (middle), and recovery of zero coefficients (right). Analogous plots for all other settings can be found in \cref{supp:method_results}.

\begin{figure}[ht]
  \phantom{} \hspace{1.8cm} weak instrument \hspace{3.8cm}  weak instrument \hspace{3cm} non-linear second stage\\
  \includegraphics[width=0.3\linewidth]{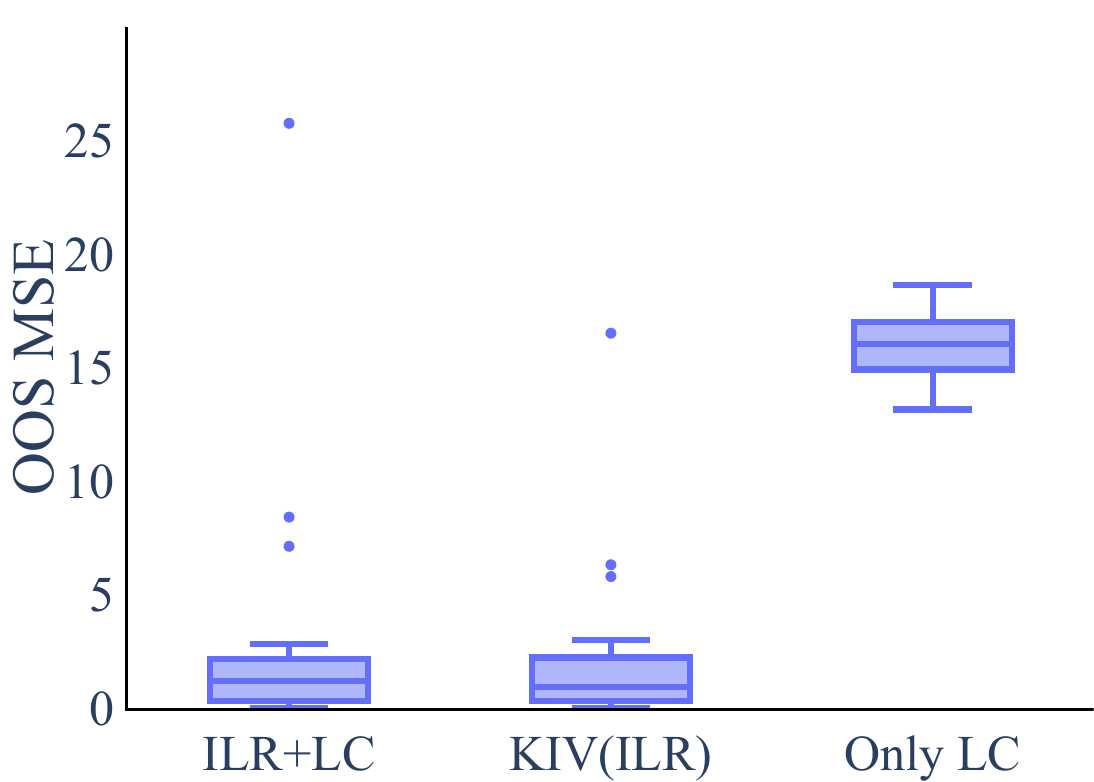}
  \hfill
  \includegraphics[width=0.3\linewidth]{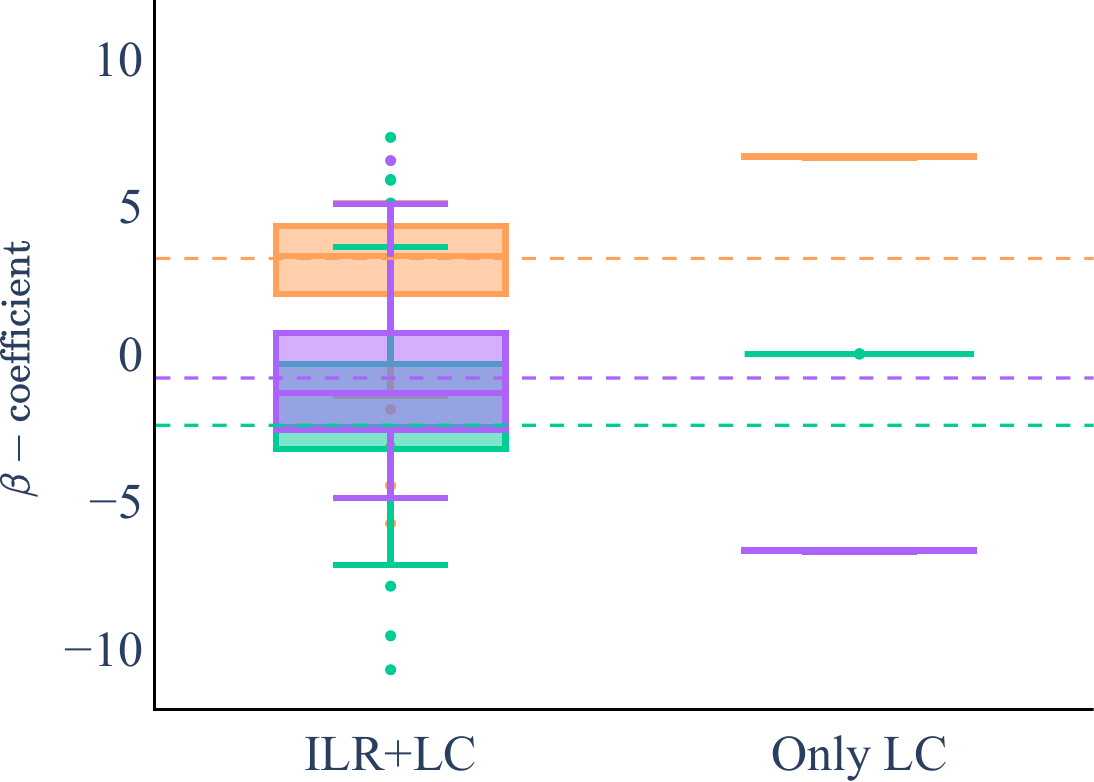}
  \hfill
  \includegraphics[width=0.3\linewidth]{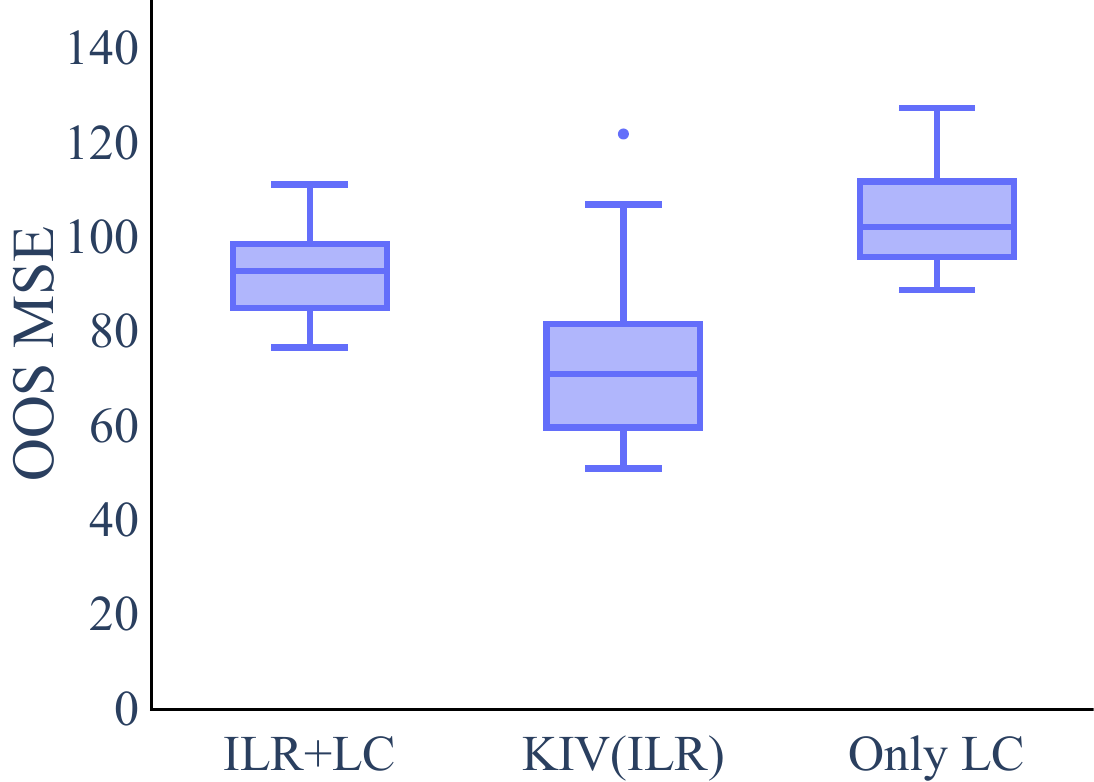}
  \bigskip
  \hrule
  \bigskip
  \phantom{} \hspace{2cm} scarce data \hspace{3.8cm}  scarce data (non-zero $\beta$s) \hspace{2.5cm} scarce data (zero $\beta$s) \\
  \includegraphics[width=0.3\linewidth]{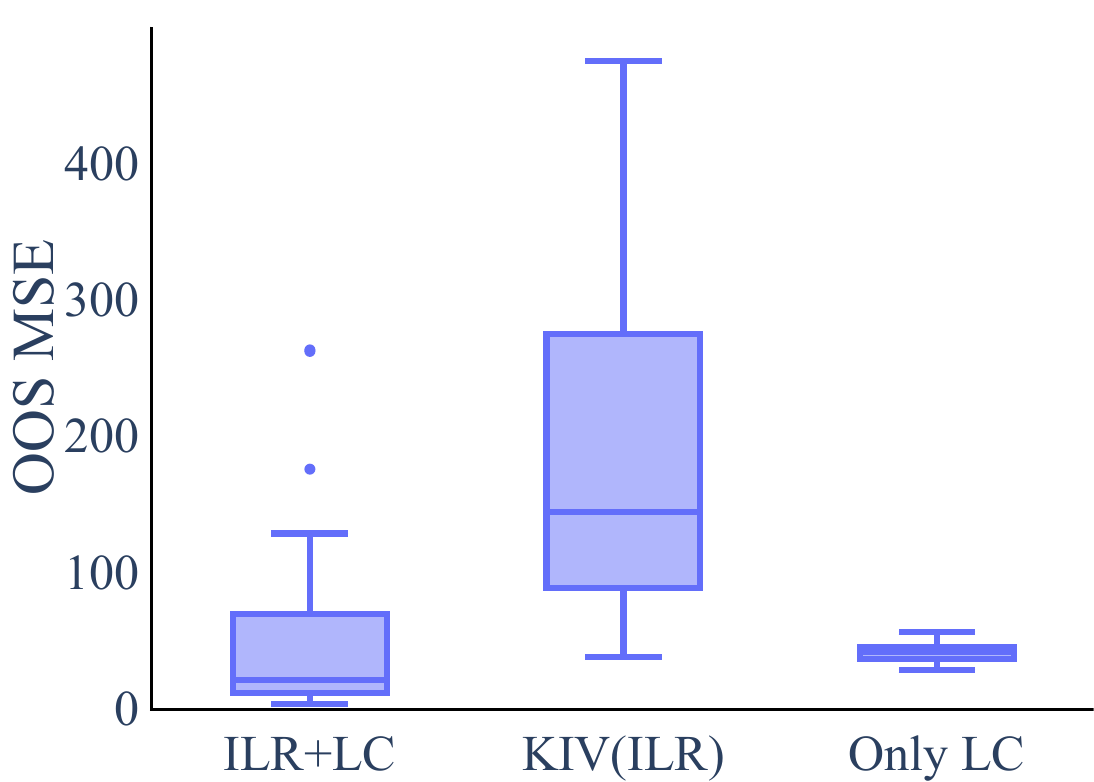}
  \hfill
  \includegraphics[width=0.3\linewidth]{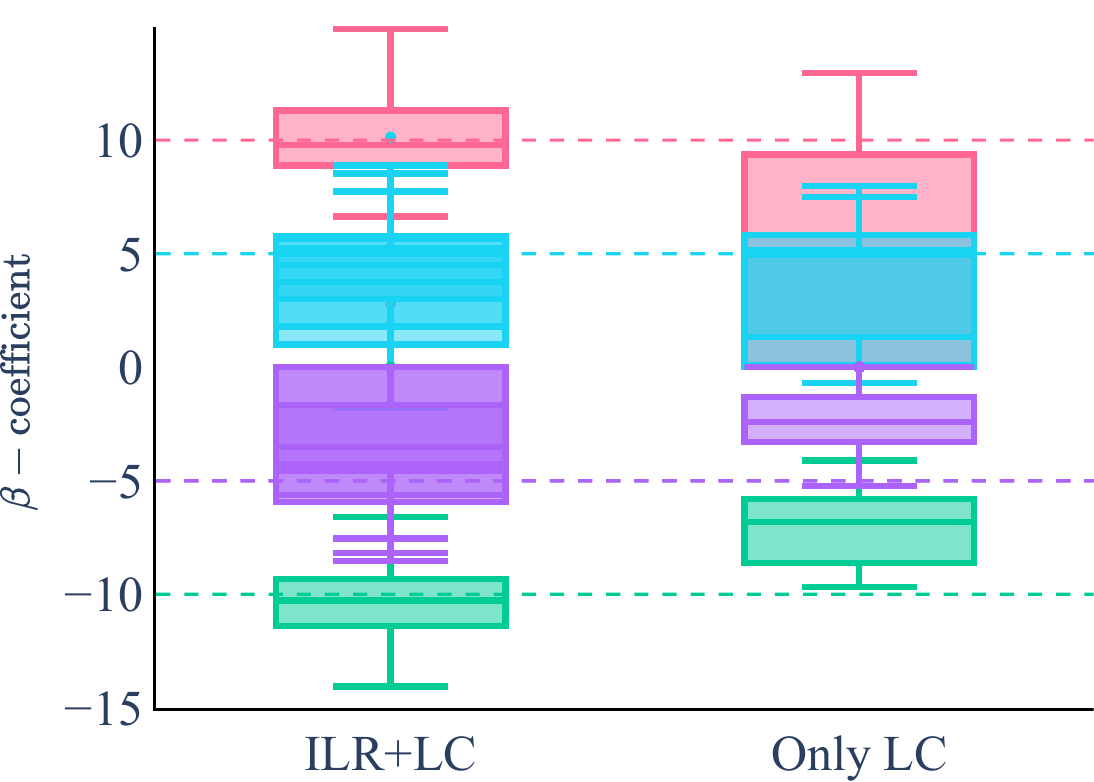}
  \hfill
  \includegraphics[width=0.3\linewidth]{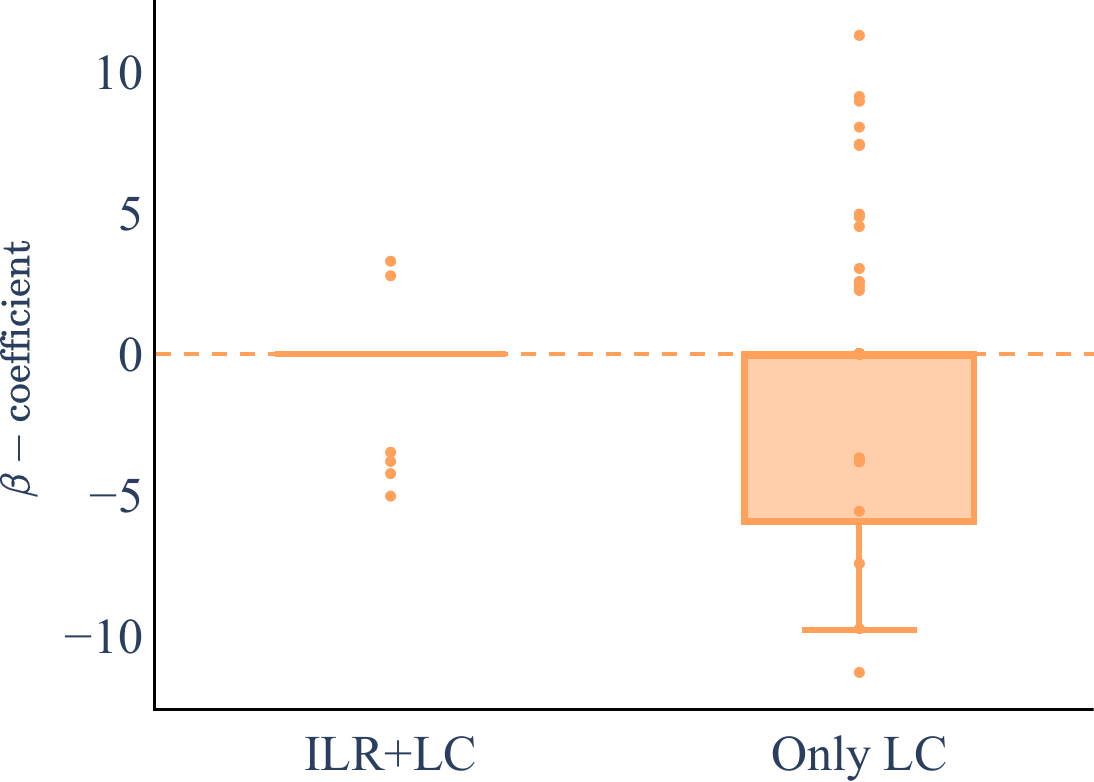}
\caption{We display OOS MSE and $\beta$-MSE (for non-zero coefficients and where applicable) for our robustness checks.
All results and further visualizations are in \cref{supp:method_results}.}\label{fig:res_robust}
\end{figure}

\subsection*{Robustness Checks}\label{sec:robust}
Due to the inherent entanglement via the unit sum constraint, analyses involving compositional data are generically hard to interpret. 
Causal analyses involving compositions in the instrumental variable setting are further challenged by potential violations of assumptions such as weak instruments or misspecification.
We assess the sensitivity of our proposed methodology to such potential pitfalls in the following scenarios.

\subsubsection*{Weak Instruments}
``Strong instruments'' are a prerequisite for successful two-stage estimation in the instrumental variable setting and one of the key discussion points in real-world applications of IV.
Nevertheless, how to measure instrument validity is not unambiguously clarified, relying on heuristics and empirically derived best practices.
In the linear setting, instrument strength for $p=1$ can be approximated via a first-stage F-statistic with a value greater than $10$ generally being considered sufficient to avoid weak instrument bias in 2SLS \citep{andrews2019weak}.
For $p>1$, measuring instrument strength is more challenging even in the linear case \citep{sanderson20162ftest}. 
Therefore, we report first-stage F-statistics for each dimension of $X$ as a proxy for instrument strength.

When instruments are weak, the estimation bias can theoretically become arbitrarily large (even in the limit of infinite data).
To assess the sensitivity of our methods to weak instrument bias, we re-analyze setting A ($p=3$ and $q=2$) only changing the dependence of $X$ and $Z$ to be weak with first-stage F-statistic values of $6.9$ and $4.7$ for the two components of $\ilr(X)$.
In the linear setting, we can directly control instrument strength via $\alpha$ (see \cref{eq:syntheticdata}).

The first row in \cref{tab:res_robust} summarizes our results for weak instruments: the two-stage methods have a substantially higher variation in their estimates, both for OOS MSE and $\hat{\beta}$ compared to the strong instrument setting in \cref{tab:res_all_settinga,tab:res_all_settingb}.
As the second stage has not changed, Only LC performs equally bad.
Notably, while the wellspecified two-stage methods \ilrlc{} and \ilrilr{} seem to do worse than Only LC, the large OOS MSE and $\beta$-MSE are mostly due to outliers. \Cref{fig:res_robust} shows that the range of $\beta$ estimates still cover the true values for \ilrlc{} and \ilrilr{}, while Only LC is systematically off with low variance (confidently wrong).
DIR+LC now not only suffers from the misspecified first stage but also the weak instrument resulting in virtually useless estimates.
The surprisingly good performance of \kivilr{} in this specific setting is unexpected and cannot be consistently reproduced over different weak instrument scenarios: the performance is highly volatile and often worse than \ilrlc.
Therefore, despite the good performance for these specific parameters, we find that more flexible methods are also affected heavily by weak instruments.
In general, while two-stage estimates generally cannot be broadly trusted when instruments are weak, reverting to Only LC is potentially even more detrimental because the estimated coefficients are systematically off.

\subsubsection*{Non-linear Second Stage}
Well-specification is typically impossible to ascertain in practice and most real-world examples are likely not perfectly linear even when the linearity assumption can be defended.
Therefore, we introduce a non-linear~$f$ for setting A with $p=3$ and $q=2$ (\cref{eq:syntheticdatamis}), resulting in a misspecified second stage for all our methods except \kivilr{}, which can in principle capture non-linearities.
Note that $\beta$ cannot be interpreted directly as causal parameters when the true causal effect depends non-linearly on $\ilr(X)$.
The results in the second row of \cref{tab:res_robust} show that DIR+LC (doubly misspecified) and 2SLS (ignoring compositionality) again fail.
Moreover, in this non-linear scenario \kivilr{} beats \ilrlc{} (both still outperforming Only LC) and we expect the difference to grow as the non-linearity of~$f$ increases.

\subsubsection*{Scarce Data}
Finally, we return to the original setting A ($p=250$, $q=10$, linear  in both stages), but mimic a scarce data scenario with $n=100$. 
The third row in \cref{tab:res_robust} clearly highlights again how the lack of regularization becomes problematic for \ilrilr{} and \kivilr{}.
Compared to the larger dataset, also our regularized two-stage methods naturally exhibit higher variation in their estimates.
Notably, Only LC appears to compare favorably to \ilrlc{} in OOS\ MSE, but $\beta$-MSE surfaces its failure to accurately recover causal parameters.
We thus conclude that despite increased variability, the \ilrlc{} is still better equipped to recover $\beta$ in the small data regime (see \cref{fig:res_robust}).

\begin{table}[ht]
\centering

\begin{tabular}{lcrr}
\toprule
 \thead{Scenario} & \thead{Method} & \thead{OOS MSE} & \thead{$\beta$-MSE} \\ 

\midrule
  \multirow{5}{*}{Weak Instruments}
  & DIR+LC    &  \err{7.1}{2.6}  & \err{43.2}{10.3}  \\
  & \ilrlc{}$^{\dagger}$    &  \err{3.6}{1.3}  & \err{26.0}{5.6}  \\
  & \kivilr{}   & \err{\bm{2.7}}{0.9}  & \na{} \\ 
  & Only LC   &\err{15.9}{0.20} & \err{52.2}{0.18} \\ 
  & 2SLS   & $>100$ & $>5$k  \\ 
  \midrule

  \multirow{5}{*}{Non-Linearity}
  & DIR+LC    &  \err{{135.6}}{6.34} & \na{} \\
  & \ilrlc{}$^{\dagger}$    &  \err{{92.0}}{1.2}  & \na{} \\
  & \kivilr{}   & \err{\bm{73.4}}{2.28}  & \na{} \\ 
  & Only LC   &\err{104.1}{1.43} & \na{} \\
  & 2SLS   & $>300$  & \na{} \\
  \midrule

  \multirow{3}{*}{Scarce Data}
  & \ilrlc{}    & \err{\bm{45.1}}{7.90}   & \err{\bm{72.8}}{8.3}  \\
  & \kivilr{}    & \err{290.8}{62.5}   & \na{}  \\
  & Only LC   & \err{\bm{40.5}}{1.00} & \err{196.4}{8.7} \\
  & \ilrilr   & $>10$k & $>2 \cdot 10^{24}$ \\
  \bottomrule
\end{tabular}%
\\
{\footnotesize {}$^{\dagger}$ Identical to \ilrilr{} in low-dimensional setting without sparsity.}
\caption{Results for various robustness checks.} \label{tab:res_robust}
\end{table}

\section*{Case study on murine sub-therapeutic antibiotic treatment} 
\label{sec:realdata}

We consider the mouse dataset described by 
\cite{Schulfer2019} 
and analyzed in 
\cite{Wang2020} 
using causal mediation. 
A total of 57 newborn mice were assigned randomly to a sub-therapeutic antibiotic treatment (STAT) during their early stages of development.
Sub-therapeutic antibiotic treatment means that the administered doses of antibiotics are too small to be detectable in the mice' bloodstream.
There were 35 mice in the treatment group and 22 mice in the control group.
After 21 days, the gut microbiome composition of each mouse was recorded. 
We are interested in the causal effect of the gut microbiome composition on body weight $Y \in \bR$ of the mice (at sacrifice). 

We assume a valid instrument due to the following characteristics in the data generation:
The random assignment of the antibiotic treatment ensures independence of potential confounders such as genetic factors ($Z \indep U$).
The sub-therapeutic dose implies that antibiotics can not be detected in the mice' blood, providing reason to assume no effect of the antibiotics on the weight other than through its effect on the gut microbiome ($Z \indep Y \mid \{U, X\}$).

Finally, we observe empirically, that there are statistically significant differences of microbiome compositions between the treatment and control groups ($Z\dep X$) based on the first stage F-statistic. Thus, the sub-therapeutic antibiotic treatment is a good candidate for an instrument $Z \in \{0, 1\}$ in estimating the effect $X 
\to Y$. Note, however, that this work is focused on methods rather than novel biological insights as more scrutiny of the IV assumptions would be 
required for substantive biological claims.

\begin{figure}[ht]
  \centering
  \includegraphics[width=.6\linewidth]{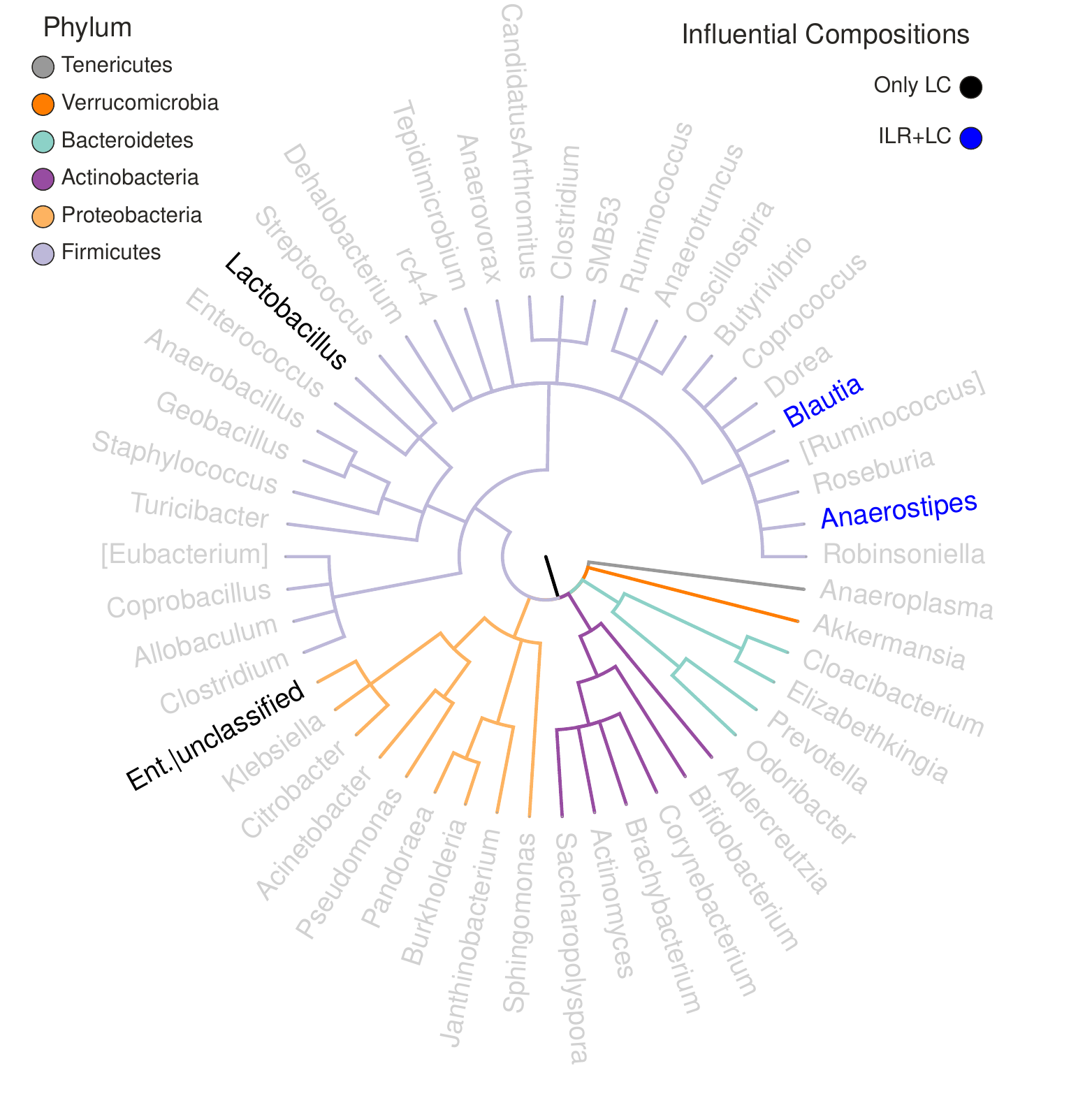}
  \caption{Taxonomic tree of the microbiome data at genus level. The influential log-ratios for both Only Second LC  and \ilrlc{} are highlighted in black and blue, respectively.
  }\label{fig:realdata}
\end{figure}

\Cref{fig:realdata} highlights the two most influential microbes on the genus level for our two-stage \ilrlc{} estimator and to the non-causal baseline Only LC, respectively. In the causal setting, we estimate the log-ratio of 
Blautia to Anaerostipes to be most influential for weight gain whereas standard log-contrast regression deems the ratio 
of an unclassified Enterobacteria genus to Lactobacillus to be the most predictive genus pair. This discrepancy 
suggests that the second stage might be subject to confounding. However, the mediation analysis on the same dataset in 
\cite{Wang2020} 
posits a negative mediation effect of Lactobacillus on weight gain, consistent with the non-causal 
baseline model. This highlights the fact that different causal models provide alternative interpretation of the data 
that can only resolved by follow-up biological experiments.   

Finally, we also assessed the influence of taxonomic aggregation levels and different loss functions on the results 
(see \cref{supp:realdataexample}). We observed that our causal model is robust to the choice of 
the loss function in terms of selected taxa whereas the baseline model found loss-function depdendent sets of 
predictive taxa (see Supplementary Material \cref{appfig:agglevel_binary}). 

\section*{Discussion}
In this work, we developed and analyzed methods for cause-effect estimation with compositional causes under unobserved confounding in instrumental variable settings.
First, we succinctly expose that the common portrayal of summary statistics as a decisive (rather than merely descriptive) description of compositions is misguided. Instead, we advocate for causal effects to be estimated from the entire composition vector directly to establish meaningful and interpretable causal links.
As a result, analysts cannot tap into a collection of well established cause-effect estimation tools for scalar data, but are instead faced with a large number of possible components (calling for sparsity-enforcing methods) and typically have to deal with unobserved confounding.
Given the potentially profound impact of microbiome or single cell RNA data on advancing human health or of species abundances on global health, it is of vital importance that we face these challenges and develop interpretable methods to obtain causal insights from compositional data.

To this end, we carefully developed and assessed the effectiveness of various methods to not only reliably recover causal effects (OOS MSE), but also yield \emph{interpretable and sparse} effect estimates for individual abundances ($\beta$-MSE, FZ, FNZ) whenever applicable.
We also put special emphasis on how IV assumptions (misspecification, weak instrument bias) interact with compositionality.
Our extensive empirical results for different two-stage methods highlight that accounting for the compositional nature as well as confounding is not optional.
The overall failure of DIR+LC shows that not any seemingly suitable compositional technique can be trusted to yield reliable estimates in a manual two-stage procedure---careful analysis is needed.
We have identified \ilrlc{}, to work reliably in wellspecified sparse and non-sparse settings as well as being relatively robust to first- and (small) second-stage misspecifications (i.e., non-linearities) and scarce data.
It also yields interpretable estimates for individual components.
When interpretability is not required or second-stage non-linearities are strong, \kivilr{} can  still perform well under these relaxed assumptions albeit being challenging to tune for large $p$ and unable to incorporate sparsity.
As expected, valid instruments are required for all our two-stage methods.
Taken together, our results on the efficacy and robustness of our methods in simulation and on real microbiome data provide first recommendations for practitioners to fully integrate compositional data into cause-effect estimation.

\section*{Acknowledgments}
We thank Dr.~Chan Wang and Dr.~Huilin Li, NYU Langone Medical Center, for kindly providing the
pre-processed murine amplicon and associated phenotype data used in this study. We thank L\'eo
Simpson, TU München, and Alice Sommer, LMU M\"unchen, for kindly and patiently providing their 
technical and scientific support. 

EA is supported by the Helmholtz Association under the joint research
school ``Munich School for Data Science - MUDS''. \vspace*{-12pt}

\newpage

\bibliography{reference}
\newpage

\section*{Author contributions statement}

NK, CLM and EA wrote the manuscript. NK and CLM reviewed the manuscript. EA conducted the analysis.

\section*{Additional Information}
No competing interest is declared.

\newpage

\renewcommand{\thesection}{S\arabic{section}}  
\setcounter{section}{0}

\renewcommand{\thetable}{S\arabic{table}}  
\setcounter{table}{0}
\renewcommand{\thefigure}{S\arabic{figure}}
\setcounter{figure}{0}


\section{Supplementary Material}
The supplementary material contains details on compositional data transformations (\cref{supp:compositional_transformations}) and the applied instrumental variables methods (\cref{supp:ivmethods}) as well as a list of the packages that have been used in the implementation (\cref{supp:packages}).
Further, the supplementary material provides additional results for the real data example of \cite{Schulfer2019} (\cref{supp:realdataexample}).
Moreover for the synthetic settings it holds a detailed description of the data generation (\cref{supp:data_generation}), the parameter settings for the training of the methods (\cref{supp:method_training}) as well as additional results and visualizations (\cref{supp:method_results}).

\section{Compositional Data Transformations}
\label{supp:compositional_transformations}
Given a compositional vector $x \in \simp^{p-1}$, the definitions of the log-transformations
are given by the additive log-ratio transformation
\begin{equation}
      \alr(x) := \Bigl (\log \frac{x_1}{x_p}, \ldots, \log \frac{x_{p-1}}{x_p} \Bigr) = \tilde{x}_{\alr}=\log(x) \cdot 
       \begin{bmatrix} 1 & 0 & \cdots & 0 \\ 0 & 1 & \cdots & 0 \\ \vdots & \vdots  & \ddots & \vdots \\ 0 & 0 & \cdots & 1 \\ -1 & -1 & \cdots & -1 \end{bmatrix}
\end{equation}
with inverse
\begin{equation}
    \alr^{-1}(\tilde{x}) = C(\exp([\tilde{x}, 0])),
\end{equation}
the  centered log-ratio transformation
\begin{equation}
  \clr(x) := \biggl( \log \frac{x_1}{g(x)}, \ldots, \log \frac{x_p}{g(x)} \biggr) = \tilde{x}_{\clr} = \frac{\log(x)}{D} \cdot 
    \begin{bmatrix} D-1 & -1 & \cdots & -1 \\ -1 & D-1 & \cdots & -1 \\ \vdots & \vdots  & \ddots & \vdots \\ -1 & -1 & \cdots & D-1 \end{bmatrix}
\end{equation}
with $g(x) := \sqrt[p]{x_1 \cdot \ldots \cdot x_p}$ and inverse
\begin{equation}
  \clr^{-1}(\tilde{x}) = C(\exp([\tilde{x}])),
\end{equation}
and the isometric log-ratio transformation
\begin{equation}
  \ilr_{V}(x) = \tilde{x}_{\ilr} = \clr(x) \cdot V
\end{equation}
for a matrix $V \in \bR^{p \times p-1}$ such that $V^T V = \mathbb{I}_{p-1}$ providing an orthonormal basis of $\bR^{p-1}$ with inverse
\begin{equation}
  \ilr^{-1}_V(\tilde{x}) = C(\exp([\tilde{x} V^T])).
\end{equation}
For the $\ilr$ transformation, a typical choice for $V^T$ is the so-called Helmert matrix with the first row removed (see for example \url{http://scikit-bio.org/docs/0.4.1/generated/generated/skbio.stats.composition.ilr.html}).

\section{IV Methods}
\label{supp:ivmethods}
We consider three different approaches that gradually relax some of the common IV assumptions.
In particular, the restrictions on the function space of $f$ are gradually relaxed in the different settings.

The Two-Stage Least Squares algorithm (2SLS) consists of two sequential OLS regressions \citep{Angrist1996}.
2SLS is one of the most prominent approaches.
It allows for unobserved confounding while still putting linear restrictions on the function space of $f$ and assuming additive noise:
\begin{equation}
     Y = \beta X + \epsilon_Y
\end{equation}
First, 2SLS fits a regression model based on $Z$ to predict $X$. The second stage uses the estimated $\hat X$ to predict $Y$. This results in the following estimator for $\beta$:
\begin{equation}
     \hat \beta = \big (X^T P_Z X \big)^{-1}\big (X^T P_Z y \big)
\end{equation}
with $P_Z = Z(Z^TZ)^{-1}Z^T$.

If $p=q$, the estimator reduces to the following form:
\begin{equation}
     \hat \beta = \big (Z^T X \big)^{-1}\big (Z^T y \big)
\end{equation}

\cite{singh2019kernel} relax the assumption of the linear setting in 2SLS towards a non-parametric generalization of the causal effect by applying kernel ridge regression (KIV).
\begin{equation}
     Y = f(X) + \epsilon_Y,
\end{equation}
for a potentially non-linear $f$, maintaining the additive noise assumption for point-identifiability.

The OLS regressions are replaced by kernel ridge regressions and thus model the relationship of $Z$, $X$ and $Y$ by non-linear functions in reproducing kernel Hilbert spaces (RKHSs). This method still requires additive noise models to produce consistent results.
Following the arguments in \cite{singh2019kernel}, this gives us a closed form solution for $f$:
\begin{align}
     W &= K_{XX}(K_{ZZ} + n \lambda Id)^{-1}K_{Z \tilde Z} \\
     \hat \alpha &= (W W^T + m \xi K_{XX})^{-1}W \tilde y, \\
     \hat f^m_{\xi}(x) &= (\hat \alpha)^T K_{Xx}
\end{align}

In the next step, we drop the assumption of additive noise, i.e., allowing $f(X, U)$ to depend on the treatment $X$ and any (potentially high-dimensional) confounder $U$ in arbitrary ways (also non-linearly).
This implies that the effect is only partially identifiable, i.e., we can only put lower and upper bounds on $\E[Y \mid do(x)]$.
The authors in \cite{kilbertus2020class}
employ the response function framework to minimize (maximize) the average causal effect over all causal models that satisfy the structural IV assumptions and simultaneously match the observed data to find the lower (upper) bound.
We refer the reader to the original paper for the details \citep{kilbertus2020class}.

\section{Package References}
\label{supp:packages}

Here, we briefly outline the software used in our empirical evaluation. Please note that the code and the requirements are all available at \url{https://github.com/EAiler/causal-compositions}.

\subsection*{Python Packages}
We use the following Python \citep{vanrossum2009python} packages:
Plotly \citep{plotly}, 
Numpy \citep{harris2020numpy}, 
Scipy \citep{mckinney-proc-scipy-2010}, 
scikit-learn \citep{pedregosa2011scikit}, 
scikit-bio \citep{skbio2020}, 
rpy2 \citep{rpy2},
Matplotlib \citep{hunter2007matplotlib},
Statsmodels \citep{seabold2010statsmodels},
Pandas \citep{reback2020pandas},
Jax \citep{bradbury2018jax},
Dirichlet \citep{suh2020dirichlet},
c-Lasso \citep{simpson2021classo}.

\subsection*{R Packages}
We use the following R \citep{rlang} packages:
SpiecEasi \citep{Kurtz2019spieceasi}, 
vegan \citep{oksanen2020vegan}, 
Compositional \citep{tsagris2021compositional} 
and metaSparSim \citep{patuzzi2019metasparsim}.

\newpage
\section{Case study on murine sub-therapeutic antibiotic treatment} \label{supp:realdataexample}

In this part we turn to the analysis of the microbiome instead of the summary statistic as the cause.
This is a more detailed examination of the murine sub-therapeutic antibiotic treatment data given in the main part. 
We provide results on higher aggregation levels, i.e., the taxonomic ranks `Order' and `Family', respectively.
Moreover, we discretize the weight outcome $Y$ and replace the squared loss of the log-contrast regression by a Hinge loss (see \cref{supp:method_training}).

\begin{figure}[!h]
  \centering
  \includegraphics[width=0.45\textwidth]{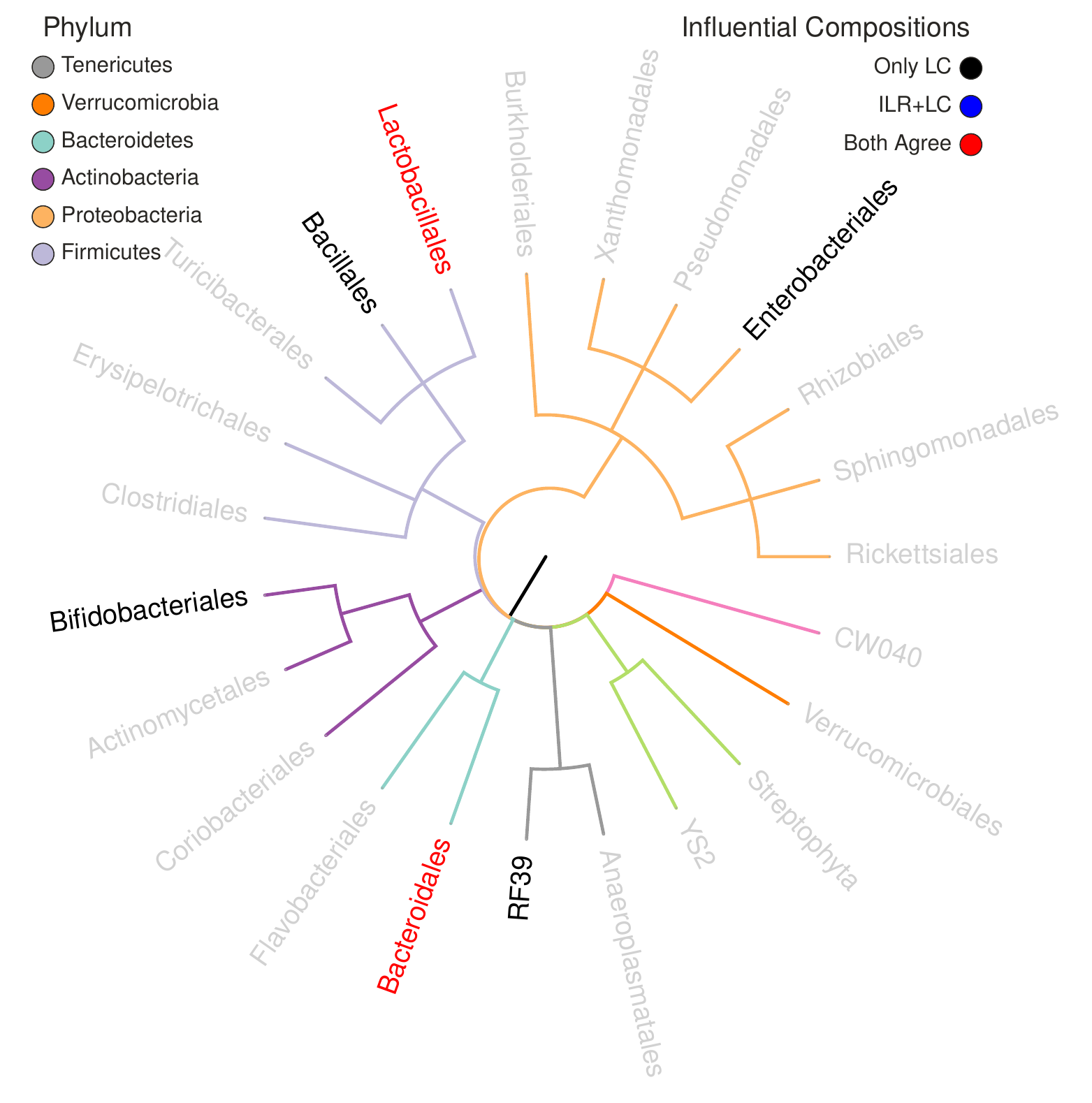} 
  \hfill
  \includegraphics[width=0.45\textwidth]{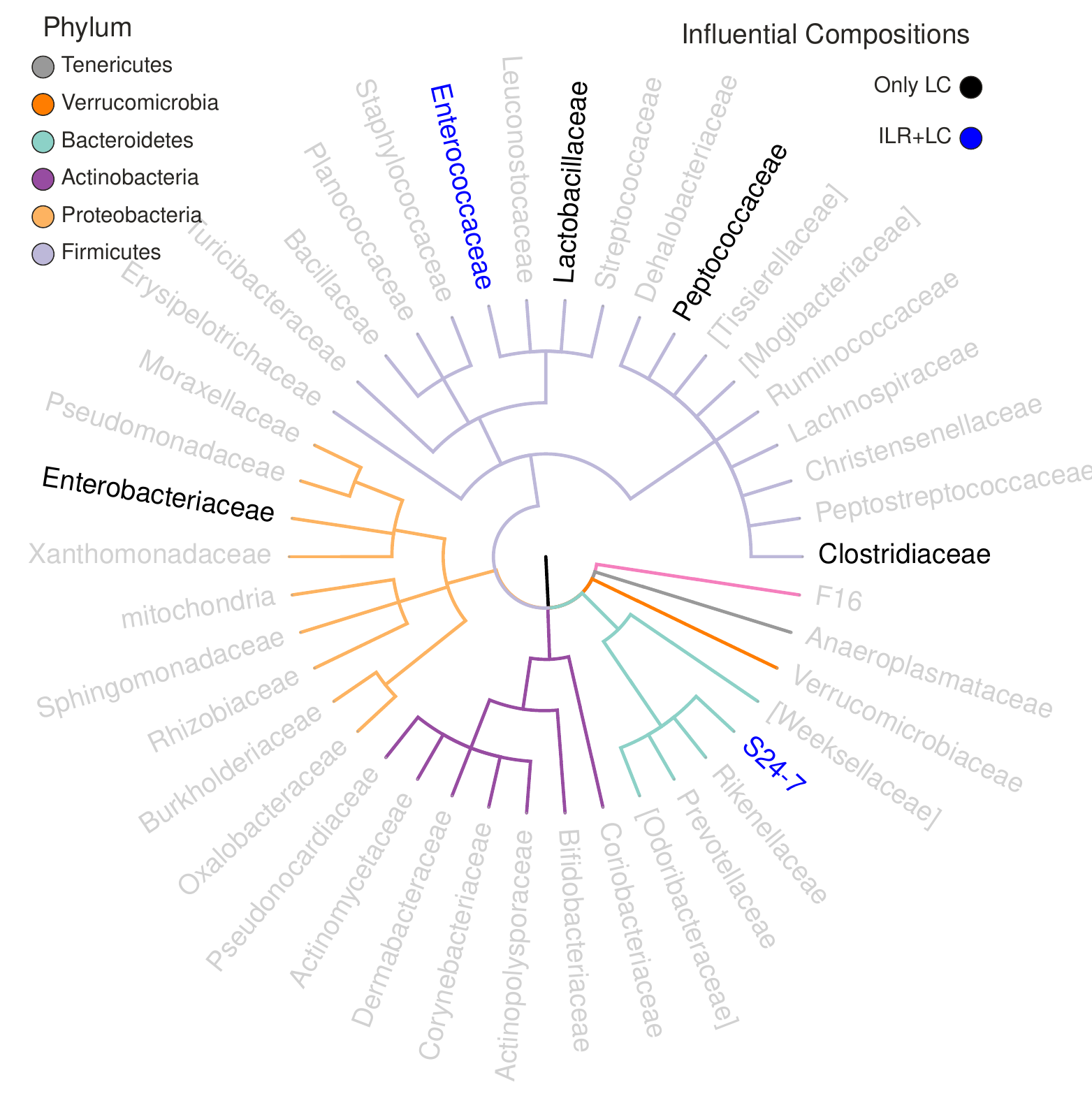} 
  \caption{\textbf{Influential Compositions on Order Level (left) and Family Level (right): } On order level, both methods agree on one influential log ratio. For the family level, there is a divide between the two-stage method and the naive regression. This could suggest that Only LC is subject to confounding on the corresponding aggregation level.}
\label{appfig:agglevel}
\end{figure}

---\emph{Further Results on different Aggregation Levels:} Naturally, for real data, we do not have ground truth labels available. 
However, the importance of being able to draw causal and actionable conclusions becomes apparent.
In the main part we provide the results for the naive regression and the two-stage method \ilrlc{} on genus level.
The methods did not agree on the influential log-ratios, thus suggesting that Only LC might be subject to confounding.
This result also holds true on family level. However, on order level both methods detect one common log-ratio (see \cref{appfig:agglevel}).

\begin{figure}[!tpb]
  \centering
  \includegraphics[width=0.45\textwidth]{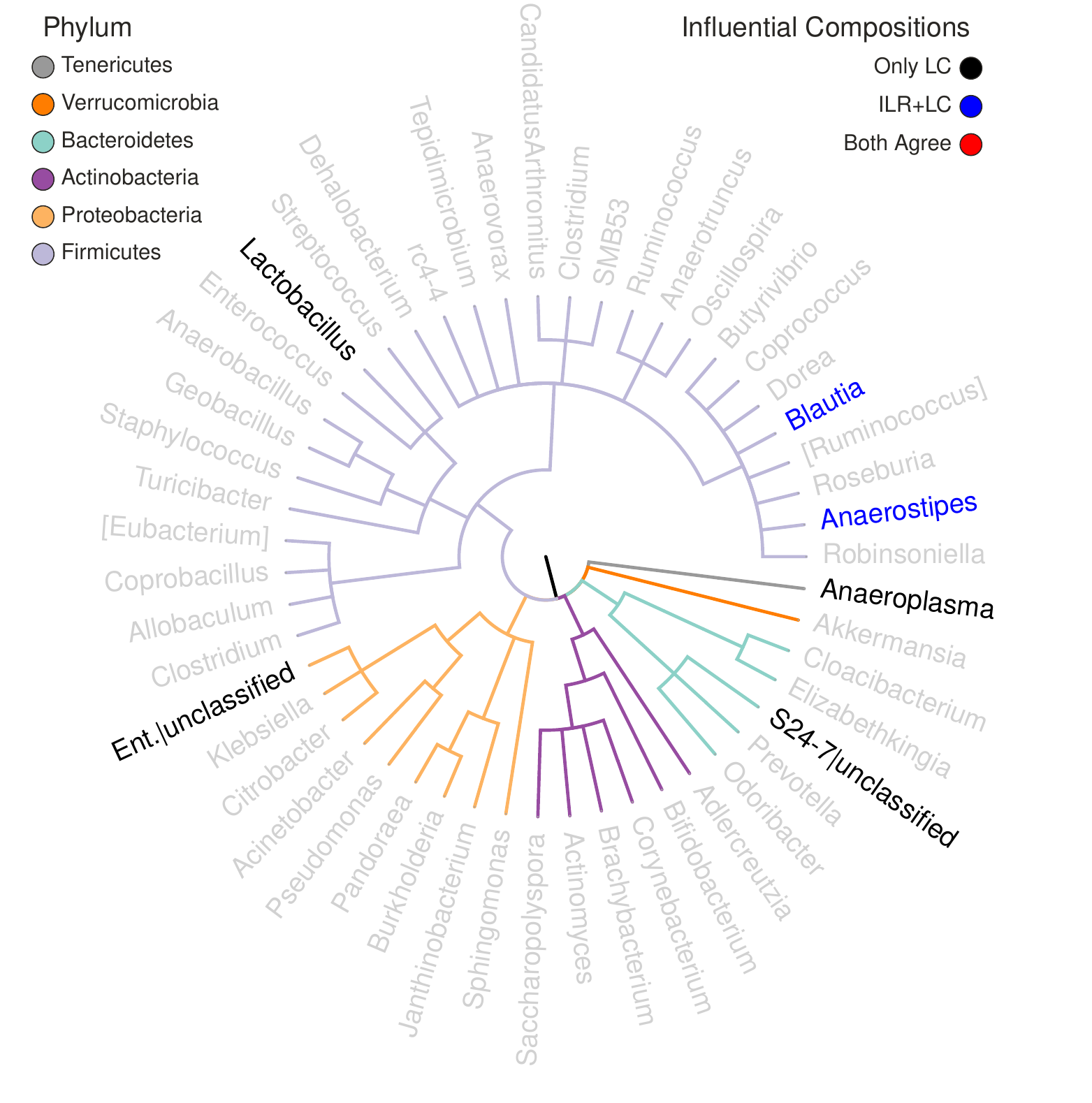} 
  \hfill
  \includegraphics[width=0.45\textwidth]{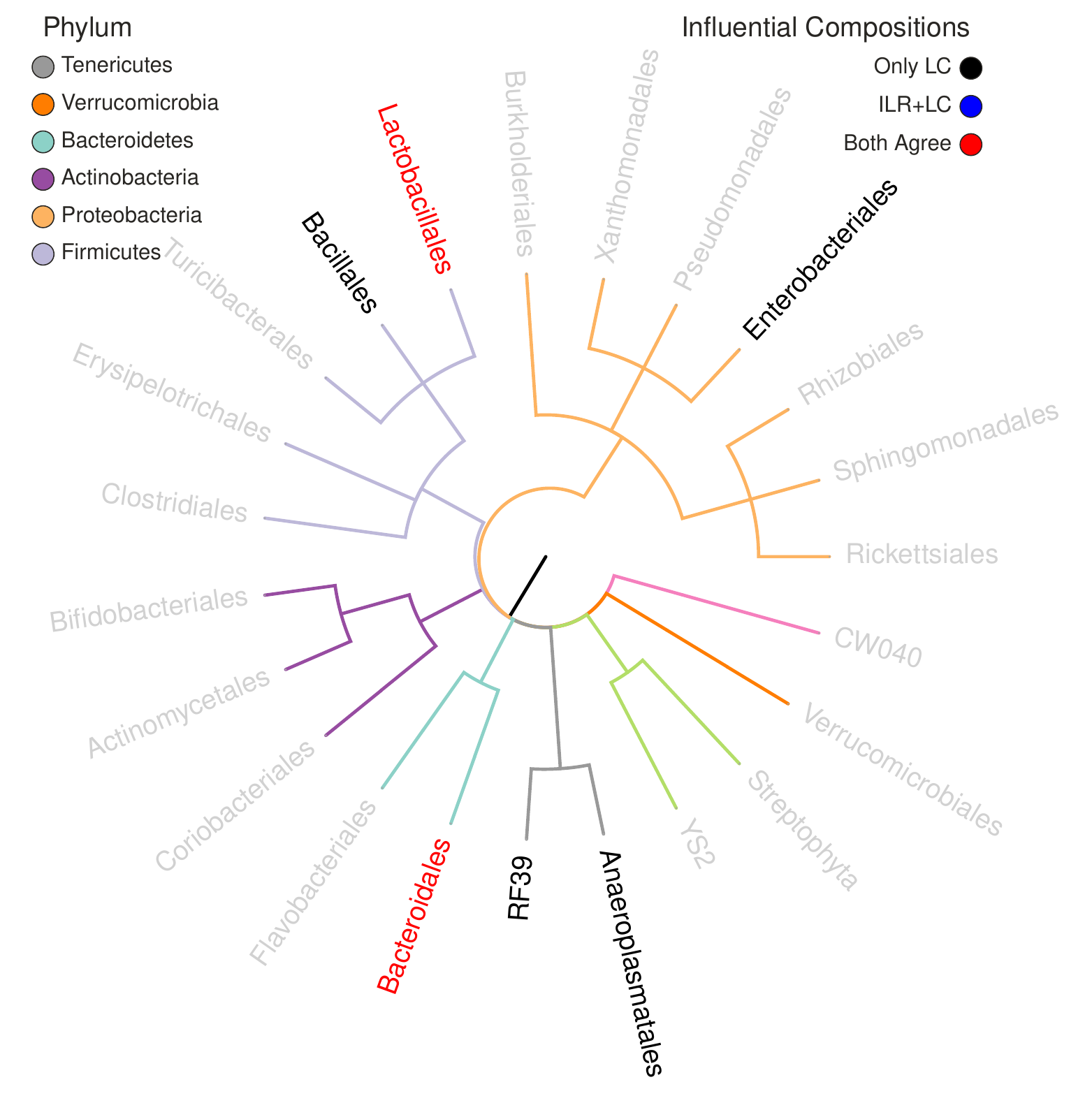} 
  \hfill
  \includegraphics[width=0.45\textwidth]{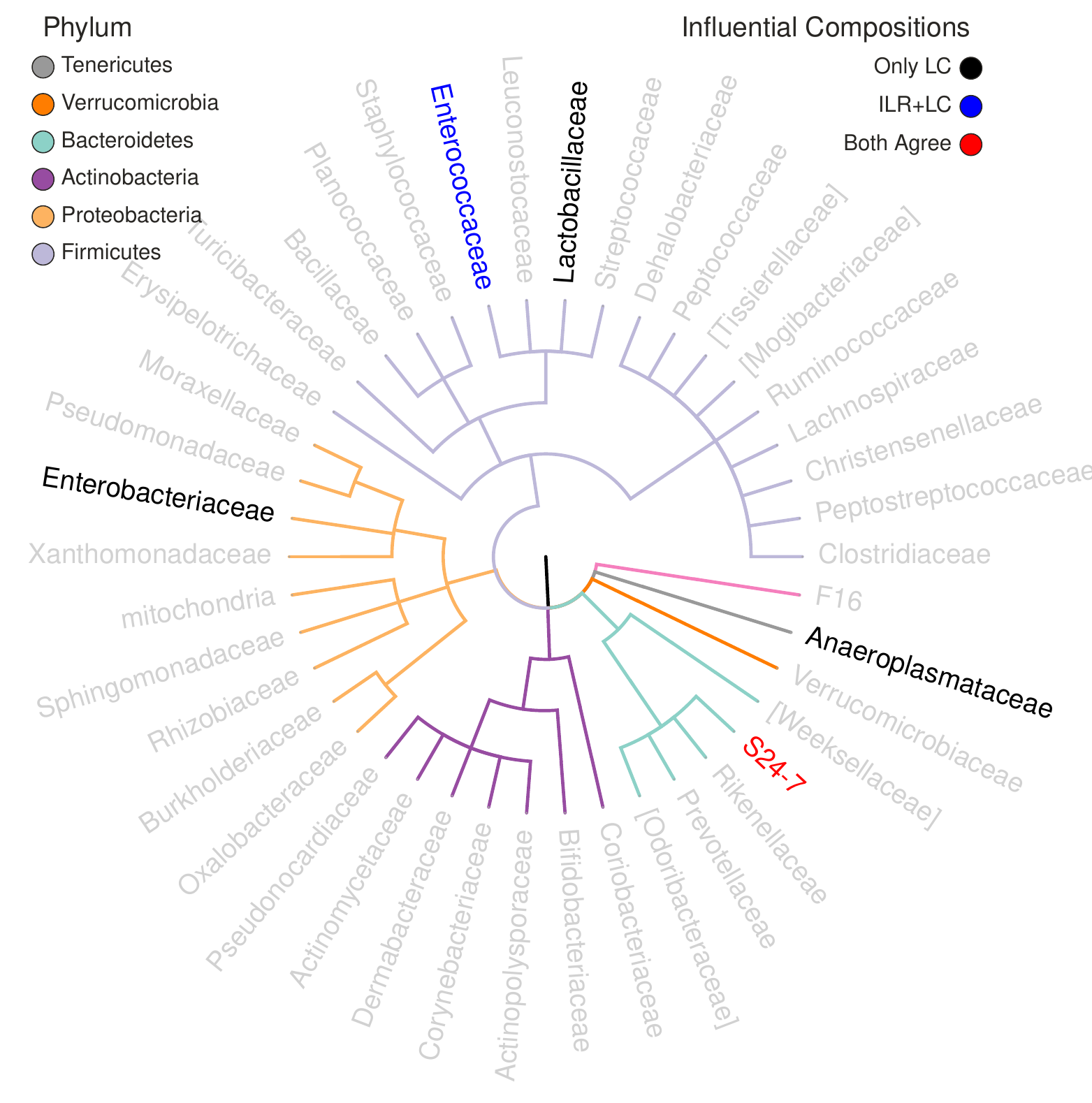} 
  \caption{\textbf{Influential Compositions on Genus Level (top left), Order Level (top right) and Family Level (bottom): } In general, the results stay the same for \ilrlc{} in the binary case and the continuous case. The influential log-ratios are slightly shifting when simply applying a naive regression.}
\label{appfig:agglevel_binary}
\end{figure}

---\emph{Categorical/Binary Outcome:}
In order to provide a more complete picture of the loss possibilities, we include results for a categorical/binary $Y$.
Originally, the real data includes weight measured in gram. To create a binary outcome, we split the data by the mean of the outcome $Y$ thus artificially generating an ``underweight'' population of $29$ mice and an ``overweight'' population of $28$ mice.
Again, we show the influential log-ratios for the naive regression Only LC and \ilrlc{} (see \cref{appfig:agglevel_binary}). 
While for \ilrlc{} the influential log-ratios stay the same for binary and continuous outcome, for the naive regression they are not entirely consistent.

\newpage

\section{Data Generation}
\label{supp:data_generation}

This section describes the details of how we generate data for our empirical evaluation.
Complementary to the real microbiome data, we consider several approaches to generate data for the compositional instrumental variable setting.
Since counterfactuals are never observed in practice, we need a setup where the ground truth is known and can be controlled.
We choose to simulate data from two different data generating models, Setting A and Setting B.
The first one will put (most of) our models in a wellspecified setting, where we have strong 
expectations and theoretical guarantees on how they will behave.
The other approach simulates compositional data by a zero-inflated negative binomial.
Thus, the first stage of all of our models will be misspecified (except for potentially KIV assuming a proper choice of the kernels).
This allows us to test our models for robustness and probe their limitations.

Based on this motivation, we also describe two additional parameter settings within Setting A that will examine robustness and limitations: a weak instrument scenario and a scenario with a non-linear second stage $f$. The first scenario will test the necessity of a strong/valid instrument, the second scenario will further look into the issue of misspecification (now in the second stage).

We describe the data generating model and the specific parameter settings.
We also provide visualizations of the resulting data distributions, which is rather tricky for compositional data with $p>3$.
We will then supplement the result section of the main text with additional comments on the evaluation of the results and show the complete set of plots for \cref{tab:res_all_settinga,tab:res_all_settingb}.

Each generated dataset for $p=3$ comprises $n=1000$ samples, resp. $10,\!000$ samples for $p=30$ and $p=250$, with an additional $n_{\text{intervention}}=250$ interventional samples for evaluation of OOS MSE. 
Note that the examples in the figures show only one of these datasets. 
To ensure reproducibility, we consistently chose the \nth{10} dataset of the confidence runs for (a representative) visualization. 

\subsection*{Setting A}
The following explanations refer to \emph{Setting A} described 
in the main part. 

Setting A generates data that enables us to assess our methods in a \emph{wellspecified} setting. 
Instead of modeling $X \in \simp^{p-1}$ directly, we model $\ilr(X)$.
The setting is strictly linear in $\ilr(X)$. This means that both $g$ and $f$ are linear functions of $U$ and $Z$, resp., $U$ and $\ilr(X)$. The generative model is as follows:
\begin{align}
    Z_j &\sim\text{Uniform}(0, 1) \nonumber\\ 
    U &\sim \cN(\mu_c, 1) \nonumber\\ 
    \ilr(X) = g(Z, U) &= \alpha_0 + \alpha Z + c_X U \nonumber\\
    Y =f(X, U) &= \beta_0 + \beta^T \ilr(X)  + c_Y U \label{eq:lineary}
\end{align}

\subsubsection*{Setting A with $p=3, q=2$}
The main characteristics of this lower dimensional dataset are the presence of all microbes and relatively seldom zero values. 
We choose the following parameters for the low-dimensional case:
\begin{equation}
\mu_c = -3, 
    \alpha_0 = [1, 1], 
    \alpha = \begin{bmatrix} 0.5 & -0.15 \\ 0.3 & 0.7 \end{bmatrix},
    c_X = [0.5, 0.5], 
    \beta_0 = 0.5, 
    \beta = [4, 1], 
    c_Y = 4    
\end{equation}
The first stage F-test for the two components of $\ilr(X)$ gives $(32.18 , 113.99)$ for the \nth{10} data sample.

We remark that in higher dimensions, the F-test does not provide a strong theoretical justification for sufficient instrument strength, but we still use it as a sensible heuristic that provides a relative measure between different settings, i.e., in which scenario the instrument is stronger.

\begin{figure}
  \centering
  \includegraphics[width=.45\textwidth]{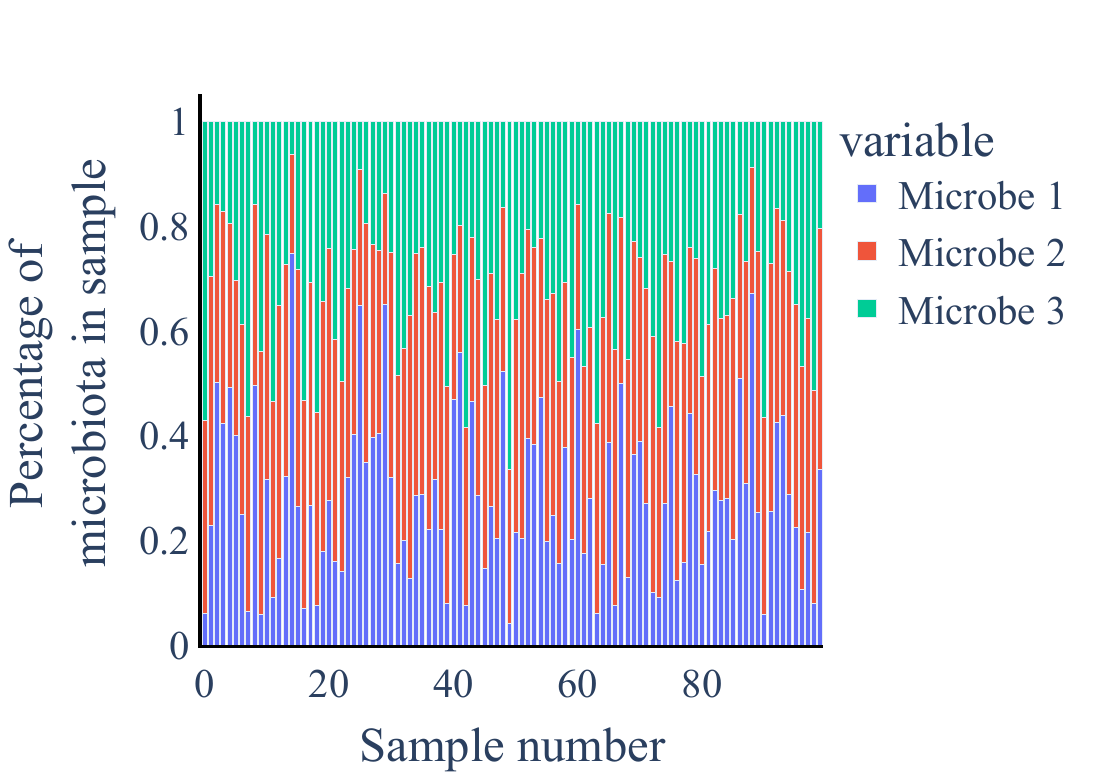}
  \caption{\textbf{Setting A with $p=3$, $q=2$: }
  The barplot shows the three-part composition of the first $100$ samples. The microbes are evenly distributed over the individual compositions.}
\label{appfig:linear_p3_barplot}
\end{figure}

\begin{figure}
  \centering
  \includegraphics[width=.45\textwidth]{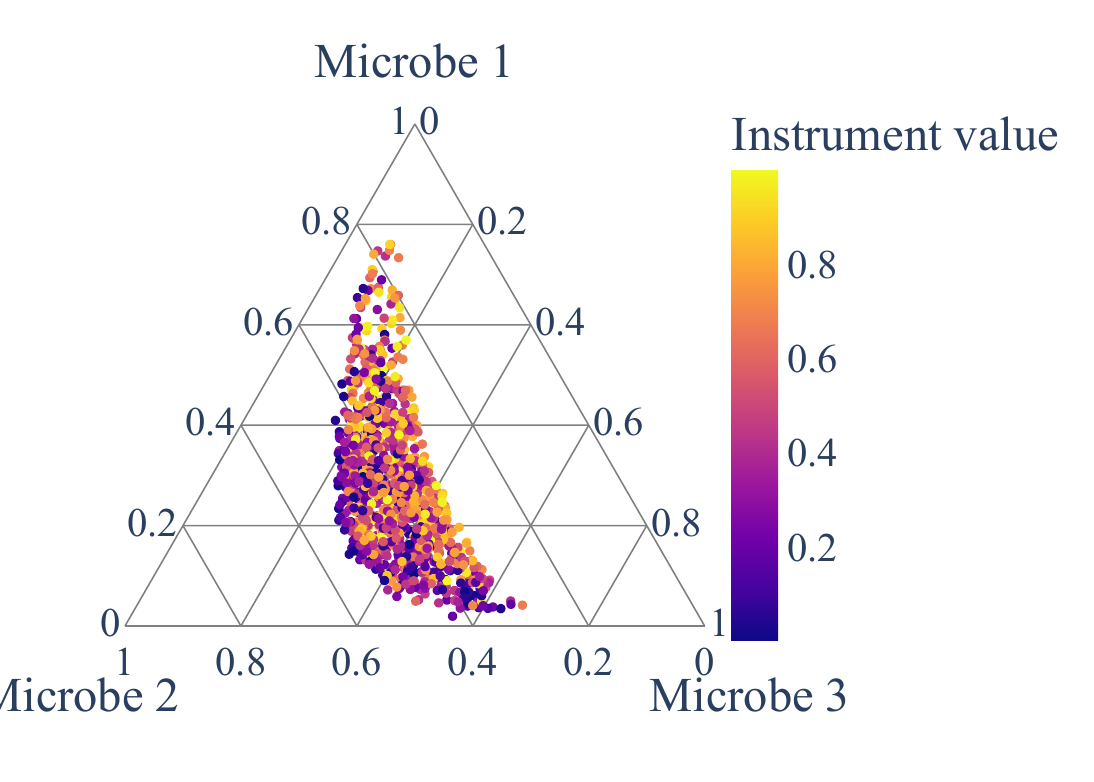}
  \hfill
  \includegraphics[width=.45\textwidth]{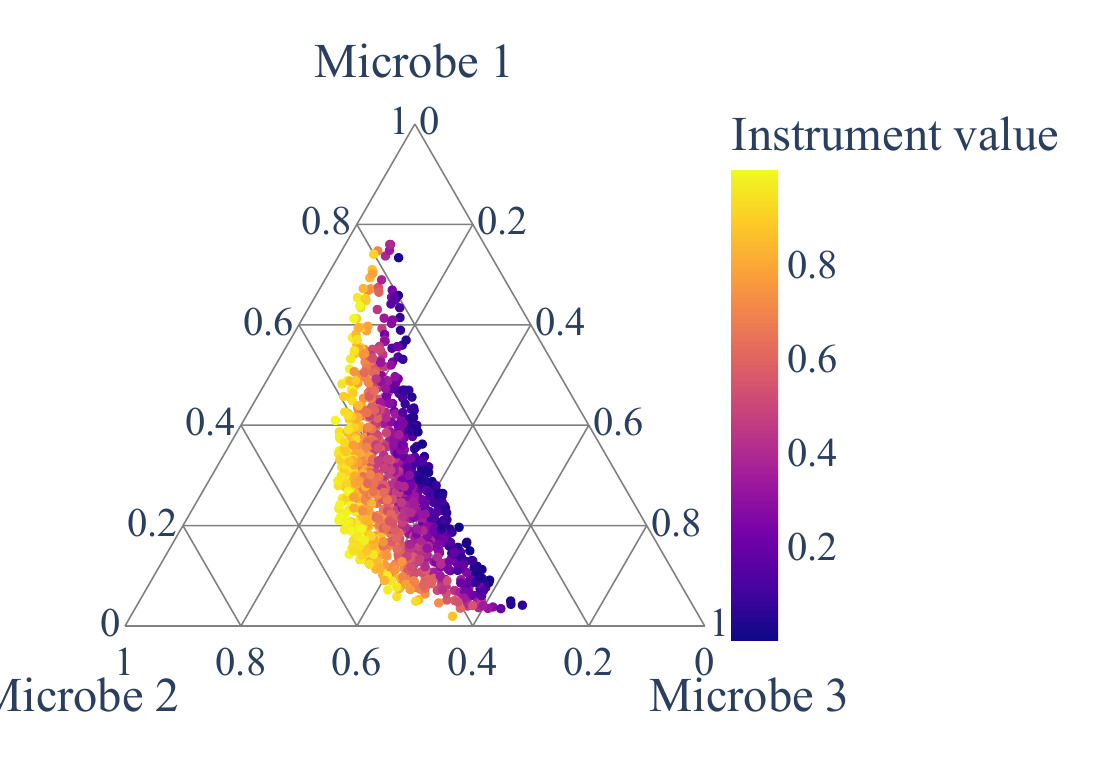}
  \caption{\textbf{Setting A with $p=3$, $q=2$:} The ternary plots are colored by first (left) and second (right) instrument value. The influence of $Z_2$ on the composition $X$ is particularly pronounced and visually supports the assumption of $Z$ being a valid instrument.}
\label{appfig:linear_p3_ternary}
\end{figure}

\begin{figure}
  \centering
  \includegraphics[width=1\textwidth]{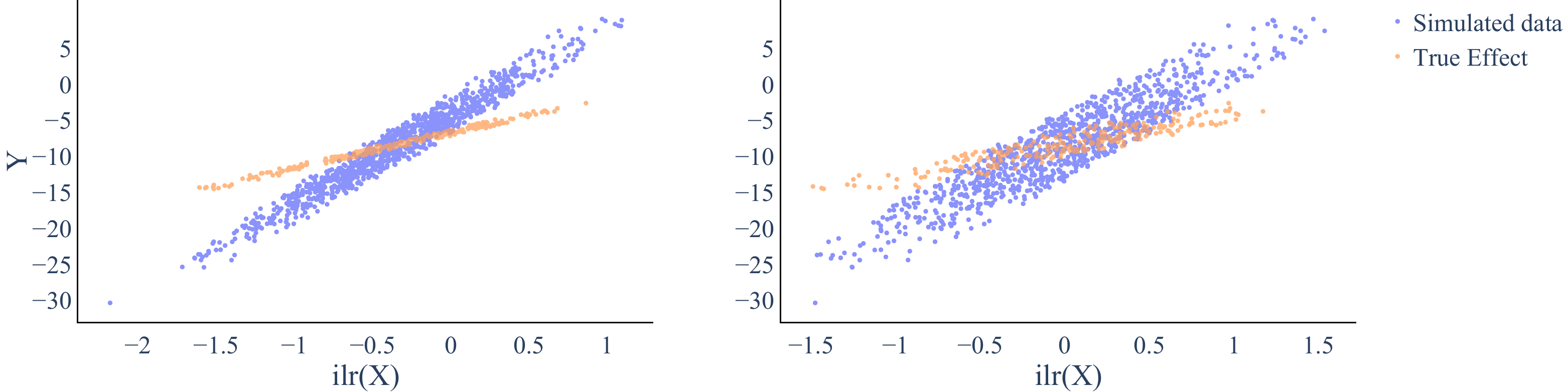}
  \caption{\textbf{Setting A with $p=3$, $q=2$:} Both plots show one component of $\ilr(X) \in \bR^2$ vs.\ the confounded outcome (blue) and the true effect (orange). Due to the confounding, the observed and the causal effect do not overlap. However, we expect the instrument $Z$ to factor out the confounding effect and enable the two-stage methods to identify the causal effect.}
\label{appfig:linear_p3_xy}
\end{figure}

For the $p=3$ case, we can visualize $X$ by its compositional coordinates not only in a barplot (\cref{appfig:linear_p3_barplot}) but also in an arguably more informative ternary plot (\cref{appfig:linear_p3_ternary}). 
To visualize the linear relationship between observed $\ilr(X)$ and $Y$ as well as the true effect $Y \mid do(X)$, we transform the data $X$ and visualize each component in a separate scatter plot (see \cref{appfig:linear_p3_xy}).

\subsubsection*{Setting A with $p=30, q=10$}
Contrary to the previous example, we now analyze a slightly higher-dimensional setting with $p=30$.
In this scenario, it makes sense to introduce sparsity in the data generation process from a practical viewpoint.
We work with the data generation setting given in \cref{eq:lineary} and choose the following parameters: 
\begin{align*}
\mu_c = 5,\,
\alpha_0 = [3, 1, 1, 1, 3, 1, 1, 1 , 0, \cdots , 0],\;
&\alpha_{ij} \begin{cases} 0, &\text{for }i \not = j \text{ and } i, j > 8, \\ 1,  &\text{for }i \not =  j \leq 8 \end{cases}, \\
c_X = [-2 , -1 , -1 , -1 , 2, 1, 1, 1, 0, \cdots , 0] ,\,
\beta_0 = 5 ,\;
&\beta_{\log} =  [10, 5, 5, 5, -10, -5, -5, -5 , 0, \cdots , 0] ,\,
\beta = V^T \cdot \beta_{\log},\,
c_Y= 5
\end{align*}
for $i \in \{1, \ldots, p-1\}, j \in \{1, \ldots, q\}$ and $V$ providing the orthonormal basis for the $\ilr$-transformation (see \cref{supp:compositional_transformations}).

\begin{figure}
  \centering
  \includegraphics[width=.45\textwidth]{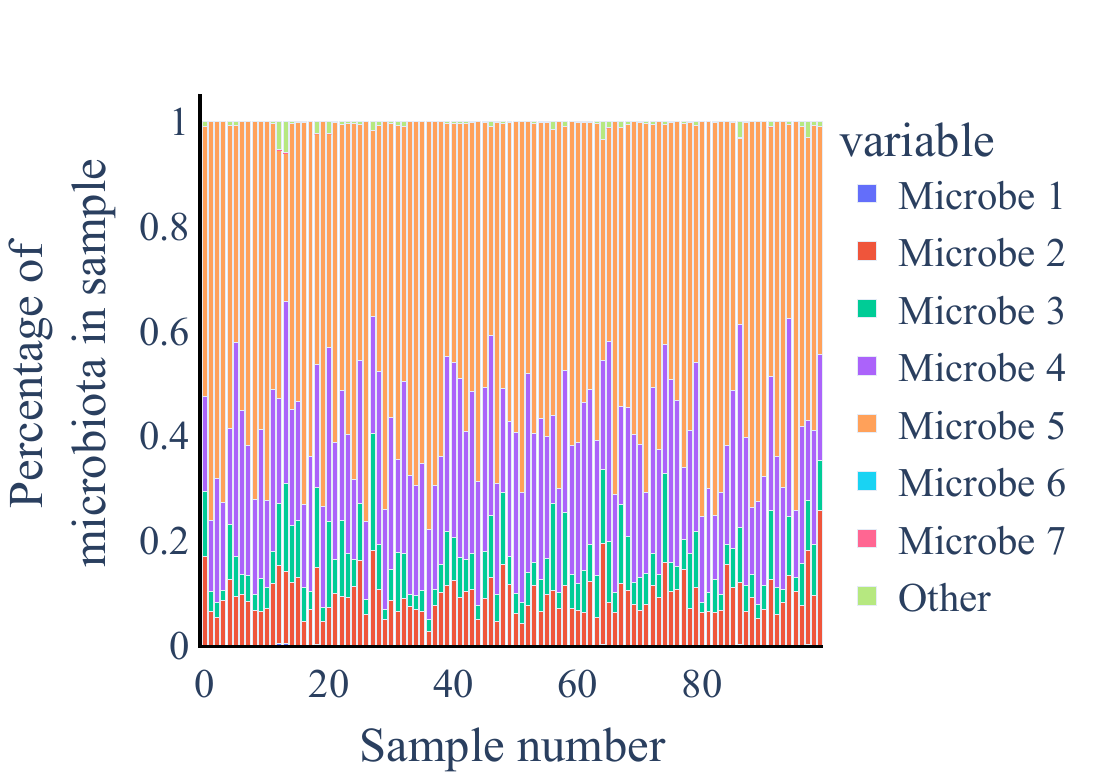}
  \caption{\textbf{Setting A with $p=30$, $q=10$: }
  The barplot shows the composition of the first $100$ samples. The compositions are dominated by a few species.}
\label{appfig:linear_p30_barplot}
\end{figure}

\begin{figure}
  \centering
  \includegraphics[width=1\textwidth]{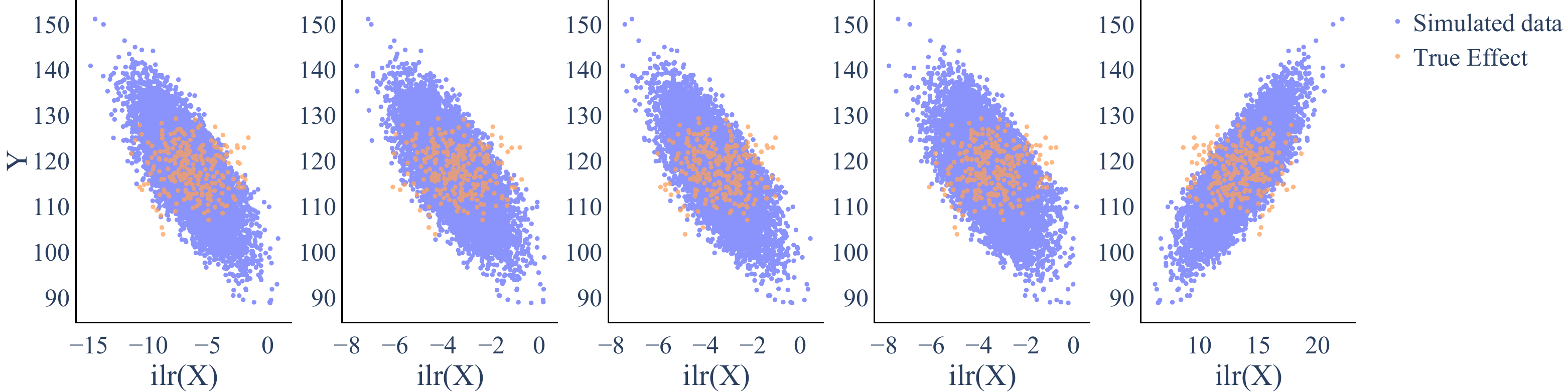}
  \caption{\textbf{Setting A with $p=30$, $q=10$ :} Both plots show one component of $\ilr(X) \in \bR^{29}$ vs.\ the confounded outcome (blue) and the true effect (orange). Due to the confounding, the observed and the causal effect do not overlap. However, we expect the instrument $Z$ to factor out the confounding effect and enable the two-stage methods to identify the causal effect.}
\label{appfig:linear_p30_xy}
\end{figure}

Since a visualization with a ternary plot is no longer feasible, we only show barplots of the data in \cref{appfig:linear_p30_barplot}.
However, scatter plots showing individual $\ilr(X)$ coordinates versus the observed $Y$ and the true causal effect are still informative.
Since the first components are the most influential ones in our setting, we show the first five $\ilr(X)$ components in \cref{appfig:linear_p30_xy}.


\subsubsection*{Setting A with $p=250, q=10$}
\begin{figure}
  \centering
  \includegraphics[width=.45\textwidth]{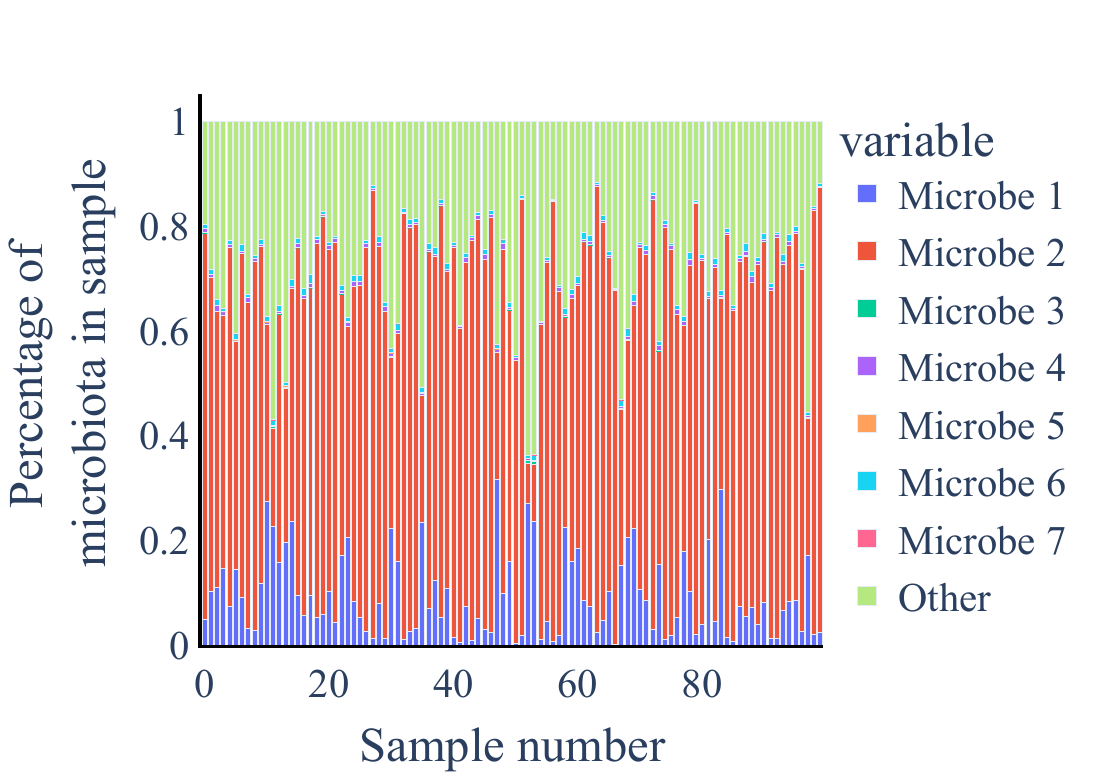}
  \caption{\textbf{Setting A with $p=250$, $q=10$: }
  The barplot shows the composition of the first $100$ samples. The compositions are dominated by a few species.}
\label{appfig:linear_p250_barplot}
\end{figure}

\begin{figure}
  \centering
  \includegraphics[width=1\textwidth]{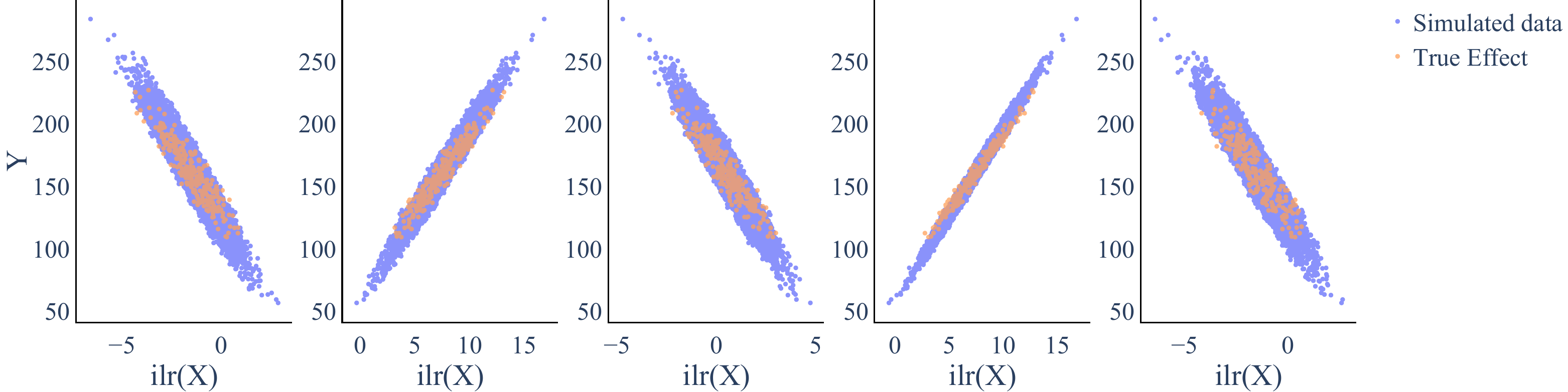}
  \caption{\textbf{Setting A with $p=250$, $q=10$:} Both plots show one component of $\ilr(X) \in \bR^{249}$ vs.\ the confounded outcome (blue) and the true effect (orange). Due to the confounding, the observed and the causal effect do not overlap. However, we expect the instrument $Z$ to factor out the confounding effect and enable the two-stage methods to identify the causal effect.}
\label{appfig:linear_p250_xy}
\end{figure}

We now analyze the second high-dimensional setting with $p=250$.
As in the scenario of $p=30$, it makes sense to introduce sparsity in the data generation process from a practical viewpoint.
We work with the data generation setting given in \cref{eq:lineary} and choose the following parameters:
\begin{align*}
\mu_c = 3,\,
\alpha_0 = [1, 1, 3, 1, 1, 1, 3, 1, 1, 1, 3, 1 , 0, \cdots , 0],\;
&\alpha_{ij} \begin{cases} 0, &\text{for }i \not = j \text{ and } i, j > 8, \\ 1,  &\text{for }i \not =  j \leq 8 \end{cases}, \\
c_X = [-1, 2, -1, 2, -1, 2, -2, 1, -2, 1, -2, 1, 0, \cdots , 0] ,\,
\beta_0 = 5 ,\;
&\beta_{\log} =  [10, 5, 5, 5, -10, -5, -5, -5 , 0, \cdots , 0] , \\
\beta = V^T \cdot \beta_{\log},\,
c_Y= 5
\end{align*}
for $i \in \{1, \ldots, p-1\}, j \in \{1, \ldots, q\}$ and $V$ providing the orthonormal basis for the $\ilr$-transformation (see \cref{supp:compositional_transformations}).
Since a visualization with a ternary plot is no longer feasible, we only show barplots of the data in \cref{appfig:linear_p250_barplot}.
However, scatter plots showing individual $\ilr(X)$ coordinates versus the observed $Y$ and the true causal effect are still informative.
Since the first components are the most influential ones in our setting, we show the first five $\ilr(X)$ components in \cref{appfig:linear_p250_xy}.



\subsection*{Setting B}
The following explanations refer to \emph{Setting B} described 
in the main part. 

Setting B serves three main purposes: (i) to assess our methods on a dataset that closely resembles real-world data in terms of its distribution, (ii) to assess our methods when the first stage is misspecified, and (iii) to allow for sparsity in the first stage of the data generating process, resembling the real data in 
\cite{Schulfer2019}. 
The sparsity of the compositional data can be accomplished by a zero-inflated negative binomial distribution. As ZINegBinomial is a frequently used distribution in modeling microbiome data, we assume a closer resemblance to real world sparsity than the resemblance we achieve in Setting A for $p=30$ and $p=250$.

The data is generated according to the following model with the parameter $\mu$ of the negative binomial as $\mu = \alpha_0 + \alpha Z$:
\begin{align}
    Z_j &\sim \text{Uniform}({Z_{\min}}, {Z_{\max}} ), \nonumber  \\ 
    U &\sim \text{Uniform}({U_{\min}}, {U_{\max}}), \nonumber  \\ 
    X =g(Z, U) &\sim C(\text{ZINegBinomial}(\mu, \Sigma, \theta, \eta )) \oplus (\Omega_C \odot U ),\nonumber  \\
     Y = f(X, U) &= \beta_0 + \beta^T \log(X) + c_Y^T \log(\Omega_C \odot U ) \label{eq:negbinom}
\end{align}
We fix $Z_{\min} = 1, Z_{\max} = 10$ and $U_{\min} = 0.2, U_{\max} = 3$ throughout. 
For the negative binomial distribution we set $\Sigma = \mathbb{I}_{p}$, i.e., assuming no additional correlation within the different components of the composition for simplicity.

\subsubsection*{Setting B with $p=3, q=2$}
The parameter setting with $p=3$ does not yet contain sparse data due to its low-dimensionality. It serves the purpose to compare the performance of the two-stage methods in a misspecified setting and a wellspecified setting (except for DIR+LC which is misspecified in both Setting A and Setting B).

Here, we consider the following generative model based on \cref{eq:negbinom}.
We fix $Z_{\min} = 0, Z_{\max} = 10, U_{\min} = 0.2, U_{\max} =  3$. 
We chose $\alpha_0$ to be $[7, 9, 8]$ and $\alpha = \begin{bmatrix} 5 & 0 & 0 \\ 0 & 5 & 0 \\ 0 & 0 & 5 \end{bmatrix}$ to guarantee for valid instruments.
We set the dispersion to $\theta=2$ and keep the zero probability at $\eta =  [0, 0, 0]$ to get valid compositions for this low-dimensional scenario.
For the confounding composition $\Omega_C$, we set it to $[0.7, 0.1, 0.2]$.
For the second stage, we fix ground truth parameters $\beta_0 = 1$, $\beta_{\log} = [-5, 3, 2]$, which results in $\beta = V^T \beta_{\log}$ and the confounding parameter $c_Y =  [2, -10, -10]$.

The first stage F-test for the two components of $\ilr(X)$ gives $(41.38 , 14.08)$ for the \nth{10} data sample. We remark that in higher dimensions, the F-test does not provide a strong theoretical justification for sufficient instrument strength, but we still use it as a sensible heuristic that provides a relative measure between different settings, i.e., in which scenario the instrument is stronger.

\begin{figure}
  \centering
  \includegraphics[width=.45\textwidth]{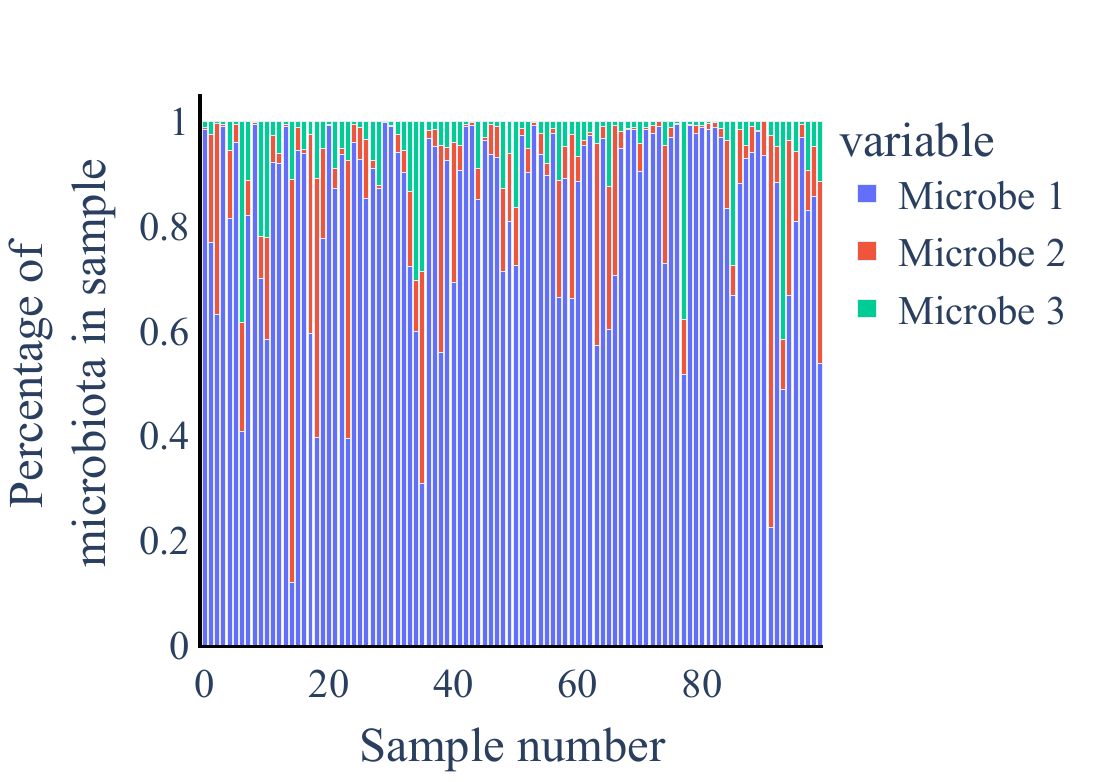}
  \caption{\textbf{Setting B with $p=3$, $q=2$: }
  The barplot shows the three-part composition of the first $100$ samples. The data sample shows some dominating species in the individual compositions while having more variation between the samples compared to Setting A.}
\label{appfig:negbinom_p3_barplot}
\end{figure}

\begin{figure}
  \centering
  \includegraphics[width=.45\textwidth]{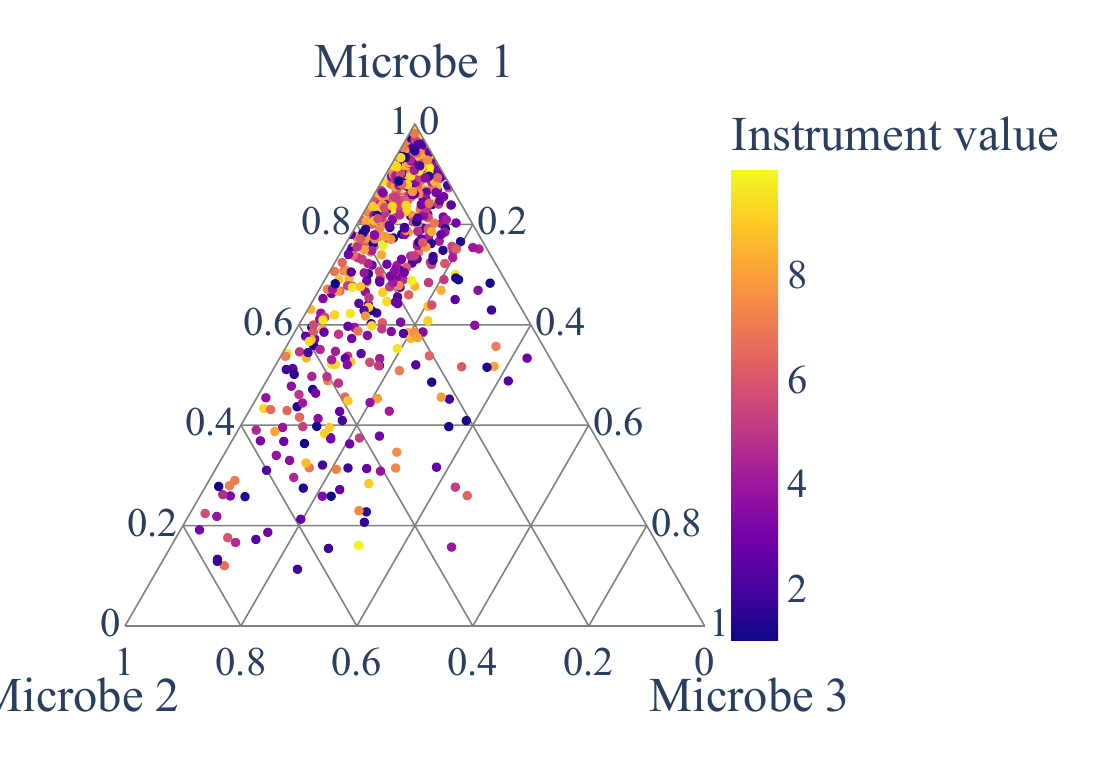}
  \hfill
  \includegraphics[width=.45\textwidth]{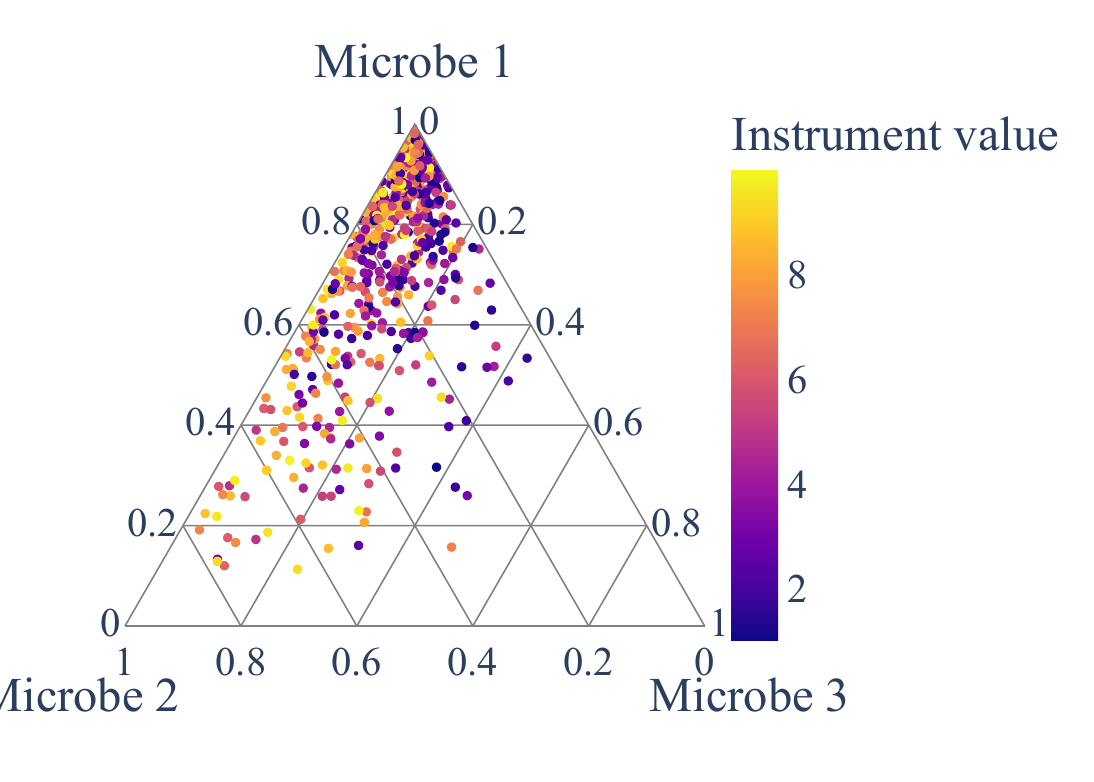}
  \caption{\textbf{Setting B with $p=3$, $q=2$:} The ternary plots are colored by first (left) and second (right) instrument value. Due to the data generation process, the influence of $Z_1$ and $Z_2$ on the composition $X$ is less visually obvious than for Setting A. Nevertheless, $Z$ can be assumed to be a valid instrument.}
\label{appfig:negbinom_p3_ternary}
\end{figure}

\begin{figure}
  \centering
  \includegraphics[width=1\textwidth]{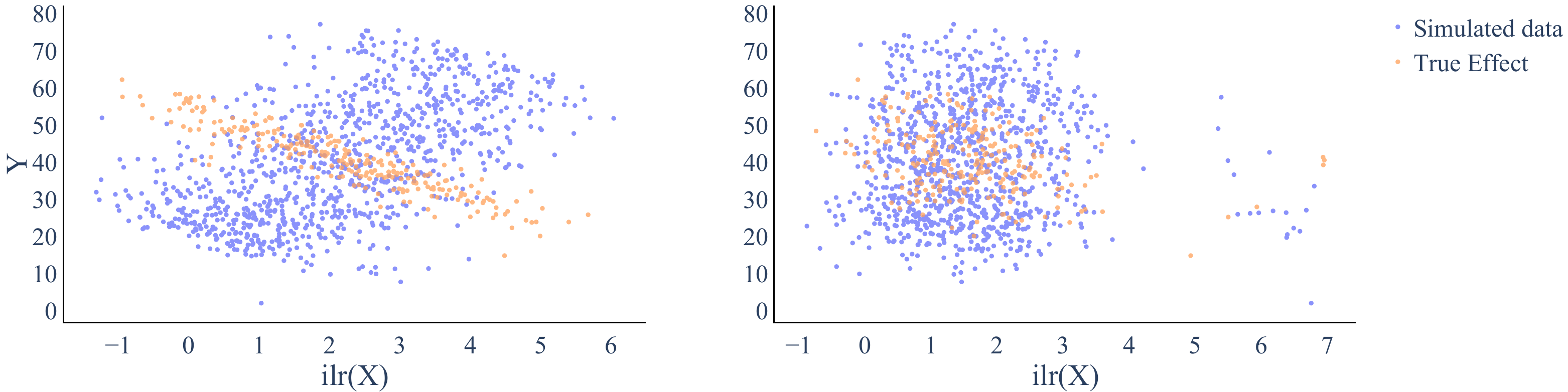}
  \caption{\textbf{Setting B with $p=3$, $q=2$:} Both plots show one component of $\ilr(X) \in \bR^2$ vs.\ the confounded outcome (blue) and the true effect (orange). Due to the confounding, the observed and the causal effect do not overlap. However, we expect the instrument $Z$ to factor out the confounding effect and enable the two-stage methods to identify the causal effect.}
\label{appfig:negbinom_p3_xy}
\end{figure}

For the $p=3$ case, we can visualize $X$ by its compositional coordinates not only in a barplot (\cref{appfig:negbinom_p3_barplot}) but also in an arguably more informative ternary plot (\cref{appfig:negbinom_p3_ternary}). 
To visualize the relationship between observed $\ilr(X)$ and $Y$ as well as the true effect $Y \mid do(X)$, we transform the data $X$ and visualize each component in a separate scatter plot (see \cref{appfig:negbinom_p3_xy}).

\subsubsection*{Setting B with $p=30, q=10$}
In the higher-dimensional scenarios we will make us of the sparsity ability of the ZINegBinom distribution.

The parameters were chosen to generate a suitable dataset that still conveys typical compositional data properties (sparsity, high variance within the composition, similar means to real data) and  significant instruments. Here, we consider the following generative model based on \cref{eq:negbinom}.
We fix $Z_{\min} = 0, Z_{\max} = 10, U_{\min} = 0.2, U_{\max} =  3$. 
To ensure a handful of components dominating the composition, we fix the first $8$ entries of $\alpha_0$ to be $[1, 1, 2, 1, 4, 4, 2, 1, 4, 4, 2, 1]$ and randomly sample the remaining ones from $\text{UniformChoice}([1, 2, 2])$. 
For $\alpha$, which mainly controls the instrument strength, we use a deterministic value to guarantee valid instruments:
\begin{equation}
    \alpha_{ij} \begin{cases} 0, &\text{for }i \not = j \text{ and } i, j > 8, \\ 1,  &\text{for }i \not =  j \leq 8 \end{cases}
\end{equation} 
We set the dispersion to $\theta=2$ and the zero probability value $\eta =  [0, \dots, 0, 0.8, \dots, 0.8]$.
For the confounding composition $\Omega_C$, we fix the first components to $[0.2, 0.3, 0.2, 0.1]$, to ensure that the most dominating parts of the composition are also more strongly influenced by confounding. 
Then we sample the remaining components of $\Omega_C$ from $\text{UniformChoice}([0.01, 0.05])$ and eventually apply the closure operator $C$ to ensure $\Omega_C$ is a composition.
For the second stage, we fix ground truth parameters $\beta_0 = 1$, $\beta_{\log} = [-10, -5, -5, -5, 10, 5, 5, 5, 0, \dots, 0]$, which results in $\beta = V^T \beta_{\log}$ and the confounding parameter $c_Y =  [10, 10, 5, 15, -5, -5, -5, -5, -5, -5, -5, -5, 0, \dots, 0]$.

\begin{figure}
  \centering
  \includegraphics[width=.45\textwidth]{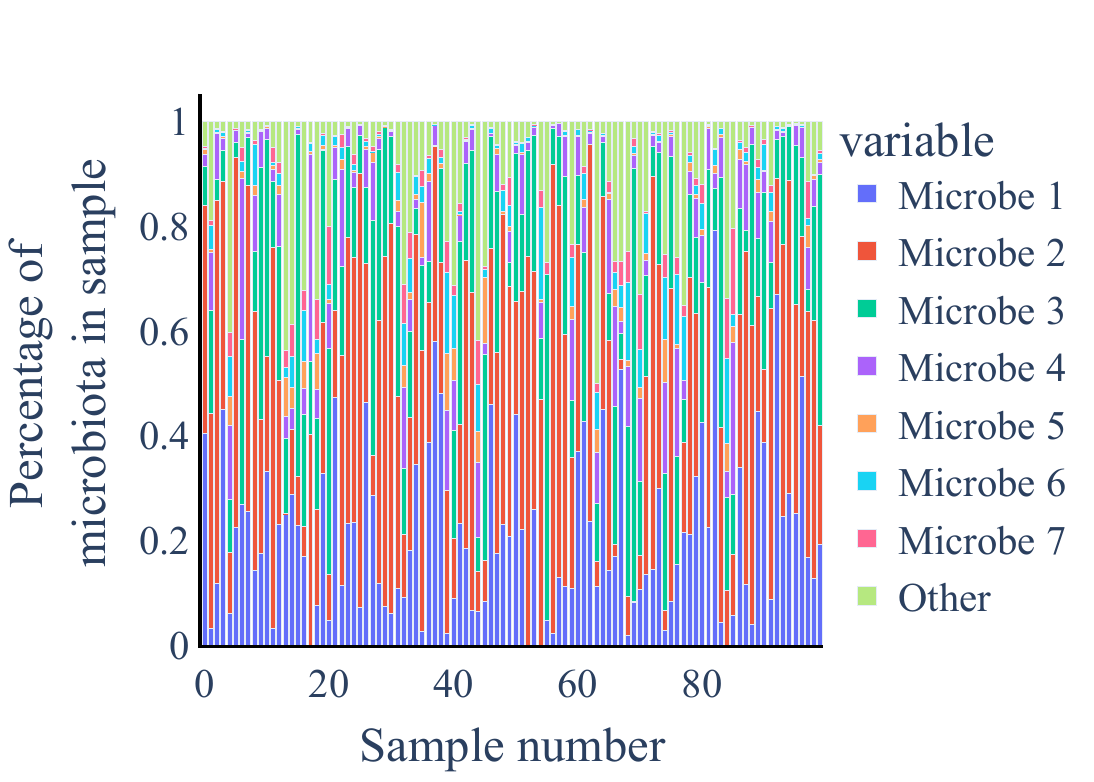}
  \caption{\textbf{Setting B with $p=30$, $q=10$:} The barplot shows the different compositions of the first $100$ samples in the dataset. We observe some dominating components and many small components with an overall high variability.}
\label{appfig:negbinom_p30_barplot}
\end{figure}

\begin{figure}
  \centering
  \includegraphics[width=1\textwidth]{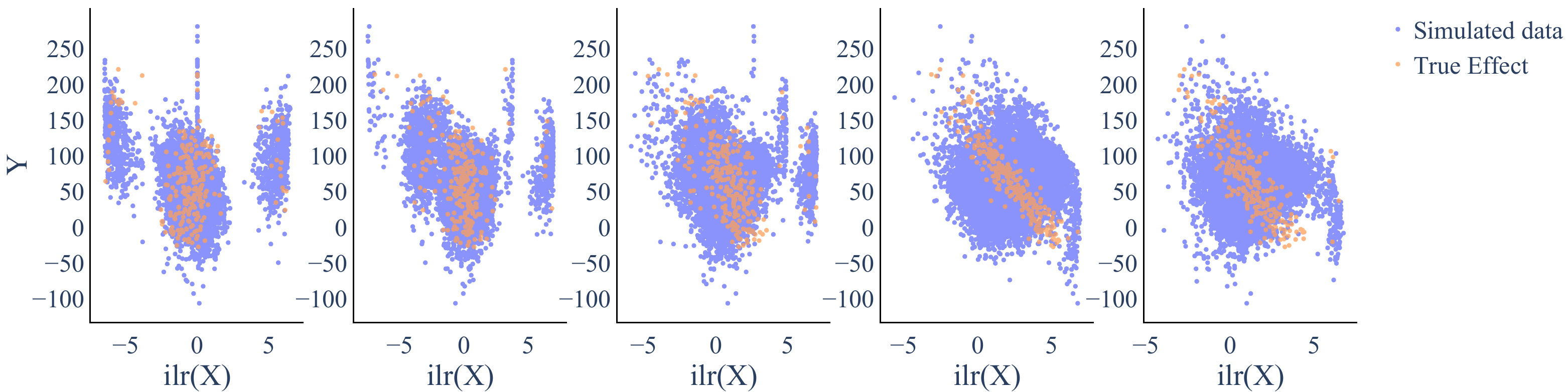}
  \caption{\textbf{Setting B with $p=30$, $q=10$: } Each plot shows one of the first five components of $\ilr(X) \in \bR^{29}$ vs.\ the confounded outcome (blue) and the true effect (orange). The dataset shows strong confounding in some of the components as the true effect and the observed effect actually contradict each other. We expect the two-stage methods to perform better than the naive regression in such scenarios. We can thus check if the two-stage methods are still able to make use of the instrument $Z$ despite the misspecified first stage.}
\label{appfig:negbinom_p30_xy}
\end{figure}

For a brief overview, we visualize the first five components of the $\ilr(X)$ coordinates versus the observed $Y$ and the true causal effect in \cref{appfig:negbinom_p30_xy} and show barplots of the generated data in \cref{appfig:negbinom_p30_barplot}.

\subsubsection*{Setting B with $p=250, q=10$}

We consider now the second high-dimensional scenario for Setting B with $p=250$.
The parameters for Setting B with $p=250$ are very close to the parameters for Setting B with $p=30$.

Again, we consider the following generative model based on \cref{eq:negbinom}.
We fix $Z_{\min} = 0, Z_{\max} = 10, U_{\min} = 0.2, U_{\max} =  3$. 
To ensure a handful of components dominating the composition, we fix the first $8$ entries of $\alpha_0$ to be $[1, 1, 2, 1, 4, 4, 2, 1, 4, 4, 2, 1]$ and randomly sample the remaining ones from $\text{UniformChoice}([1, 2, 2])$. 
For $\alpha$, which mainly controls the instrument strength, we use a deterministic value to guarantee valid instruments:
\begin{equation}
    \alpha_{ij} \begin{cases} 0, &\text{for }i \not = j \text{ and } i, j > 8, \\ 1,  &\text{for }i \not =  j \leq 8 \end{cases}
\end{equation} 
We set the dispersion to $\theta=2$ and the zero probability value $\eta =  [0, \dots, 0, 0.8, \dots, 0.8]$.
For the confounding composition $\Omega_C$, we fix the first components to $[0.2, 0.3, 0.2, 0.1]$, to ensure that the most dominating parts of the composition are also more strongly influenced by confounding. 
Then we sample the remaining components of $\Omega_C$ from $\text{UniformChoice}([0.01, 0.05])$ and eventually apply the closure operator $C$ to ensure $\Omega_C$ is a composition.
For the second stage, we fix ground truth parameters $\beta_0 = 1$, $\beta_{\log} = [-10, -5, -5, -5, 10, 5, 5, 5, 0, \dots, 0]$, which results in $\beta = V^T \beta_{\log}$ and the confounding parameter $c_Y =  [10, 10, 5, 15, -5, -5, -5, -5, -5, -5, -5, -5, 0, \dots, 0]$.

\begin{figure}
  \centering
  \includegraphics[width=.45\textwidth]{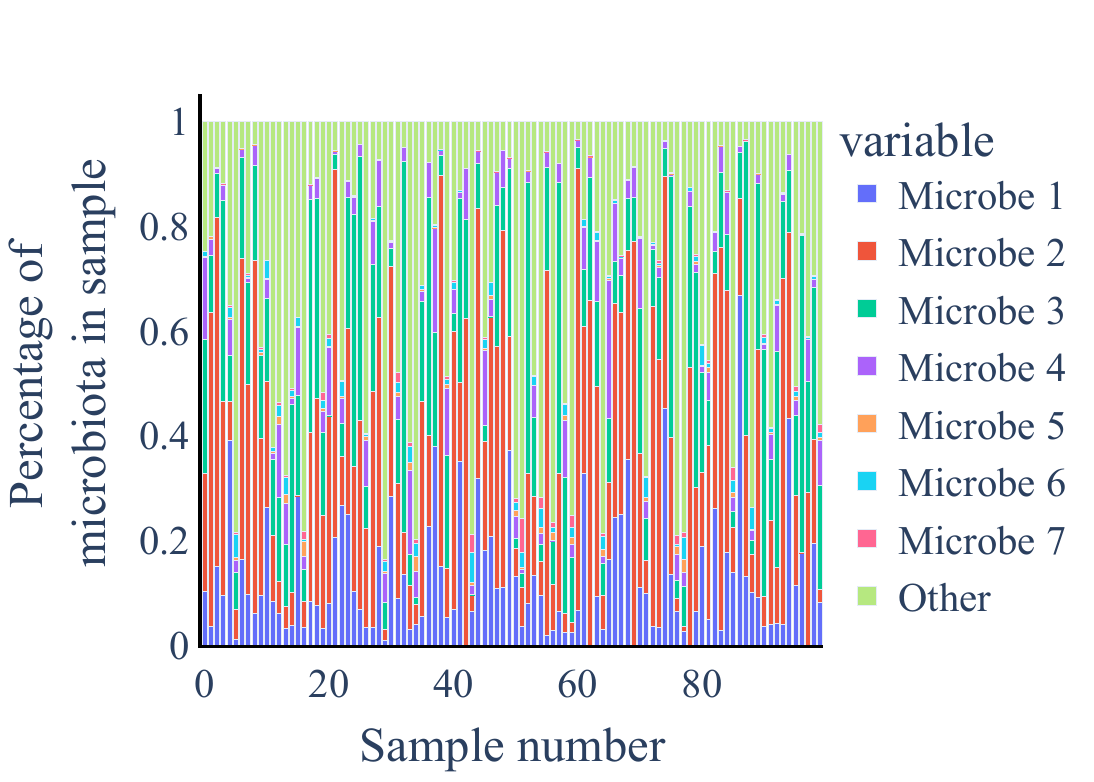}
  \caption{\textbf{Setting B with $p=250$, $q=10$:} The barplot shows the different compositions of the first $100$ samples in the dataset. We still observe a few dominating components and many small components with an overall high variability.}
\label{appfig:negbinom_p250_barplot}
\end{figure}

\begin{figure}
  \centering
  \includegraphics[width=1\textwidth]{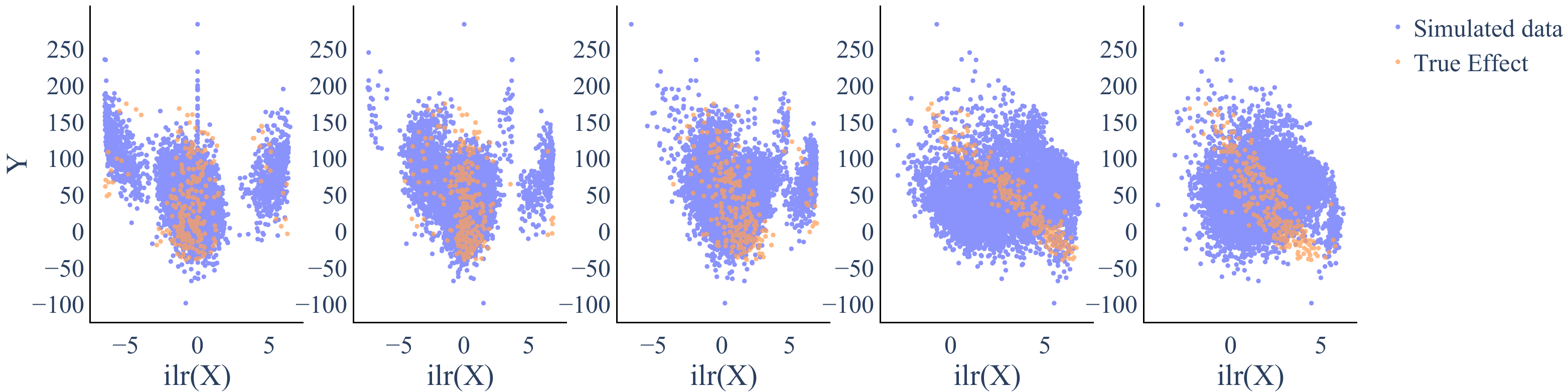}
  \caption{\textbf{Setting B with $p=250$, $q=10$: } Each plot shows one of the first five components of $\ilr(X) \in \bR^{249}$ vs.\ the confounded outcome (blue) and the true effect (orange). The dataset shows strong confounding in some of the components as the true effect and the observed effect actually contradict each other. We expect the two-stage methods to perform better than the naive regression in such scenarios. We can thus check if the two-stage methods are still able to make use of the instrument $Z$ despite the misspecified first stage.}
\label{appfig:negbinom_p250_xy}
\end{figure}

For a brief overview, we visualize the first five components of the $\ilr(X)$ coordinates versus the observed $Y$ and the true causal effect in \cref{appfig:negbinom_p250_xy} and show barplots of the generated data in \cref{appfig:negbinom_p250_barplot}.

\subsection*{Further Settings for Robustness Evaluation}

By assuming a misspecified first stage in Setting B via the ZINegBinom distribution, we already started to evaluate the robustness of our methods. Nevertheless, we will further relax different requirements within Setting A.
We evaluate the robustness via two additional scenarios
\begin{enumerate}
    \item We relax the assumption of a valid instrument and test the sensitivity of the methods with respect to weak instruments.
    \item We assume a non-linear ground truth relationship $f$ for the second stage, a scenario for which all the considered models are misspecified.
\end{enumerate}

\subsubsection*{Weak Instrument}
\label{supp:instrumentstrength}

``Strong instruments'' resp. ``valid instruments'' are a prerequisite for successful two-stage estimation and one of the key discussion points in applications of two-stage instrumental variable estimation.
Instrument strength for $p=1$ is typically measured via the first-stage F-statistic with a value $>10$ being considered sufficient to avoid weak instrument bias in 2SLS \citep{andrews2019weak}.
For $p>1$, measuring instrument strength is not as straightforward \citep{sanderson20162ftest} and we thus report F-statistics for each dimension of the treatment (either $X \in \simp^{p-1}$ or $\ilr(X) \in \bR^{p-1}$) separately.
Theoretically, the estimation bias can become arbitrarily large (even in the large data limit) for weak instruments.
To quantitatively assess the effect of weak instruments in our specific applications, we provide an additional simulation scenario and its results (see \cref{supp:method_results}) for a weak instrument settings.

\paragraph{Setting A with $p=3, q=2$ and weak instruments}

\begin{figure}
  \centering
  \includegraphics[width=.45\textwidth]{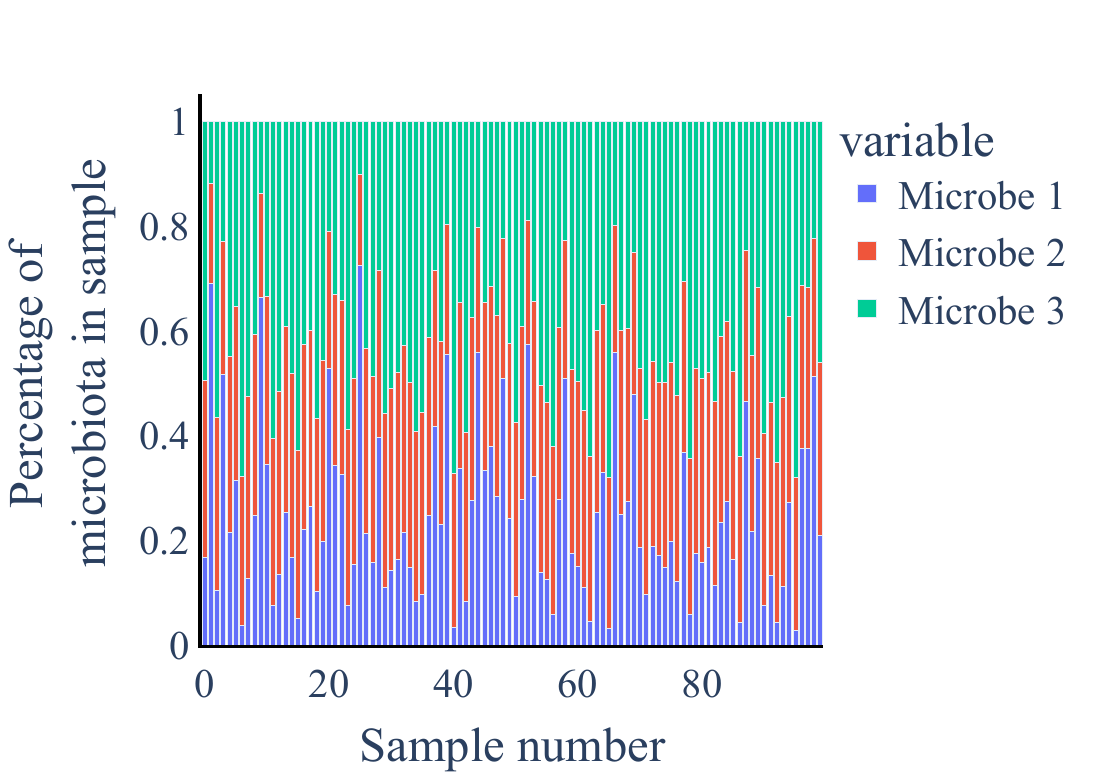}
  \caption{\textbf{Setting A with $q=2$, $p=3$ and weak instruments:} The barplot shows the different composition in each sample (plotted here for the first $100$ samples). Microbe 2 has a relatively small value whereas microbe 1 and microbe 3 dominate the composition by high variation.}
\label{appfig:linear_p3_barplot_weak}
\end{figure}

\begin{figure}
  \centering
  \includegraphics[width=.45\textwidth]{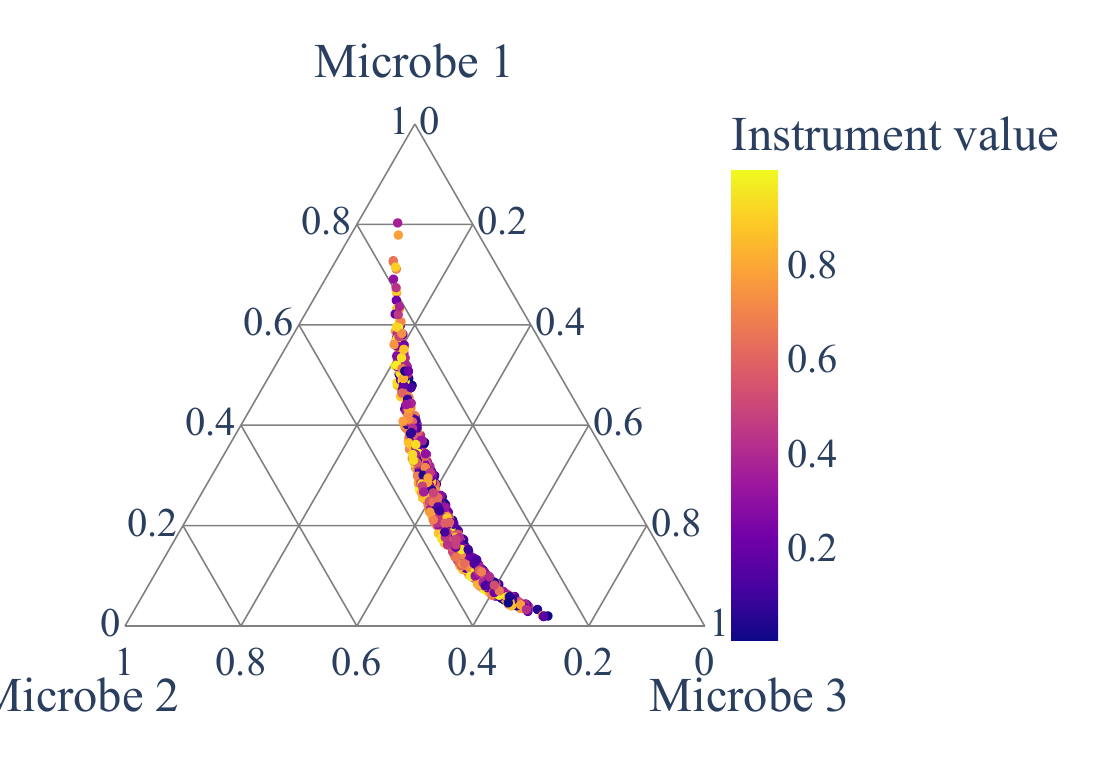}
  \hfill
  \includegraphics[width=.45\textwidth]{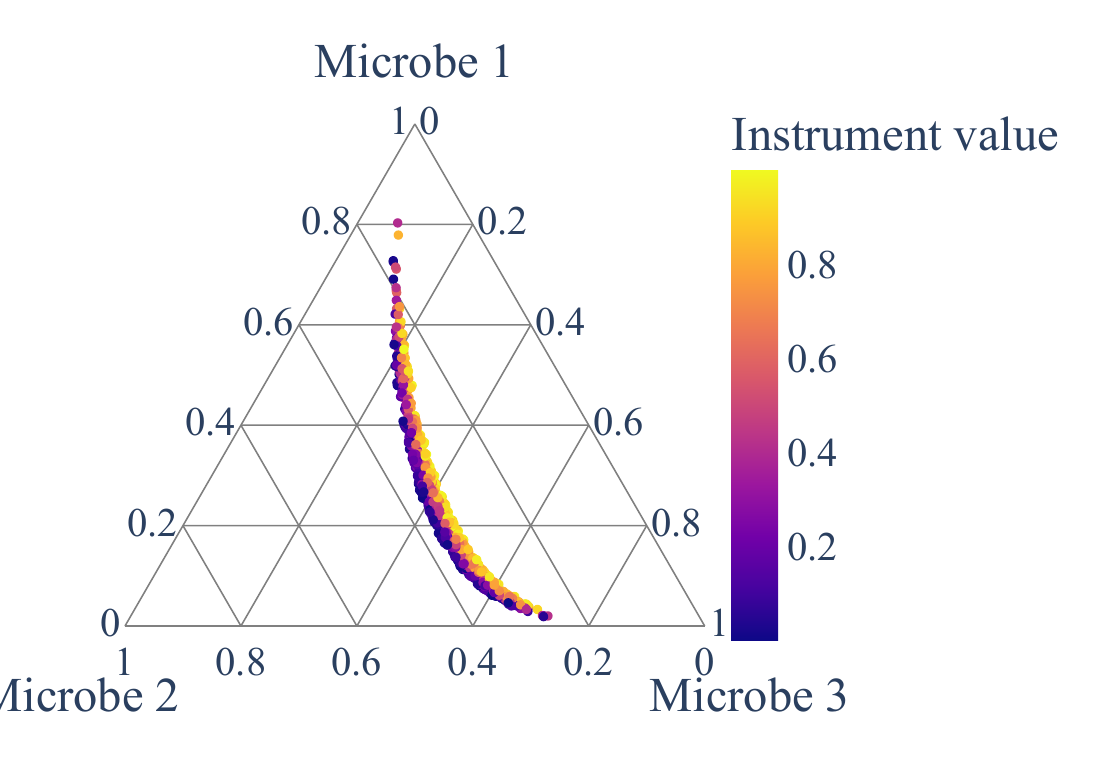}
  \caption{\textbf{Setting A with $q=2$, $p=3$ and weak instruments: }The ternary plots are colored by first (left) and second (right) instrument. The composition of the instrument is barely influenced by the value of $Z$.}
\label{appfig:linear_p3_ternary_weak}
\end{figure}

\begin{figure}
  \centering
  \includegraphics[width=1\textwidth]{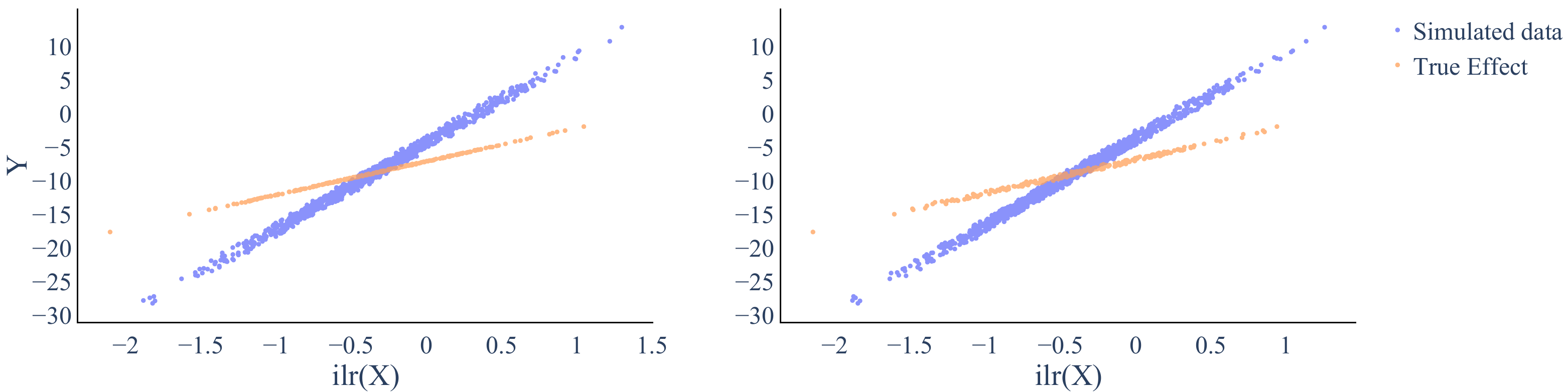}
  \caption{\textbf{Setting A with $q=2$, $p=3$ and weak instruments: } Both plots show one component of $\ilr(X) \in \bR^2$ vs.\ the confounded outcome (blue) and the true effect (orange). Due to the confounding, both effects do not overlap. As we are in the weaker instrument setting, we expect the methods to perform not as stable as in the previous cases where we had a stronger instrument available.}
\label{appfig:linear_p3_xy_weak}
\end{figure}

For testing in a weak instrument setting, we return to Setting A.
We mostly control the instrument strength via $\alpha$ and use higher or lower $\alpha$ values to obtain a strong or weak instrument setting.
We choose the following parameters for a weak instrument:
\begin{equation}
    \mu_c = -2 ,
    \alpha_0 = [4, 1] ,
    \alpha = \begin{bmatrix} 0.05 & 0.01 \\ 0.2 & 0 \end{bmatrix} ,
    c_X = [1, 1] ,
    \beta_0 = 2 ,
    \beta = [6, 2] ,
    c_Y = 4
\end{equation}
The first stage F-test for the two components of $\ilr(X)$ gives $(6.9,  4.7)$, much weaker than the previous settings.
Again we show a barplot (\cref{appfig:linear_p3_barplot_weak}) and a ternary plot (\cref{appfig:linear_p3_ternary_weak}) of the generated data.
The observed data as well as the true causal effect are shown in \cref{appfig:linear_p3_xy_weak}.

\subsubsection*{Nonlinear Second Stage}

Contrary to the previous scenarios, we now consider a non-linear $f$, resulting in a misspecified second stage for most of our methods. Note that in this scenario all two-stage methods as well as the naive regression will be misspecified in the second stage.

\begin{figure}
  \centering
  \includegraphics[width=.45\textwidth]{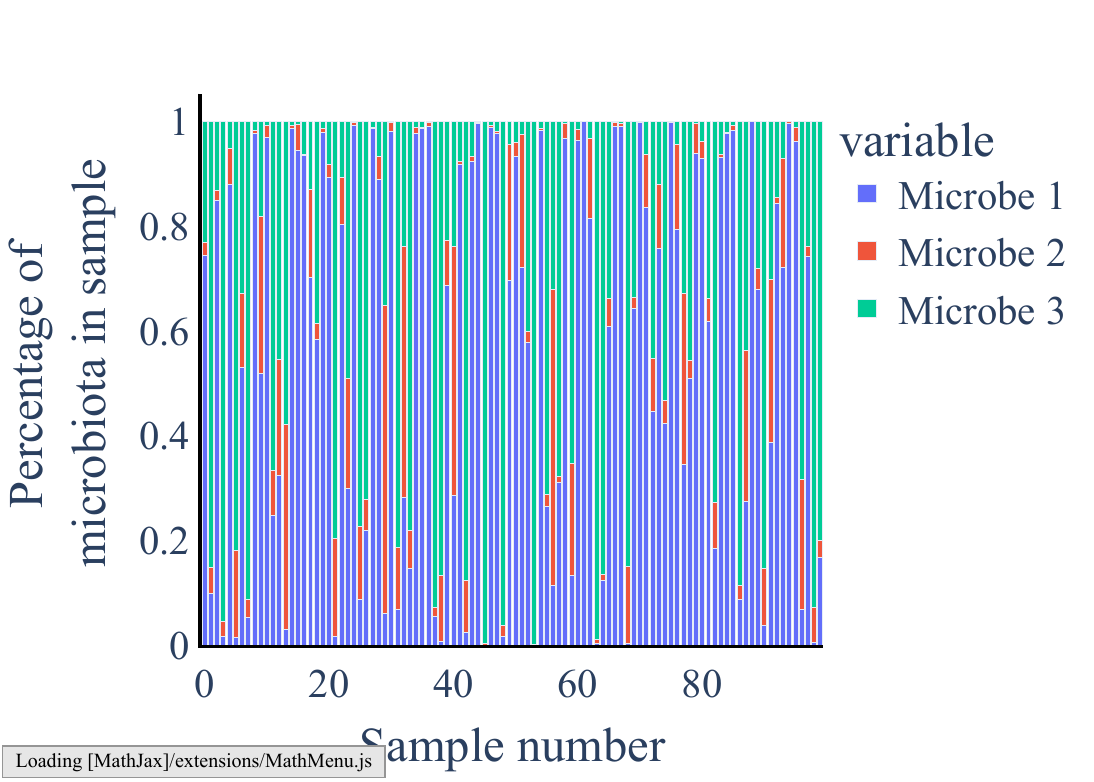}
  \caption{\textbf{Setting A with $p=3$, $q=2$ and a non-linear function form of $f$: } The barplot shows the different composition for each sample (for the $100$ first data points). Microbe 1 and 2 dominate the composition with high variance.}
\label{appfig:nonlinear_p3_barplot}
\end{figure}

\begin{figure}
  \centering
  \includegraphics[width=.45\textwidth]{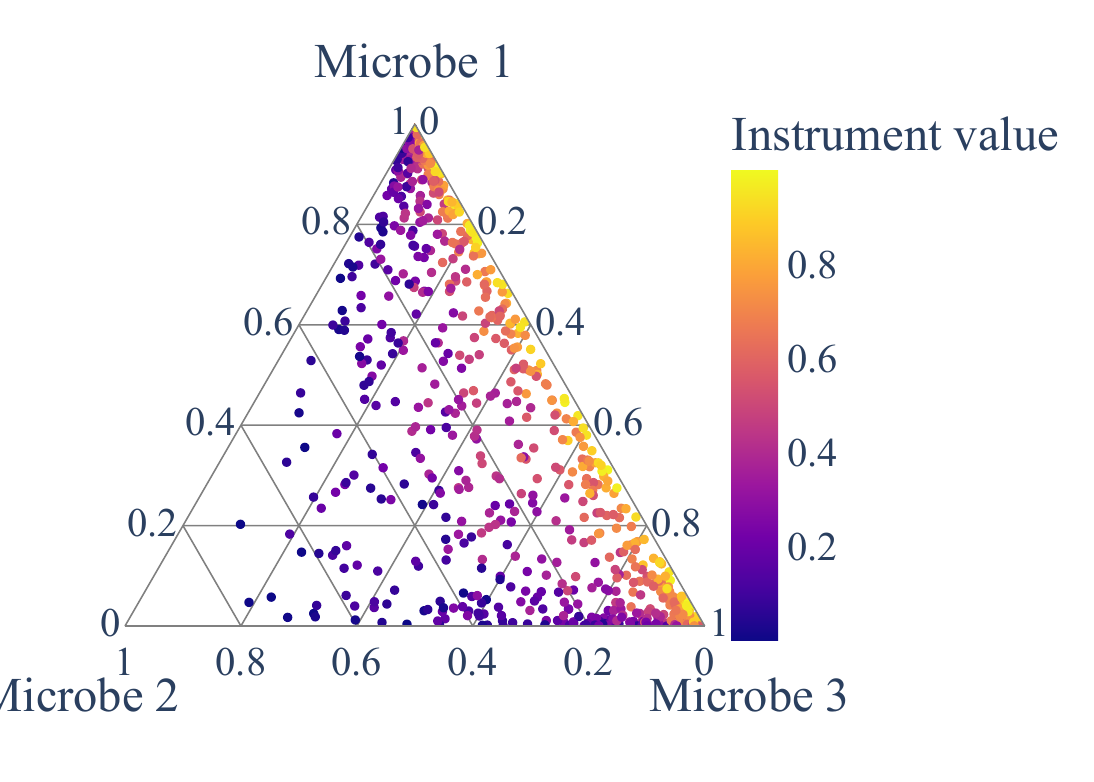}
  \hfill
  \includegraphics[width=.45\textwidth]{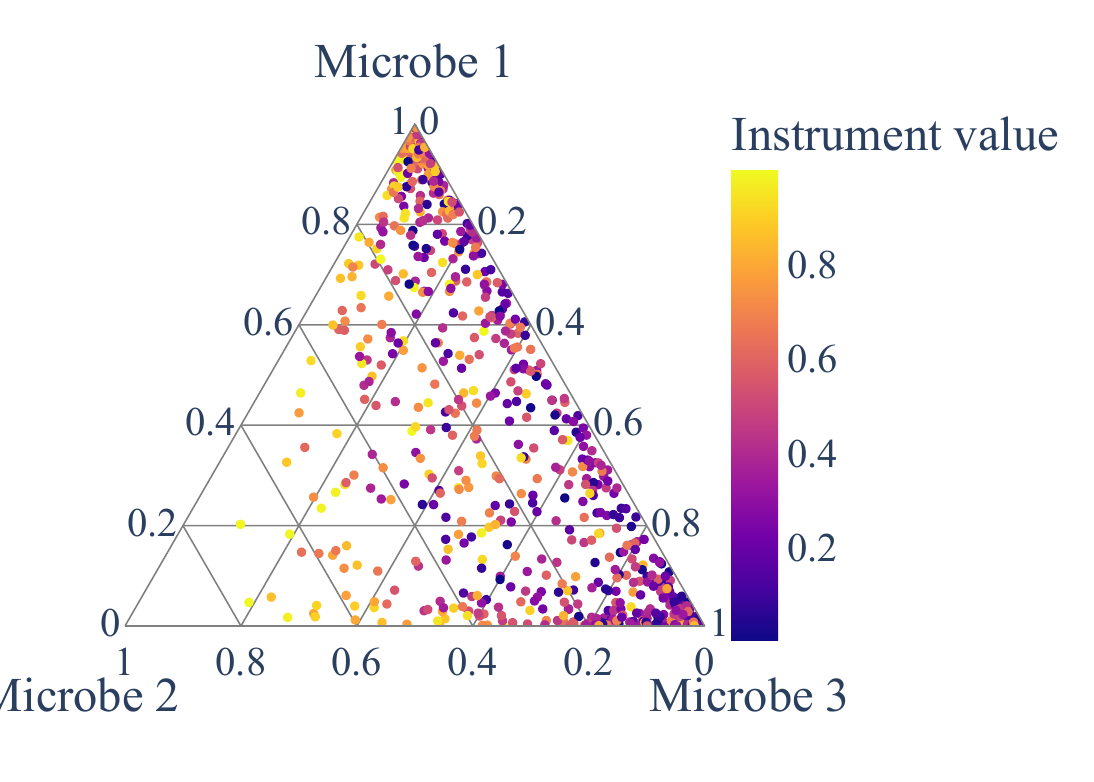}
  \caption{\textbf{Setting A with $p=3$, $q=2$ and a non-linear function form of $f$: } The ternary plots for the non-linear setup with $q=2$, colored by first (left) and second (right) instrument. Note that the first stage is still linear in $\ilr(X)$. Thus, the generation of the $X$ values is not affected by the change in $f$.}
\label{appfig:nonlinear_p3_ternary}
\end{figure}

\begin{figure}
  \centering
  \includegraphics[width=1\textwidth]{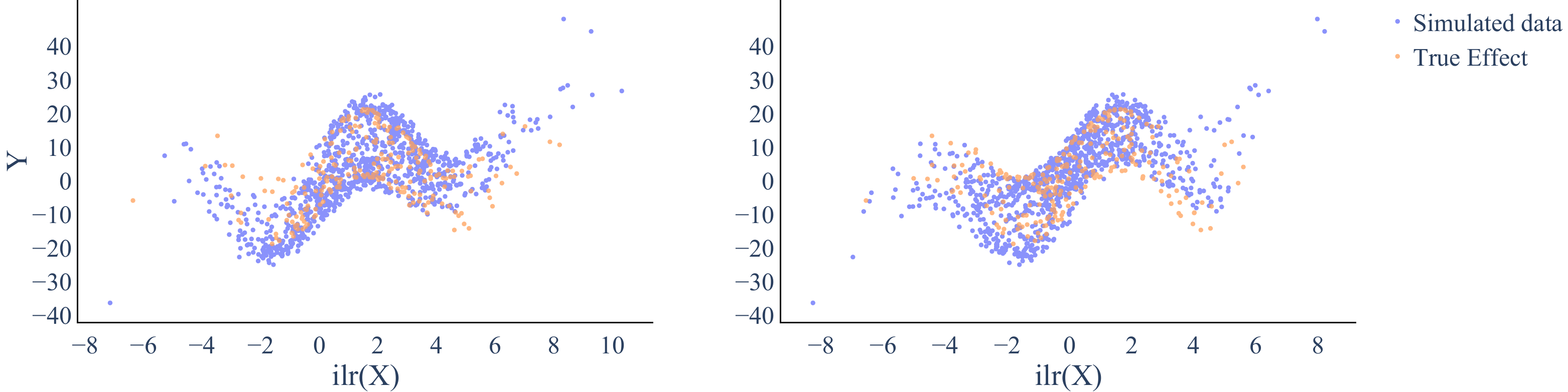}
  \caption{\textbf{Setting A with $p=3$, $q=2$ and a non-linear function form of $f$: } Both plots show one component of $\ilr(X) \in \bR^2$ vs.\ the confounded outcome (blue) and the true effect (orange). The effect both of the confounded outcome and the true effect show a non-linear dependency towards the individual $\ilr(X)$ components.}
\label{appfig:nonlinear_p3_xy}
\end{figure}

\paragraph{Setting A with $p=3, q=2$ and non-linear $f$}
Specifically, we replace the linear function for $Y$ in \cref{eq:lineary} with
\begin{equation}
Y = \beta_0 + \frac{1}{100} \B{1}^T (\ilr(X) + 1)^3 + 10 \cdot \B{1}^T\sin(\ilr(X))  + c_Y U.
\end{equation}
The remaining parameters are chosen to yield a strong instrument, ensuring that any performance differences are not (in addition) due to weak instrument bias:
\begin{equation}
    \mu_c = -1 ,
    \alpha_0 = [1 , 1] ,
    \alpha = \begin{bmatrix} 4 & 1 \\ -1 & 3 \end{bmatrix} ,
    c_X = [2, 2]
    \beta_0 = 5 ,
    \beta = [6  , 2] ,
    c_Y = 4
\end{equation}
Note that in this setting $\beta$ cannot be interpreted directly as the causal parameters, since the true causal effect also has a non-linear dependence on $\ilr(X)$.
Since the first stage remains unchanged, we can still use an F-test to assess instrument strength, which results in $(164.7, 76.4)$, a solid indicator for a strong instrument.
Again we show a barplot (\cref{appfig:nonlinear_p3_barplot}) and a ternary plot (\cref{appfig:nonlinear_p3_ternary}) of the generated data.
The observed data as well as the true causal effect are shown in \cref{appfig:nonlinear_p3_xy}.

\subsubsection*{Scarce Data Example $p \gg n$}

We return to Setting A with linear dependencies in both stages. 
However, in the scenarios before, we assumed a large dataset ($n=10,\!000$) for the methods to work on. 
In many real applications, this might not be the case.
Thus we choose to include an additional robustness aspect concentrating on the scenario $p \gg n$. In this particular case we chose $p=250$ and $n=100$.


\paragraph{Setting A with $p=250, q=10$ and $n=100$}
The choice of parameter is the same to Setting A with $n=10,\!000$, however we only include the first $100$ samples for the estimation:
\begin{align*}
\mu_c = 3,\,
\alpha_0 = [1, 1, 3, 1, 1, 1, 3, 1, 1, 1, 3, 1 , 0, \cdots , 0],\;
&\alpha_{ij} \begin{cases} 0, &\text{for }i \not = j \text{ and } i, j > 8, \\ 1,  &\text{for }i \not =  j \leq 8 \end{cases}, \\
c_X = [-1, 2, -1, 2, -1, 2, -2, 1, -2, 1, -2, 1, 0, \cdots , 0] ,\,
\beta_0 = 5 ,\;
&\beta_{\log} =  [10, 5, 5, 5, -10, -5, -5, -5, 0, \cdots , 0] , \\
\beta = V^T \cdot \beta_{\log},\,
c_Y= 5
\end{align*}
for $i \in \{1, \ldots, p-1\}, j \in \{1, \ldots, q\}$ and $V$ providing the orthonormal basis for the $\ilr$-transformation (see \cref{supp:compositional_transformations}).

\begin{figure}
  \centering
  \includegraphics[width=.45\textwidth]{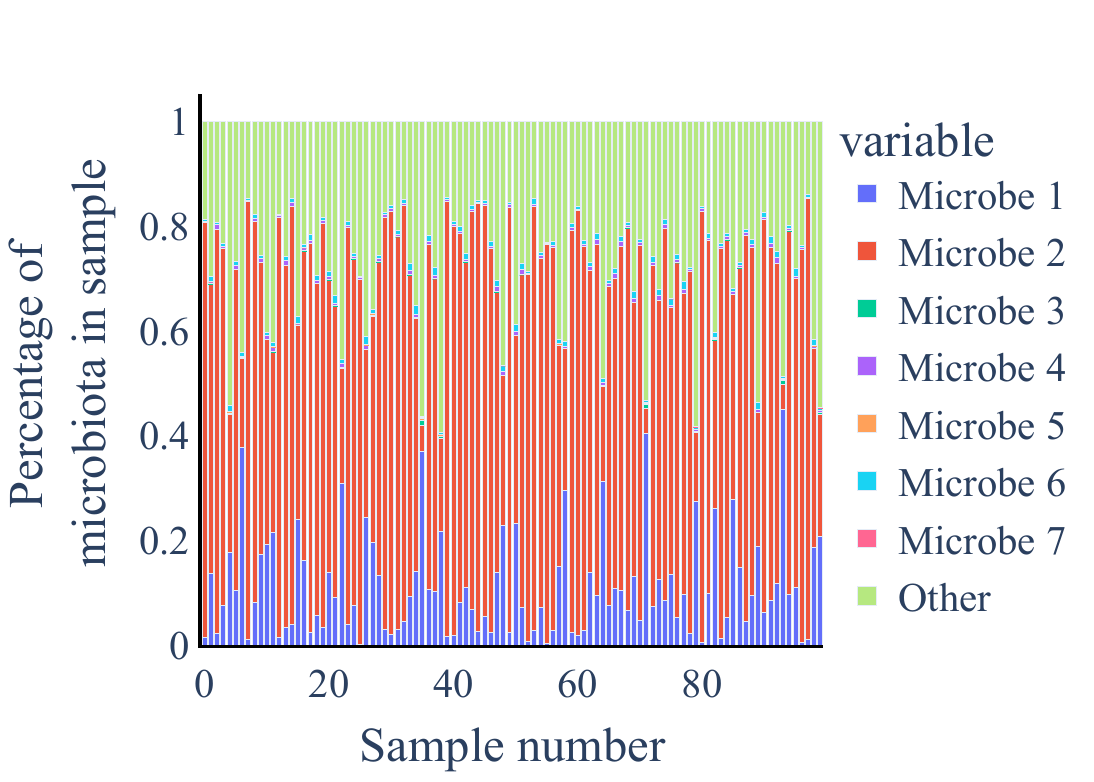}
  \caption{\textbf{Setting A with $n=100$, $p=250$, $q=10$: }
  The barplot shows the composition of the $100$ samples. The compositions are dominated by a few species.}
\label{appfig:linear_p250_barplot_n100}
\end{figure}

\begin{figure}
  \centering
  \includegraphics[width=1\textwidth]{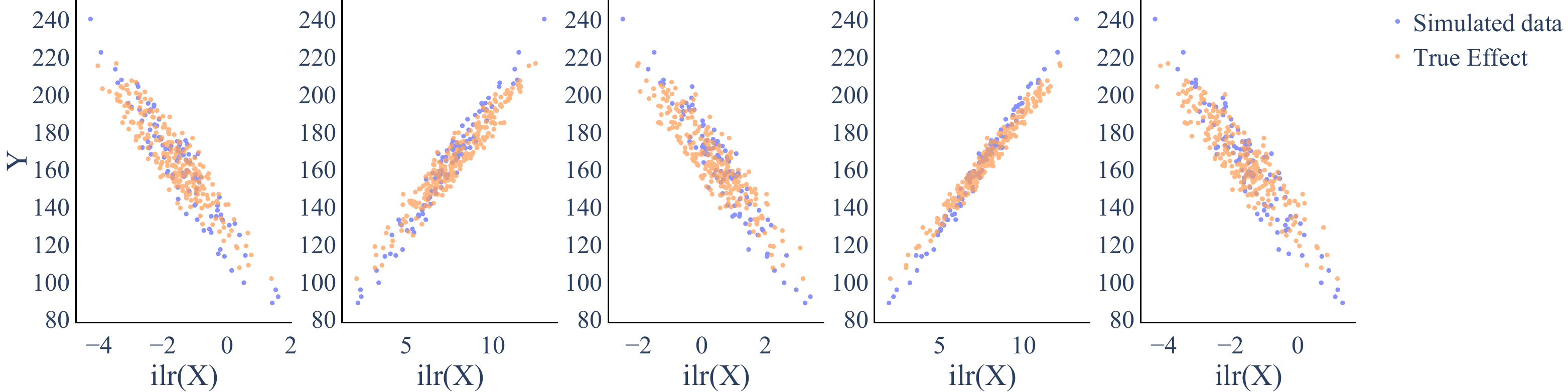}
  \caption{\textbf{Setting A with $n=100$, $p=250$, $q=10$:} Both plots show one component of $\ilr(X) \in \bR^{249}$ vs.\ the confounded outcome (blue) and the true effect (orange). Due to the confounding, the observed and the causal effect do not overlap. However, we expect the instrument $Z$ to factor out the confounding effect and enable the two-stage methods to identify the causal effect. Note that now we consider a sample size of $n=100$.}
\label{appfig:linear_p250_xy_n100}
\end{figure}

For the sake of completeness, we show a barplot (\cref{appfig:linear_p250_barplot_n100}) of the generated data. We note that the samples are the first $100$ samples of the larger dataset of the original Setting A with $p=250, q=10$ and $n=10,\!000$.
The observed data as well as the true causal effect are shown in \cref{appfig:linear_p250_xy_n100}.


\section{Method Training}
\label{supp:method_training}

\subsection*{Dirichlet Regression}

The mean of the Dirichlet distribution is given by $\mu_{\text{Diri}} = \frac{\alpha_j}{\sum_{j=1}^p \alpha_j}$.
Here, we consider the following model for the mean components
\begin{align}
\E[X_{ij}] &= \frac{\alpha_j}{\sum_{j=1}^p \alpha_j} = \frac{\alpha_j(Z_i)}{\sum_{j=1}^p \alpha_j(Z_i)} \\
\log(\alpha_j(Z_i)) &= \omega_{0j} + \omega_j Z_j.
\end{align}
The maximum likelihood function is then given by
\begin{align}
    l(\alpha; X, Z) &= \frac 1 n \sum_{i=1}^n \log \Gamma \Big ( \sum_{j=1}^p \exp \{\omega_{0j} + \omega_j Z_{i}\} \Big )  \\ &{} + \frac 1 n \sum_{i=1}^n \sum_{j=1}^p \Bigg ( \log (X_{ij}) \Big( \exp \{\omega_{0j} + \omega_j Z_{i}\} - 1\Big)  - \log \Gamma \Big( \exp \{\omega_{0j} + \omega_j Z_{i}\}\Big)\Bigg ).
\end{align}
Additionally, we introduce a sparsity enforcing regularization term to arrive at the following objective function
\begin{align}
    \min_{\omega} - l(\alpha; X, Z) + \lambda_{\text{dirichlet}} \sum_{j=1}^p |\omega_j|
\end{align}
with $\lambda_{\text{dirichlet}} \geq 0$.
For each Dirichlet regression, we pick $\lambda_{\text{dirichlet}}$ from the set $\{0.1, 1, 2, 5, 10\}$ by model selection via the Bayesian Information Criterion ($BIC = q \cdot \log(n) - 2 \cdot(\hat{L})$, with $\hat{L}$ being the likelihood value). We train the model for each available $\lambda$ value in the set and choose the model with minimal BIC.
For the starting point $\alpha_{\text{start}}$ we fit a Dirichlet distribution on those $X$ for which all $|Z| < 0.2$ by maximum likelihood estimation.  

\subsection*{Log-contrast Regression}

The log-contrast regression is enforcing sparsity via an $\ell_1$ penalty on the $\beta$ parameters.
\begin{equation}
  \min_{\beta} \sum_{i=1}^n \cL(x_i, y_i, \beta) + \lambda \| \beta \|_1 \quad \text{s.t.}\: \sum_{i=1}^p \beta_i = 0 \:.
\end{equation}
This estimation respects the compositional nature of $x$ while retaining the association between the entry $\beta_i$ and the relative abundance of the individual component $x_i$.

In our examples, we focus mainly on continuous $y \in \bR$ and the squared loss $\cL(x, y, \beta) = (y - \beta^T \log(x))^2$.
However, the framework 
also supports different loss functions. 

For robust Lasso regression, the Huber loss can be applied.
\begin{equation}
    \cL(x_i, y_i, \beta) = \mathcal{H}_\delta(x_i, y_i, \beta)=
    \begin{cases} 
        \frac{1}{2}(y_i - \beta^T \log(x_i))^2 &\text{ for } |y_i - \beta^T \log(x_i)| < \delta \\ 
        \delta(|y_i - \beta^T \log(x_i)| - \frac{1}{2}\delta), & \text{ otherwise.} 
        \end{cases}
\end{equation}
The Huber Loss combines the squared loss and the absolute loss. It is less sensitive to outliers than the squared loss, but remains differentiable at $0$ in contrast to the absolute loss.

Moreover, for classification tasks with $y_i \in \{-1, 1\}$, we can directly use the squared Hinge loss  for $\cL$ with:
\begin{equation}
    \cL(x_i, y_i, \beta) =  l(x_i, y_i, \beta) \text{ with } l(x_i, y_i, \beta) =
    \begin{cases} 
        (1 - (y_i  \beta^T \log(x_i))^2,  &\text{ if } y_i \beta^T \log(x_i) \leq 1 \\ 
        0,  & \text{ if } y_i \beta^T \log(x_i) > 1 
        \end{cases}\label{eq:classo_classification}
\end{equation} 
or a ``Huberized'' version thereof:
\begin{equation}
    \cL(x_i, y_i, \beta) =  l_\delta(x_i, y_i, \beta) \text{ with } l_\delta(x_i, y_i, \beta) =
    \begin{cases} 
        (1 - (y_i  \beta^T \log(x_i))^2,  &\text{ if } \delta \leq y_i \beta^T \log(x_i) \leq 1 \\ 
        (1 - \delta) (1 + \delta - 2 y_i \beta^T \log(x_i)), & \text{ if } y_i \beta^T \log(x_i) \leq \delta \\
        0,  & \text{ if } y_i \beta^T \log(x_i) > 1 
        \end{cases}
\end{equation}

We refer to \cite{simpson2021classo} for further loss functions and a more detailed overview.

We now continue with the description of the setup used in the following result section. The results on the synthetic data and the real data in \cref{supp:method_results} are based on the squared loss:
\begin{equation}
  \min_{\beta} \sum_{i=1}^n \|y_i - \beta^T \log(x_i) \|_2^2 + \lambda \| \beta \|_1 \quad \text{subject to}\: \sum_{i=1}^p \beta_i = 0 \:.
\end{equation}
Furthermore, for the real data we also show the results for a binary outcome $y_i \in \{-1, 1\}$ based on the squared Hinge loss (\cref{eq:classo_classification}).

We solve the underlying optimization problems with the c-lasso package, a Python package for constrained sparse regression \citep{simpson2021classo}. The c-lasso packages comprises several model selection schemes, including a theoretically-derived  $\lambda_0$ parameter, k-fold cross-validation, and stability selection. 

Here, we consider stability selection for tuning $\lambda$. The method comprises the hyperparameter $t_{\text{threshold}}$ which determines the number of coefficients included in the final model. 
In our training, we set the same $t_{\text{threshold}}$ for the naive regression as well as the two-stage methods to have a fair comparison. In all our training scenarios with generated data we find $t_{\text{threshold}}=0.7$ to be a reasonable default value. For the real data scenario we found $t_{\text{threshold}}=0.65$ to be more sensible.

\begin{figure}
  \centering
  \includegraphics[width=1\textwidth]{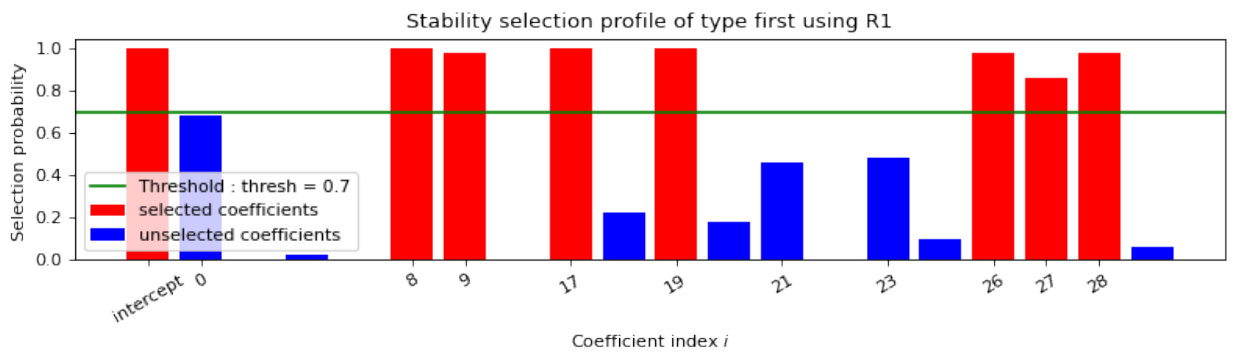}
  \hfill
  \includegraphics[width=1\textwidth]{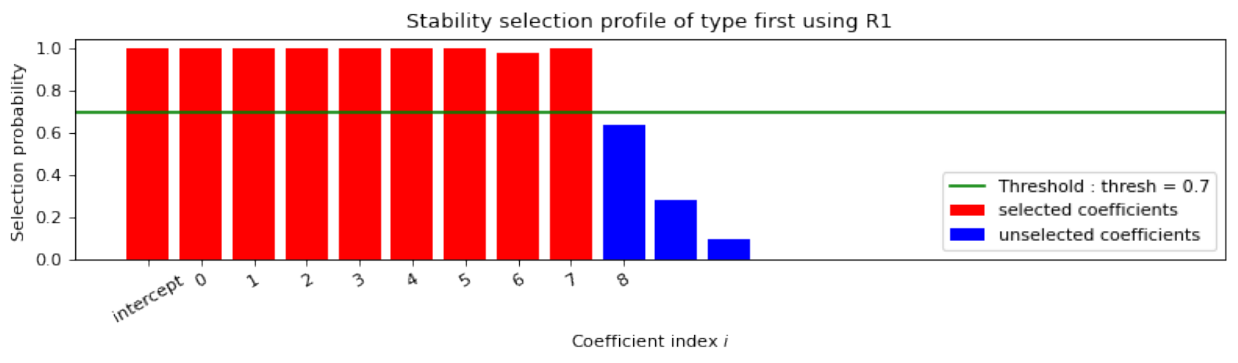}
  \caption{\textbf{Stability profiles for sparse log contrast regression with c-lasso: } The barplots show the model selection probability of the $\beta$ coefficients. The upper panel shows the example for the naive regression. The lower panel shows the results for the same setting for ILR+LC regression. Both models are fairly certain about the main drivers.}
\label{appfig:BetaSignificance}
\end{figure}

We use Setting B with $p=30$ and $q=10$ as a representative example to illustrate the impact of the threshold value. \Cref{appfig:BetaSignificance} shows the stability profile of the $\beta$ coefficients and their attributed probability of entering the model. The threshold value $t_{\text{threshold}}=0.7$ works as a cut off for the relevant coefficients. The upper panel shows the results for the naive regression, whereas the lower panel shows the results for the ILR+LC regression (working on the exact same data).

\begin{figure}
  \centering
  \includegraphics[width=1\textwidth]{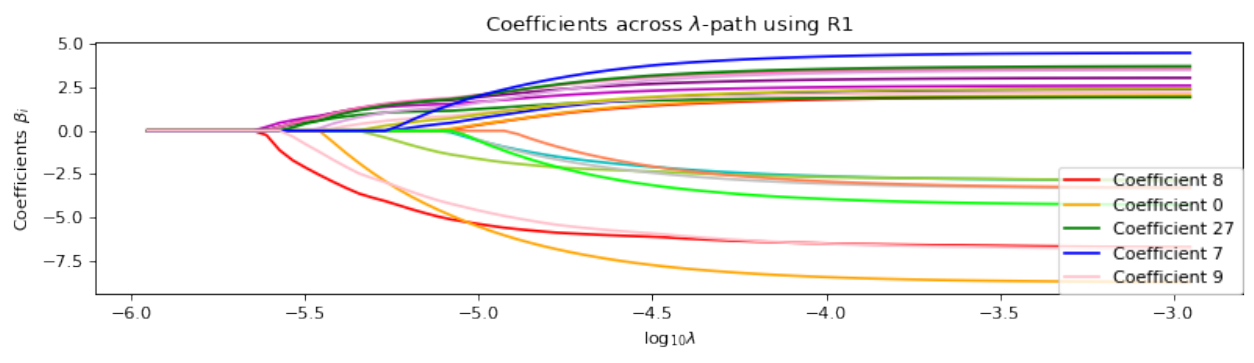}
  \hfill
  \includegraphics[width=1\textwidth]{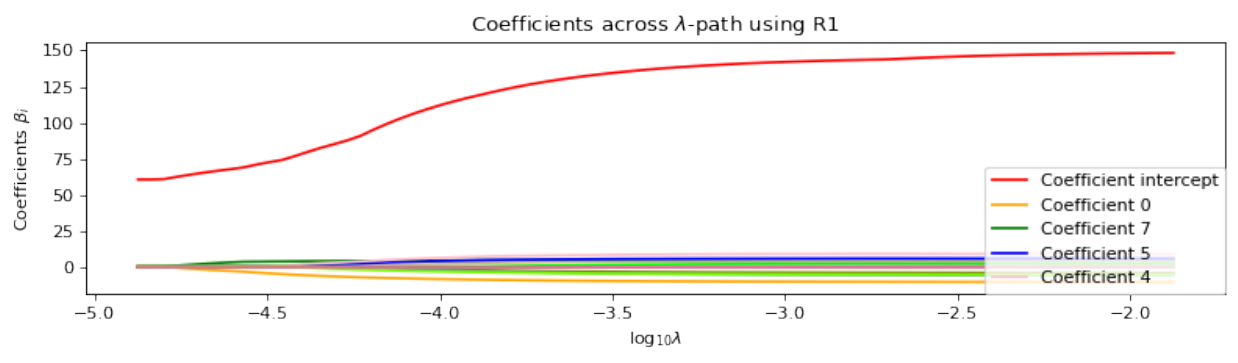}
  \caption{\textbf{Corresponding $\lambda$-path for $\beta$ coefficients: } The plots show the individual coefficients for the different $\lambda$ values. The upper plot shows the $\beta$ coefficients for the naive regression, the lower plot presents the coefficients for the two-stage method ILR+LC.}
\label{appfig:LambdaPath}
\end{figure}

Moreover, the method also returns the coefficient values across the $\lambda$-path, i.e., the entry of coefficients into the model for the corresponding $\lambda$ (see \cref{appfig:LambdaPath}).
Further improvements may be achieved by taking the path and individual analysis into account instead of proposing a general $t_{\text{threshold}}$, however, this simple yet effective approach was sufficient for our purposes in this work.

\section{Method Results}
\label{supp:method_results}

For the comparison of the different methods, we make use of three approaches:
\begin{enumerate}
    \item $\hat{\beta}$-MSE: As long as the second stage is wellspecified and linear, we can compare the estimated causal parameters $\hat{\beta}$ for the various approaches (where applicable).
    \item \emph{FZ/FNZ}: As long as the second stage is wellspecified and linear, we can additionally compare the number of false zero values and false non-zero values to quantify support recovery.
    \item \emph{OOS MSE}: In the general case, the causal performance measure is measured by an ``out of sample error''(\emph{OOS MSE}) which denotes the mean squared error between the true value of $Y$ under an intervention $do(X=x)$ and the predicted causal effect $\E[Y\mid do(X)]$ of our model, given by $\hat{f}(x)$. For the interventional $X$, we simulate $250$ additional compositional data points according to the underlying model, but using a different seed and thus disconnecting them from the instrument $Z$ and the confounder $U$. Thus, we receive a true interventional $X$ which still preserves data characteristics.
    
\end{enumerate}

For each data generating setup, we provide confidence intervals for the methods' results by performing the data generation and the method evaluation $50$ times on different random seeds. In each run, we sample $n=1000$ datapoints in the $p=3$ scenario and $10,\!000$ datapoints in the $p=30$ and $p=250$ scenario. We compute the OOS MSE as well as the $\hat{\beta}$-MSE and FZ/FNZ (if applicable).
Some of the figures in this section are extended or more complete versions of the numbers given in the table in the main body (see \cref{tab:res_all_settinga,tab:res_all_settingb}), where some less relevant results have been omitted for readability.

\subsection*{Setting A}

\subsubsection*{Setting A with $p=3, q=2$}

\begin{figure}
  \centering
  \includegraphics[width=.45\textwidth]{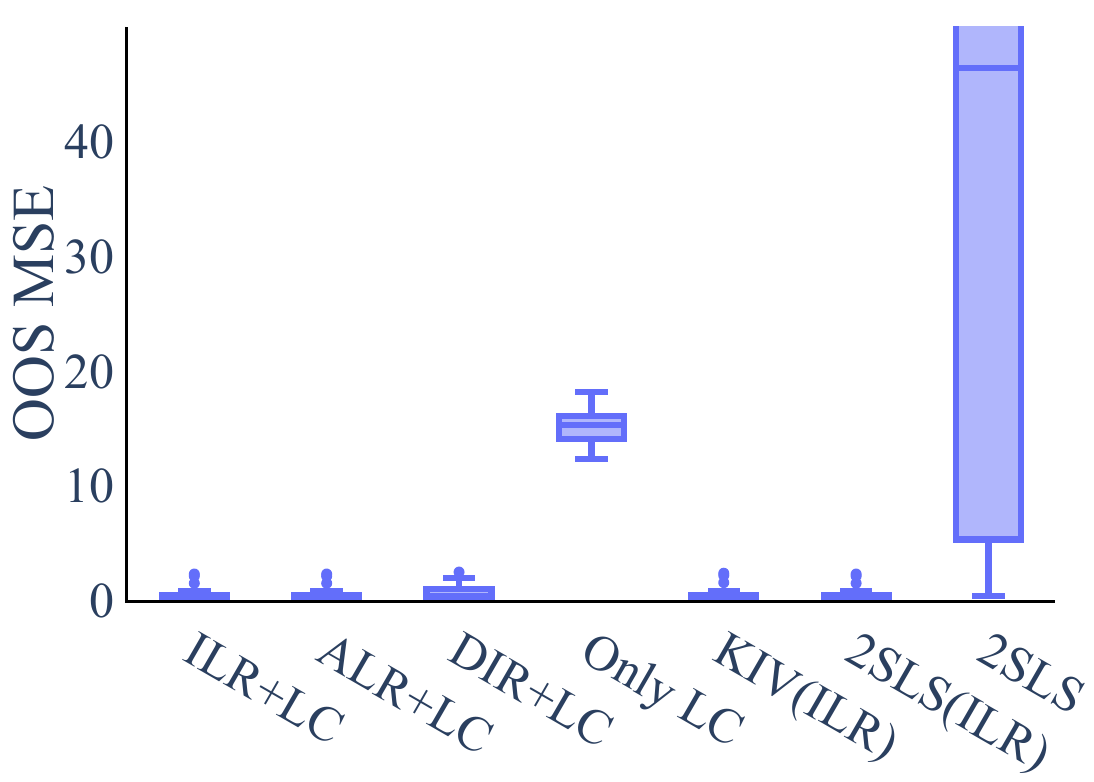}
  \hfill
  \includegraphics[width=.45\textwidth]{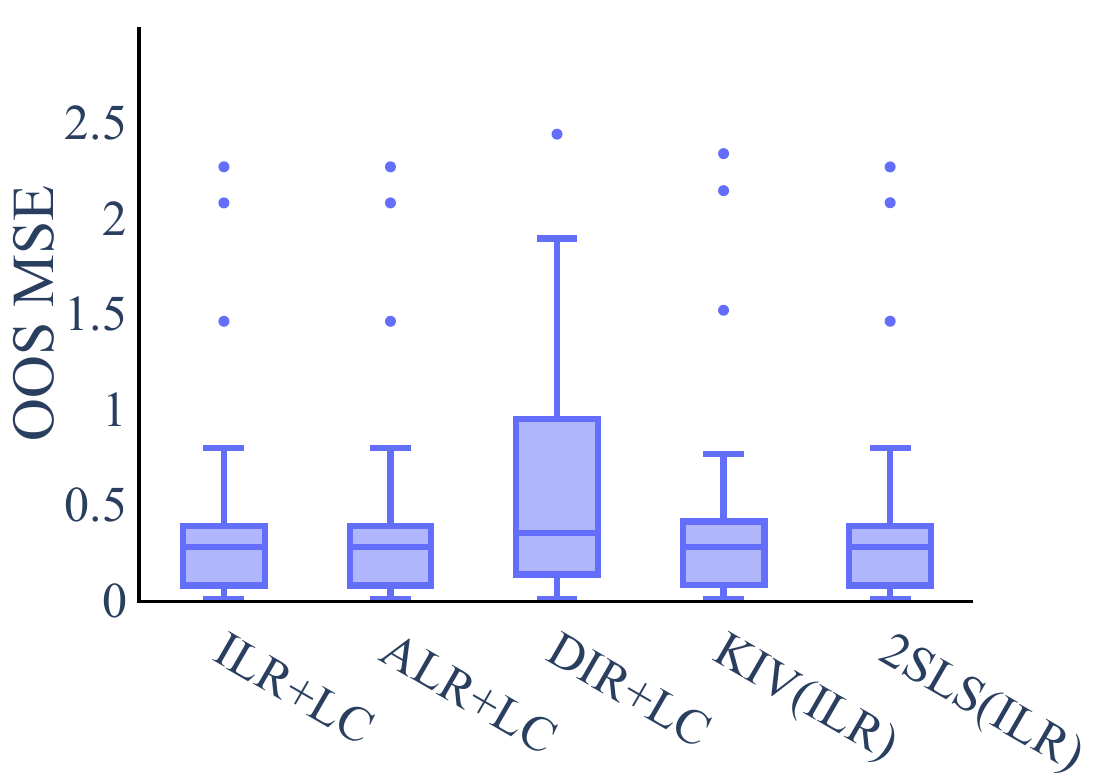}
  \caption{\textbf{Setting A with $p=3$, $q=2$: } The boxplots show the OOS MSE of $50$ runs. The naive regression Only LC and 2SLS (left) perform way worse compared to the other approaches. When we adjust the y-scale (right), DIR+LC also shows a higher OOS MSE than \ilrlc{} etc. DIR+LC possibly suffers from the misspecified first stage. Note that ALR+LC,\ilrlc{}, \ilrilr{} are equivalent in the low-dimensional case.}
\label{appfig:linear_p3_MSE}
\end{figure}

\begin{figure}
  \centering
  \includegraphics[width=.45\textwidth]{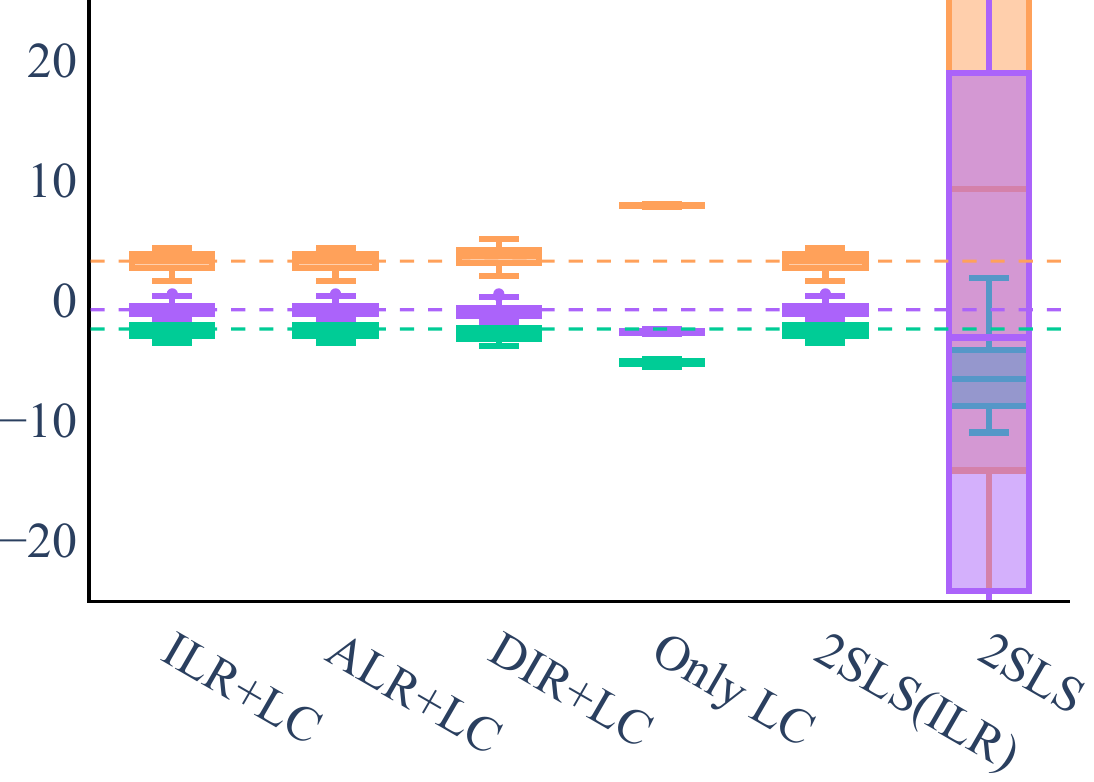}
  \hfill
  \includegraphics[width=.45\textwidth]{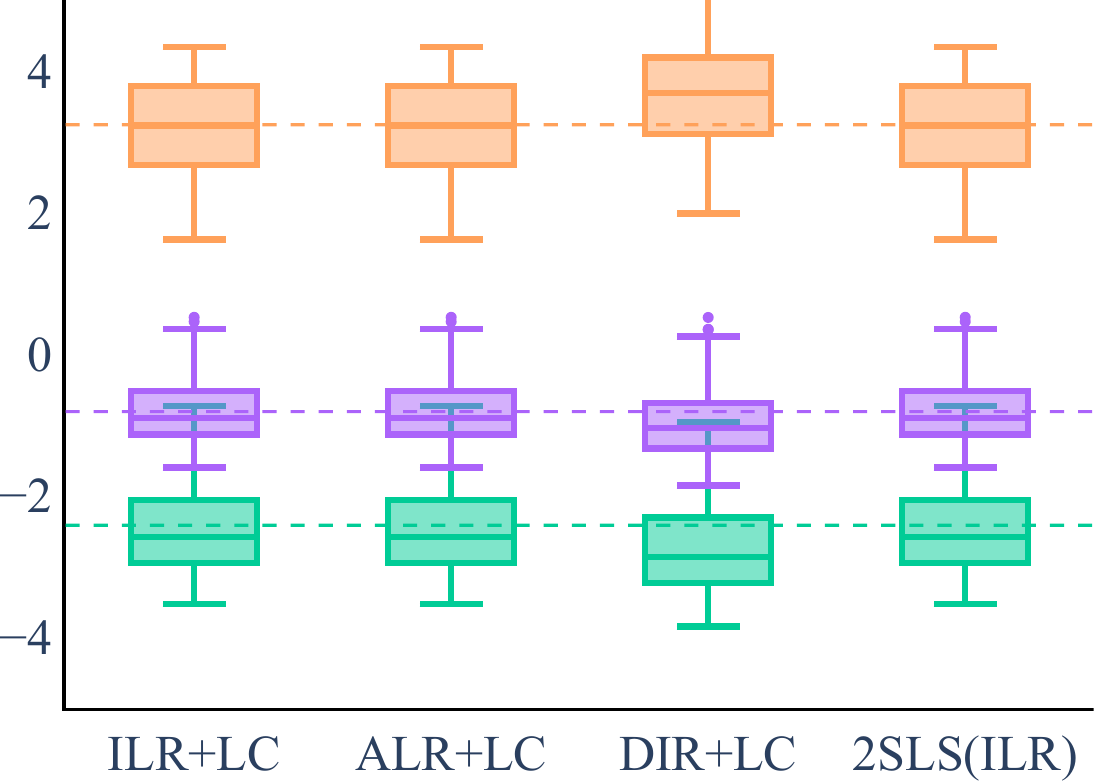}
  \caption{\textbf{Setting A with $p=3$, $q=2$: } The boxplots show the $\hat{\beta}$ values for the $50$ runs for each of the $3$ $\beta$ coefficients (dashed lines). The two-stage methods, except 2SLS, are able to recover the causal effect on average. The naive regression method overestimates the effect. Moreover, it does so with a high degree of confidence as there is barely any variation in the $\hat{\beta}$ estimates (left). When we adjust the y-scale (right), DIR+LC shows a notable bias towards the solution of the naive regression (left). This might suggest that DIR+LC indeed suffers from the misspecified first stage and thus is not able to make use of the instrument $Z$ as efficiently.}
\label{appfig:linear_p3_Beta}
\end{figure}

This setting is a wellspecified setting for ALR+LC, \ilrilr{} and ILR+LC. Moreover, confounding is present (see \cref{appfig:linear_p3_xy}) which additionally gives us reason to expect a much better performance of the two-stage methods than the naive regression $X \to Y$ in terms of OOS MSE. 
The results in \cref{appfig:linear_p3_MSE}, largely verify this expectation. 
The naive regression has a clear disadvantage due to confounding and picks up on spurious correlations as an effect coming from $X$. 
Two-stage methods work well when relying on a strong instrument, helping the methods to factor out the confounding and identifying the true casual effect. 
\Cref{appfig:linear_p3_Beta} shows the causal parameter estimates $\hat{\beta}$ and further corroborates our claims that two-stage methods significantly outperform naive regression. 
The effects found via naive regression overestimate the direct causal effect strength from $X$, whereas all two-stage methods recover the true causal parameters $\beta$ well. Only DIR+LC suffers slightly from the misspecified first stage compared to the other wellspecified two-stage approaches.
It is noteworthy that DIR+LC works reasonably well despite our manual two-stage procedure with a ``forbidden'' non-linear regression in the first stage.
Since we are in the low-dimensional setting with no sparsity regularization, the results of ILR+LC, ALR+LC and \ilrilr{} are equivalent.

\subsubsection*{Setting A with $p=30, q=10$}

Microbiome compositional data is typically high-dimensional and comprises many zero values. Moreover, it is often assumed that only a few microbial compositions (and hence $\beta$ parameters) influence an outcome of interest $Y$. 
Thus, in the following, we aim to be close to such a scenario by assuming a sparse $\beta$ as ground truth and by simulating $X$ with a few dominating compositions in the data generating process (see \cref{supp:data_generation}).


Note that for higher-dimensional approaches, we omit results for DIR+LC due to computational issues stemming from the maximum likelihood estimation of the $\alpha_0$ and $\alpha$ parameters in the first stage. 2SLS, which ignores the compositionality of $X$ altogether, is not able to converge at all.

\begin{figure}
  \centering
  \includegraphics[width=.45\textwidth]{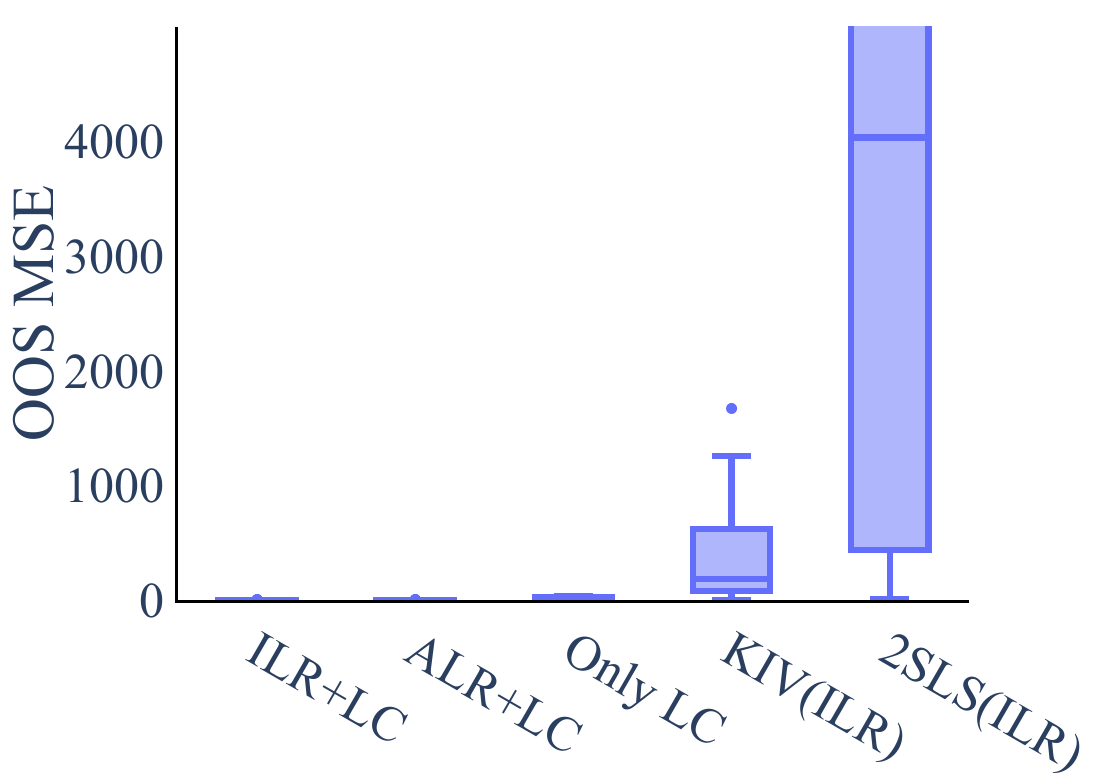}
  \hfill
  \includegraphics[width=.45\textwidth]{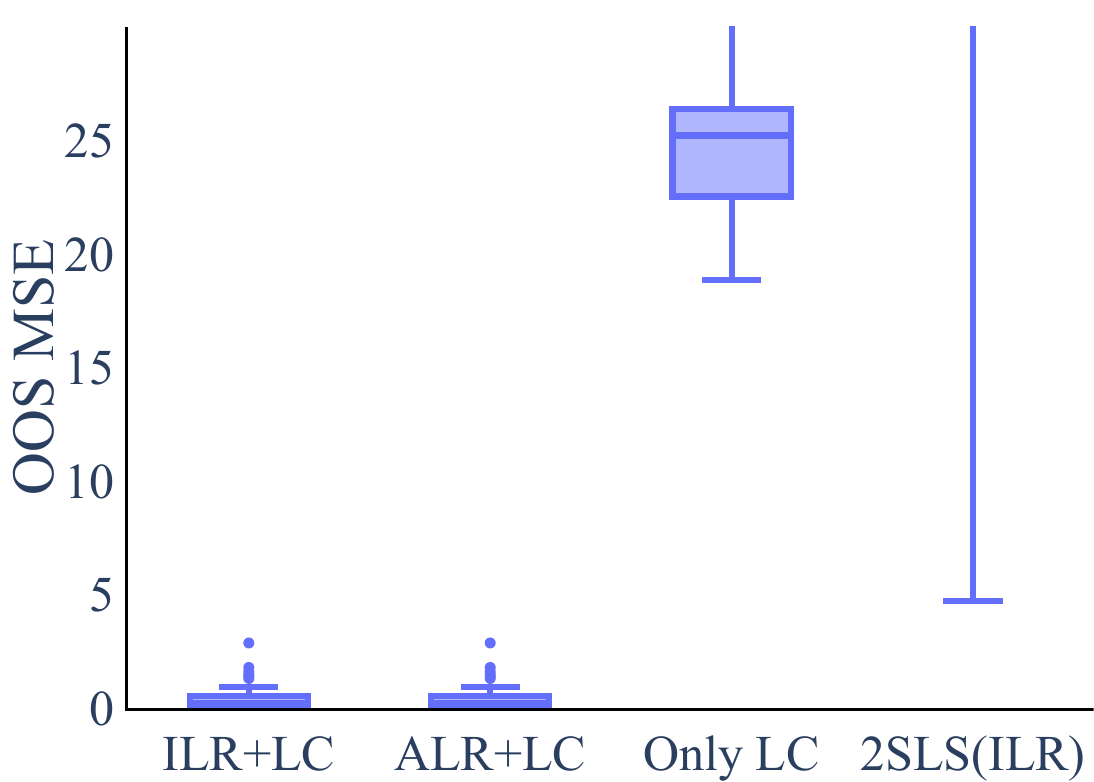}
  \caption{\textbf{Setting A with $p=30$, $q=10$: } The boxplots show the OOS MSE of $50$ runs. \ilrilr{} and \kivilr{} are volatile and lack sensible regularization (left). When we adjust the y-scale, we see that Only LC (right) performs also way worse compared to the regularized two-stage approaches.}
\label{appfig:linear_p30_MSE}
\end{figure}

\begin{figure}
  \centering
  \includegraphics[width=.45\textwidth]{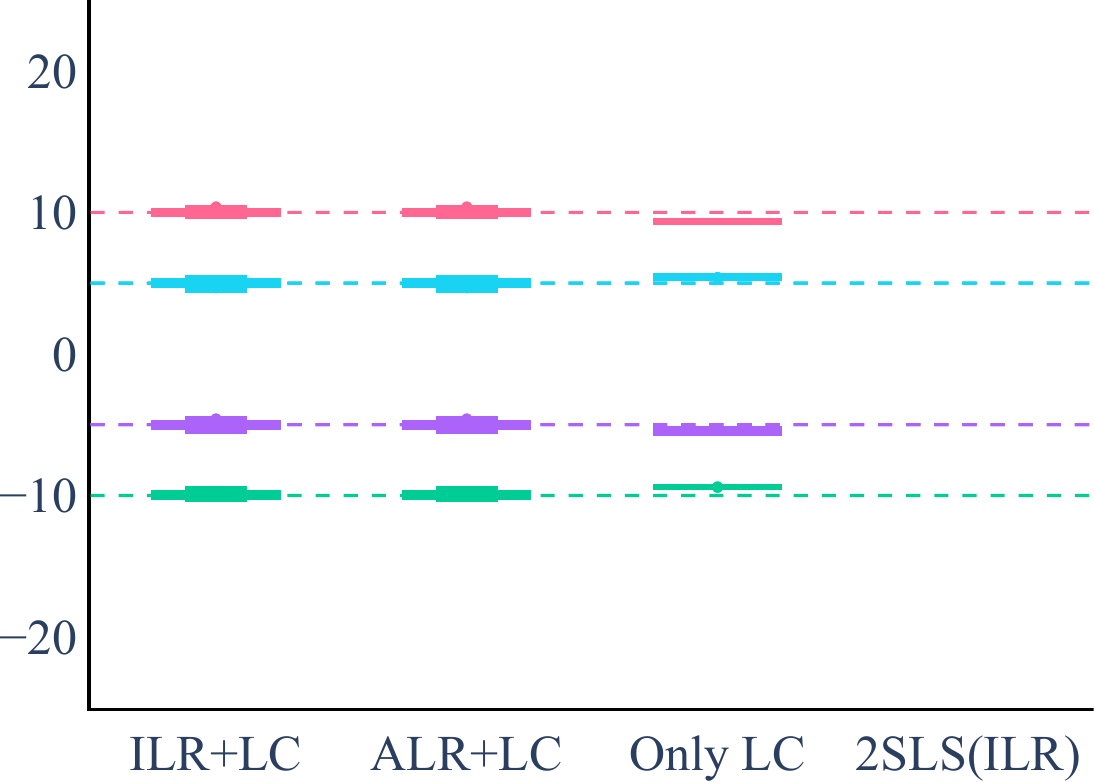}
  \hfill
  \includegraphics[width=.45\textwidth]{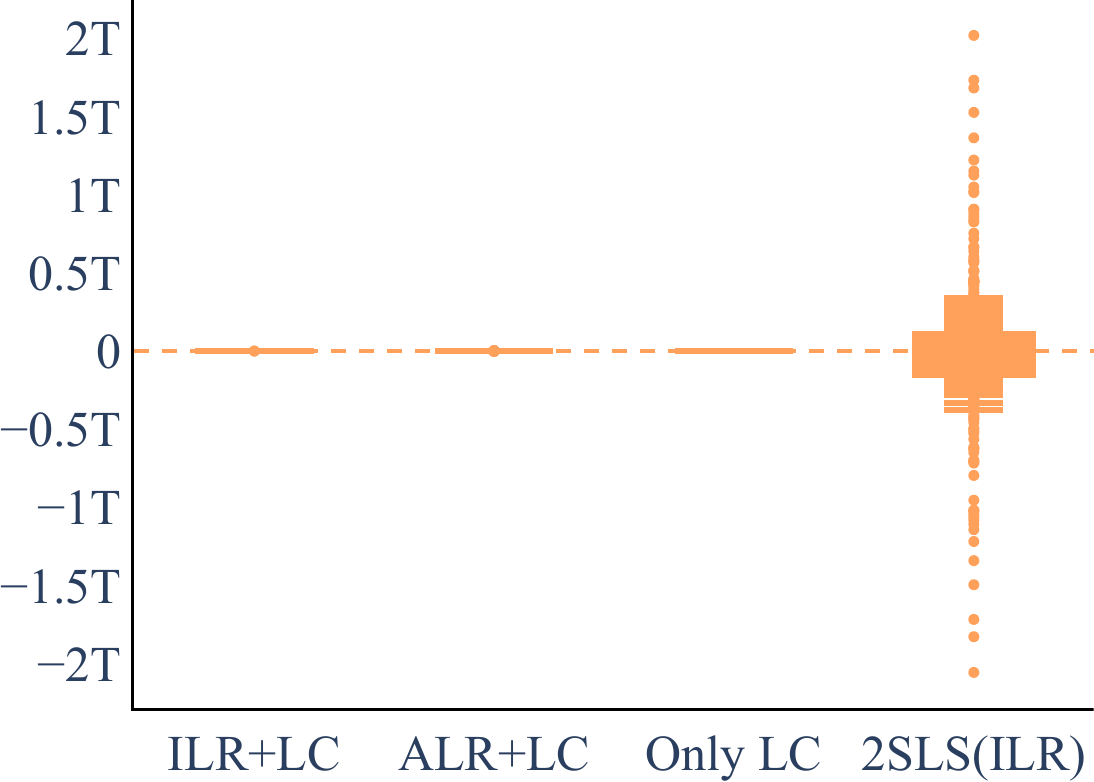}
  \caption{\textbf{Setting A with $p=30$, $q=10$:} The boxplots show the $\hat{\beta}$ values for the $50$ runs for each of the $8$ non-zero $\beta$ coefficients (dashed lines, left) and the $22$ zero $\beta$ coefficients (dashed line, right). The two-stage methods are able to recover the causal effect on average, whereas the naive regression methods overestimate the effect (left). Moreover, Only LC does so with a high degree of confidence as there is barely any variation in the $\hat{\beta}$ estimates. \ilrilr{} does not produce sensible estimates due to the missing regularization.}
\label{appfig:linear_p30_Beta}
\end{figure}

For higher dimensions, the lack of regularization in the ILR methods becomes obvious (\cref{appfig:linear_p30_MSE}), both for \ilrilr{} and \kivilr{}. The methods become more volatile and \ilrilr{} is unable to detect any zero values in $\beta$ (see \cref{appfig:linear_p30_Beta}). 
On the other hand, the naive regression is able to identify zero $\beta$s correctly, but suffers from confounding and thus over- or underestimates the true influential $\beta$s.
Only the regularized two-stage approaches are able to recover the true causal $\beta$s, both the influential coefficients as well as the zero values. 

\subsubsection*{Setting A with $p=250, q=10$}

To further test the approaches, we use another high-dimensional setup with $p=250$. Again, we make use of the common assumption that only a few microbial compositions (and hence $\beta$ parameters) influence an outcome of interest $Y$. 
We assume a sparse $\beta$ as ground truth and run the models on $X$ which has a few dominating species.

Note that for higher-dimensional approaches, we omit results for DIR+LC due to computational issues stemming from the maximum likelihood estimation of the $\alpha_0$ and $\alpha$ parameters in the first stage. 2SLS, which ignores the compositionality of $X$ altogether, is not able to converge at all.

\begin{figure}
  \centering
  \includegraphics[width=.45\textwidth]{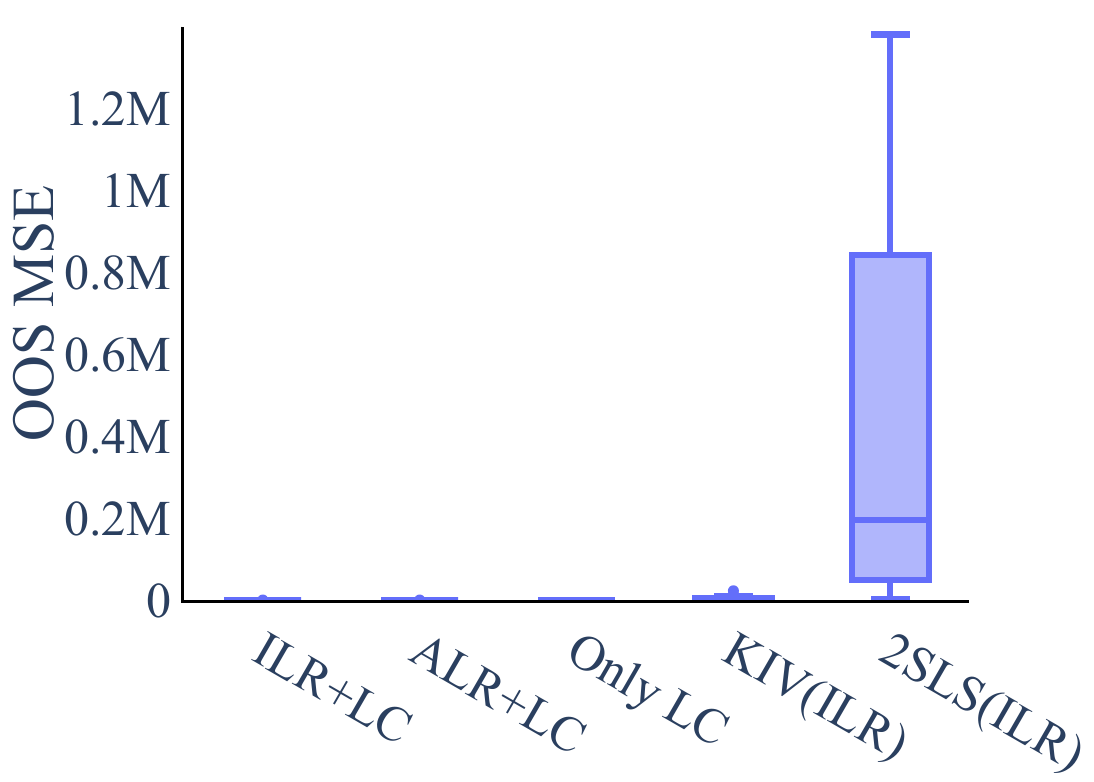}
  \hfill
  \includegraphics[width=.45\textwidth]{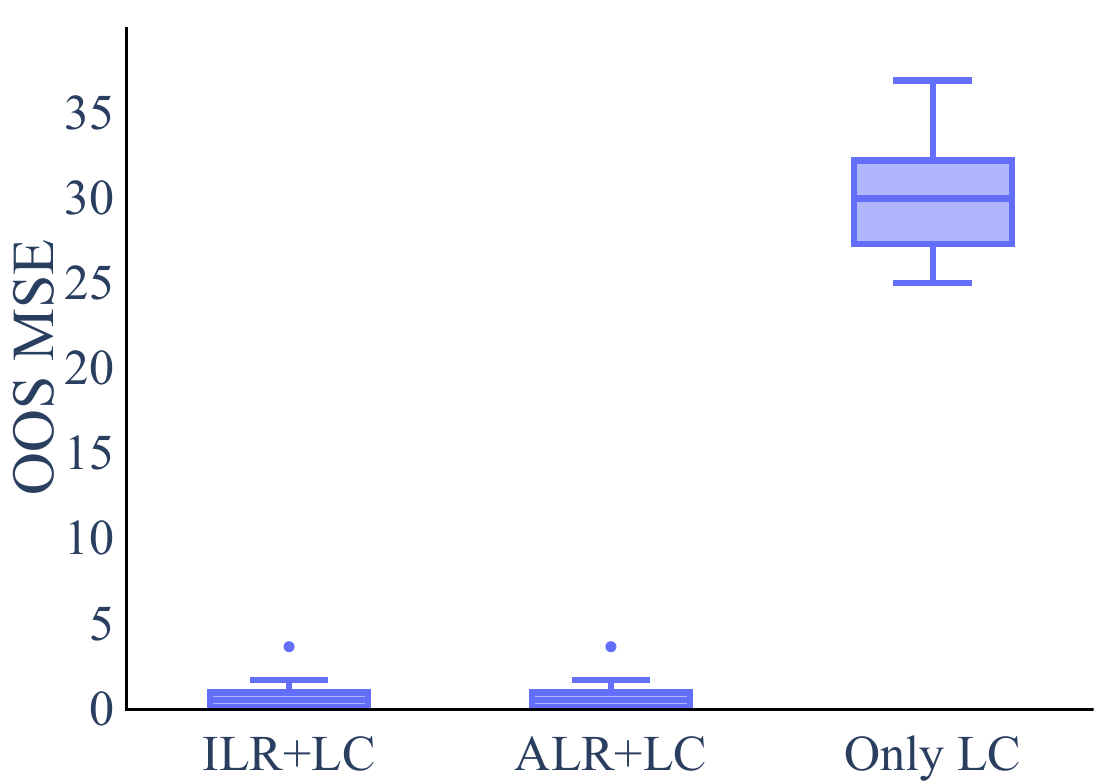}
  \caption{\textbf{Setting A with $p=250$, $q=10$: } The boxplots show the OOS MSE of $50$ runs. \ilrilr{} and \kivilr{} are volatile and lack sensible regularization (left). This problem is more pressing as the dimensionality grows. When we adjust the y-scale, we see that also Only LC (right) performs worse compared to the regularized two-stage approaches.}
\label{appfig:linear_p250_MSE}
\end{figure}

For $p=250$, the problem of missing regularization in the ILR methods (\ilrilr{} and \kivilr{}) becomes even more pronounced (\cref{appfig:linear_p250_MSE}).
For readability we thus omitted \ilrilr{} from the $\beta$ plots. 
Moreover, the naive regression is not even able to recover the full support, as it only identifies most, but not all, of the zero and non-zero $\beta$s correctly (see \cref{appfig:linear_p30_Beta}).
Only the regularized two-stage approaches are able to recover the true causal $\beta$s. 

\begin{figure}
  \centering
  \includegraphics[width=.45\textwidth]{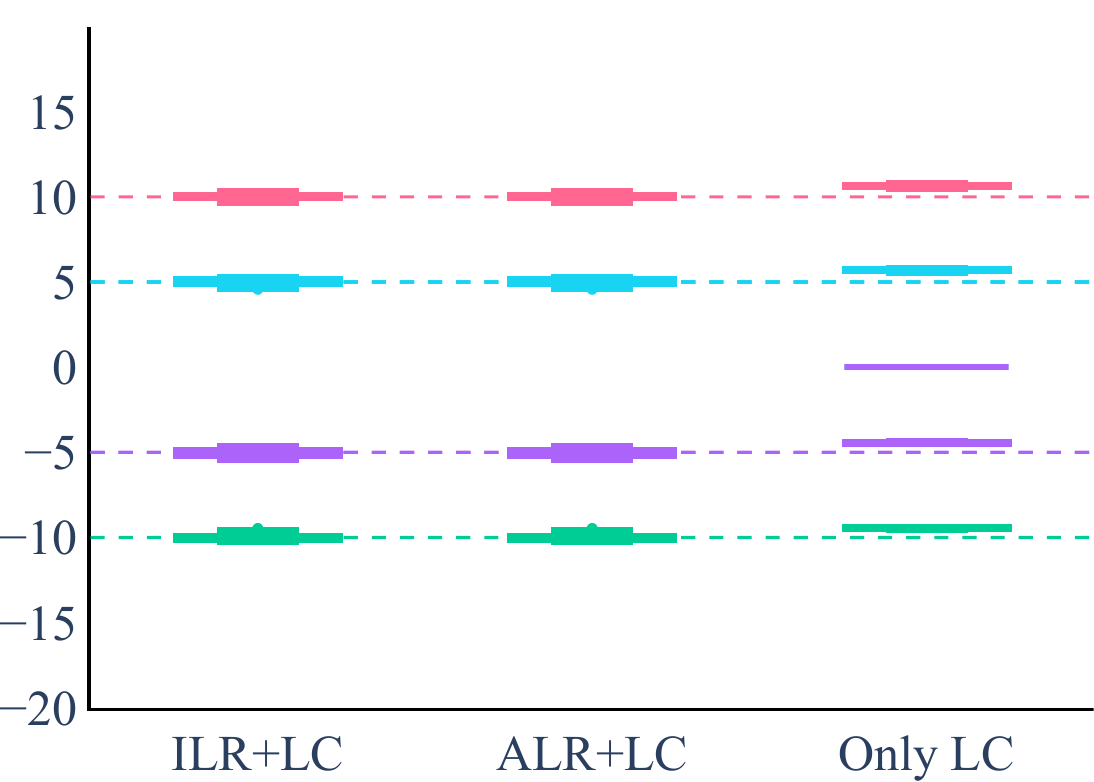}
  \hfill
  \includegraphics[width=.45\textwidth]{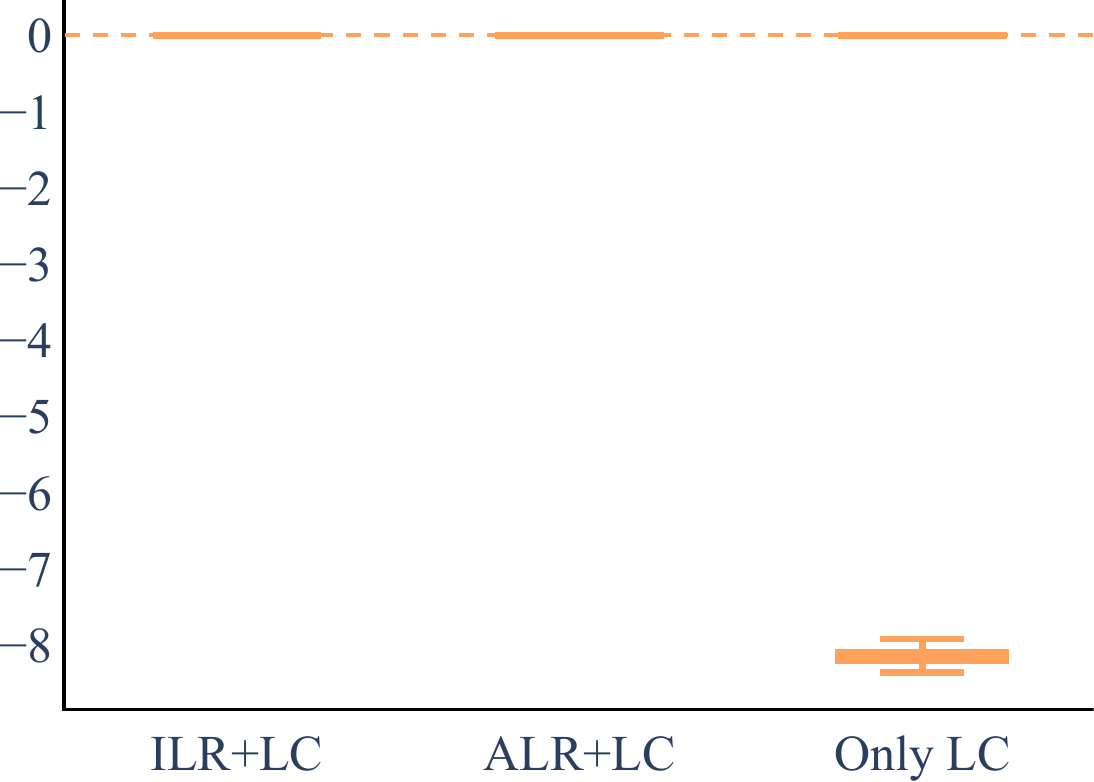}
  \caption{\textbf{Setting A with $p=250$, $q=10$: } The boxplots show the $\hat{\beta}$ values for the $50$ runs for each of the $8$ non-zero $\beta$ coefficients (dashed lines, left) and the $242$ zero $\beta$ coefficients (dashed line, right). The two-stage methods are able to recover the causal effect on average, whereas the naive regression method is not able to recover the true support. \ilrilr{} does not produce sensible estimates due to the missing regularization and is omitted for better readability.}
\label{appfig:linear_p250_Beta}
\end{figure}

\subsection*{Setting B}
\label{supp:method_semisyntheticsetup}

In this part we will examine the methods for Setting B \cref{eq:semisyntheticdata}. Note that the first stage is misspecified for the two-stage approaches, whereas the second stage is wellspecified for all methods.

\subsubsection*{Setting B with $p=3, q=2$}

\begin{figure}
  \centering
  \includegraphics[width=.45\textwidth]{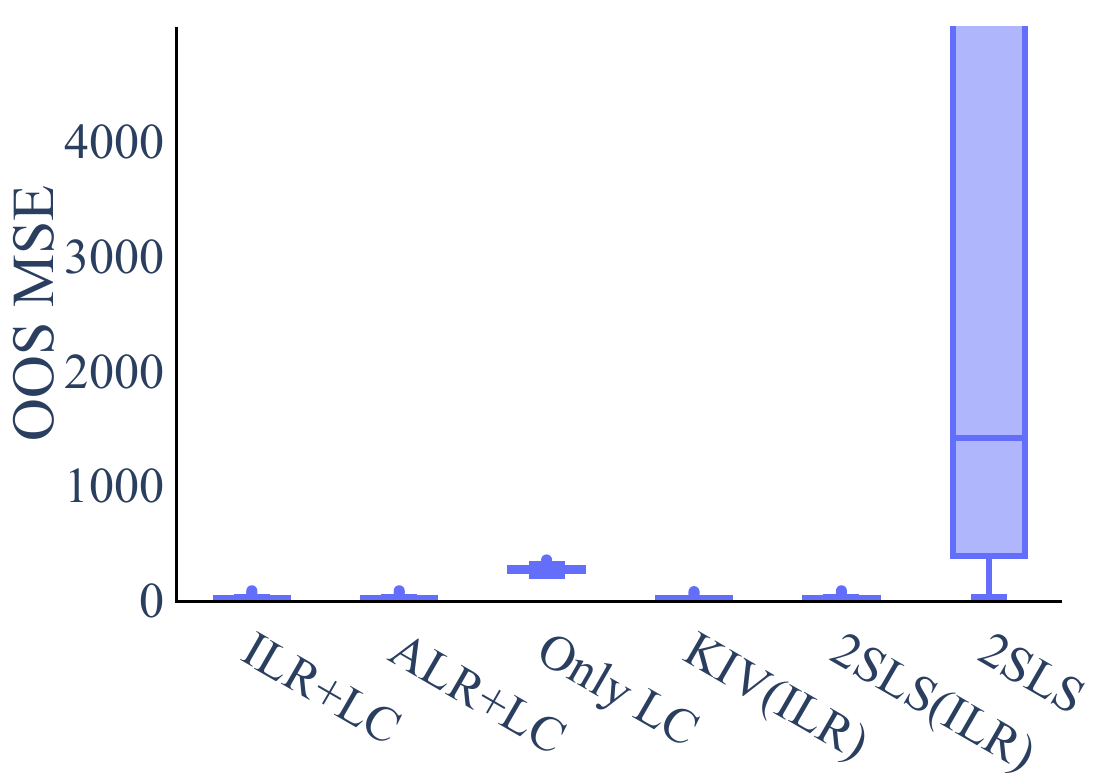}
  \hfill
  \includegraphics[width=.45\textwidth]{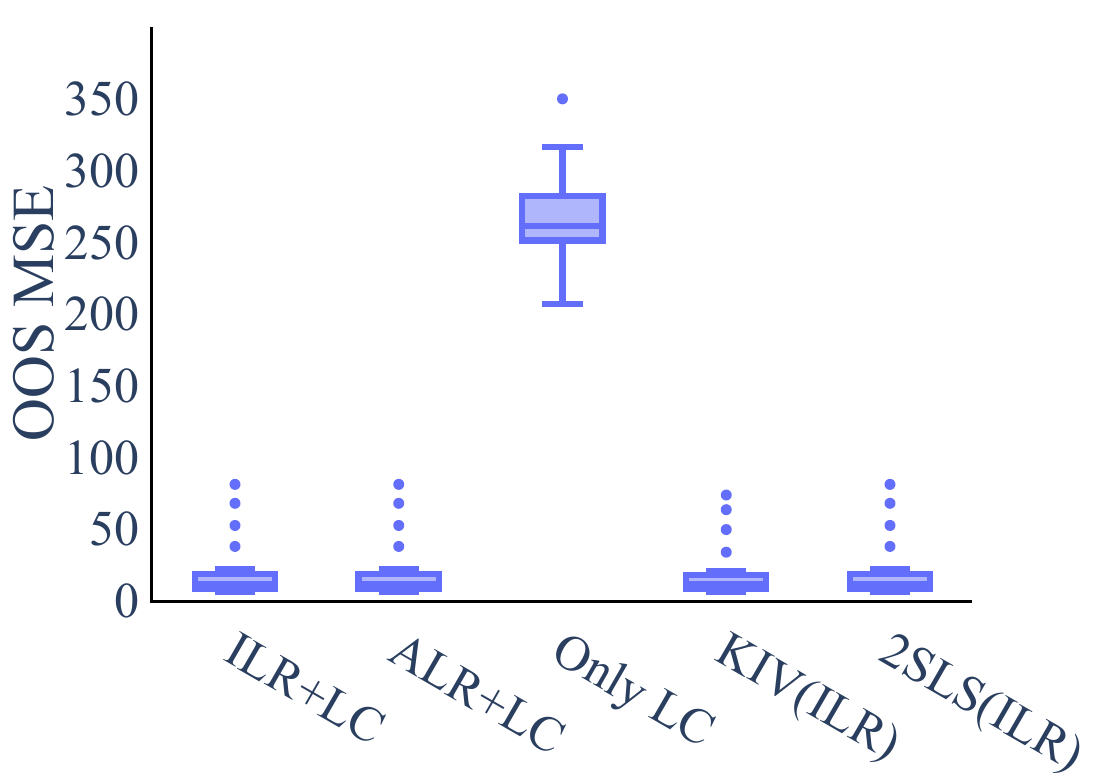}
  \caption{\textbf{Setting B with $p=3$, $q=2$:} The boxplots show the OOS MSE of $20$ runs. 2SLS and DIR+LC perform way worse as compared to the two-stage approaches (left). When we adjust the y-scale (right), Only LC also cannot compare to the remaining two-stage approaches. ALR+LC, ILR+LC, \ilrilr{} are equivalent in the low-dimensional case.}
\label{appfig:negbinom_p3_MSE}
\end{figure}

\begin{figure}
  \centering
  \includegraphics[width=.45\textwidth]{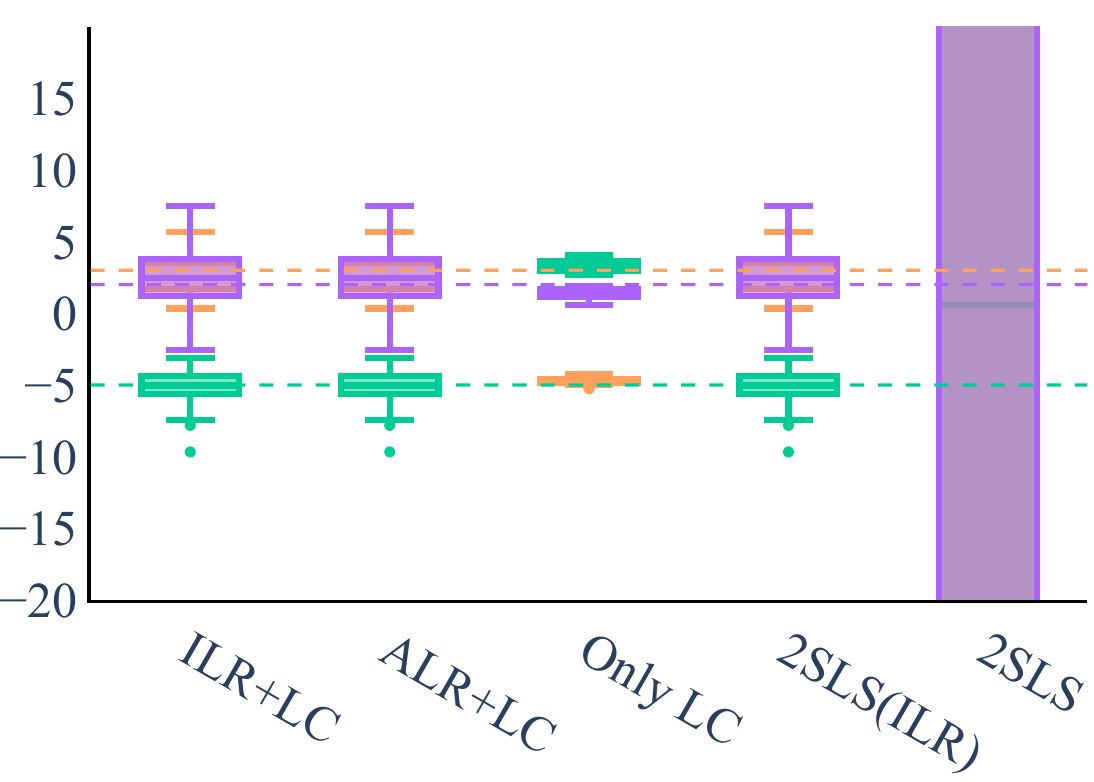}
  \hfill
  \includegraphics[width=.45\textwidth]{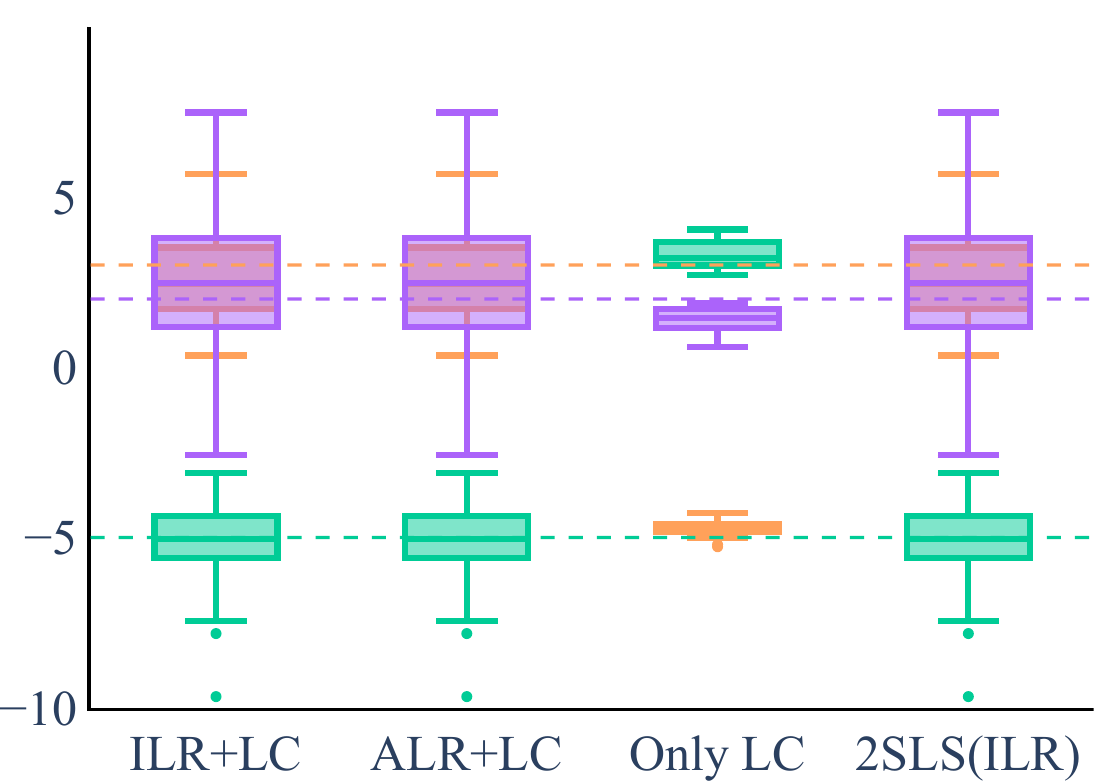}
  \caption{\textbf{Setting B with $p=3$, $q=2$: } The boxplots show the $\hat{\beta}$ values for the $20$ runs for each of the $3$ $\beta$ coefficients. The two-stage methods (except for DIR+LC and 2SLS) are able to recover the true causal $\beta$s on average. However, when we adjust the y-scale (right), the problem of confounding becomes apparent: Only LC flips the sign of two of the non-zero $\beta$ values.}
\label{appfig:negbinom_p3_Beta}
\end{figure}

Even in this low-dimensional scenario, DIR+LC suffers substantially from the misspecified second stage. It is not able to produce sensible estimates. We argue that this might be due to the ``forbidden regression'' issue. 
Furthermore, the naive regression is highly influenced by confounding. It even flips the estimated effect of two components, see \cref{appfig:negbinom_p3_Beta}.
Nevertheless the remaining two-stage methods, except 2SLS which ignores compositionality, perform reasonably well in recovering the true causal effect (see \cref{appfig:negbinom_p3_MSE}). 

\subsubsection*{Setting B with $p=30, q=10$}

Microbiome compositional data is typically high-dimensional and comprises many zero values. Moreover, it is often assumed that only a few microbial compositions (and hence $\beta$ parameters) influence an outcome of interest $Y$. 
Thus, in the following, we will emulate such a scenario and assume a sparse $\beta$ as ground truth and additionally---as ZINegBinom can incorporate sparsity also on $X$---run the models on relatively sparse $X$ (see \cref{supp:data_generation}). 

Note that for higher-dimensional approaches, we omit results for DIR+LC due to computational issues stemming from the maximum likelihood estimation of the $\alpha_0$ and $\alpha$ parameters in the first stage. 

\begin{figure}
  \centering
  \includegraphics[width=.45\textwidth]{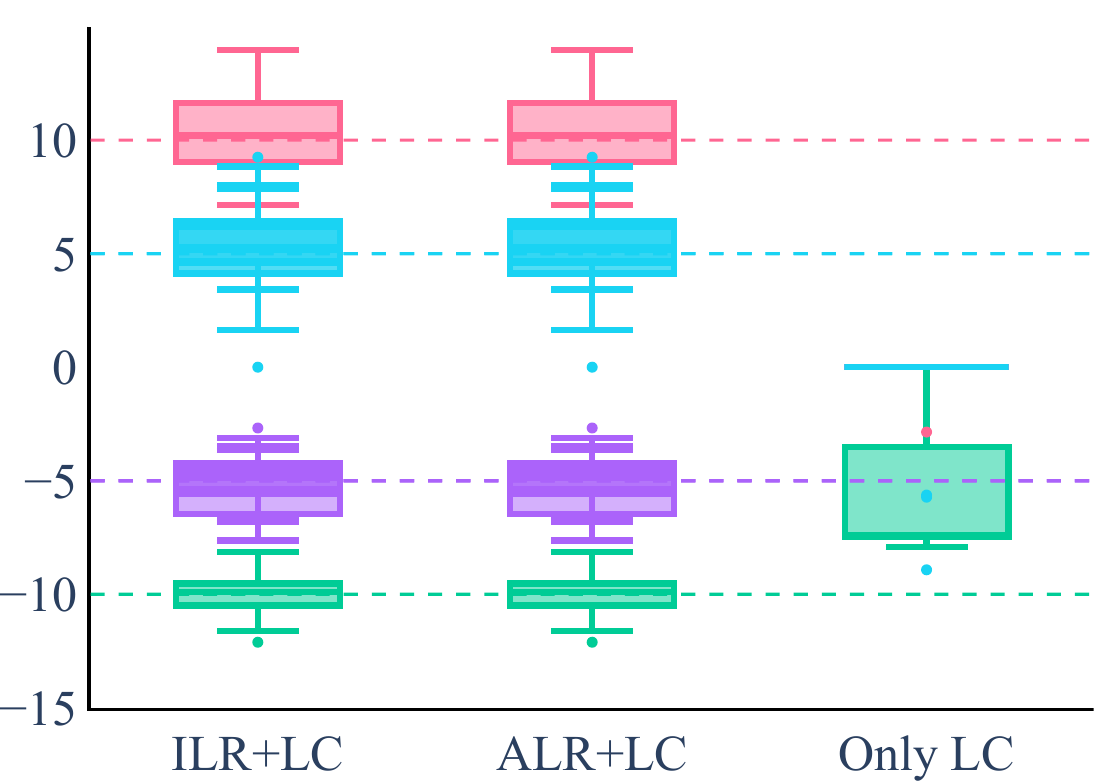}
  \hfill
  \includegraphics[width=.45\textwidth]{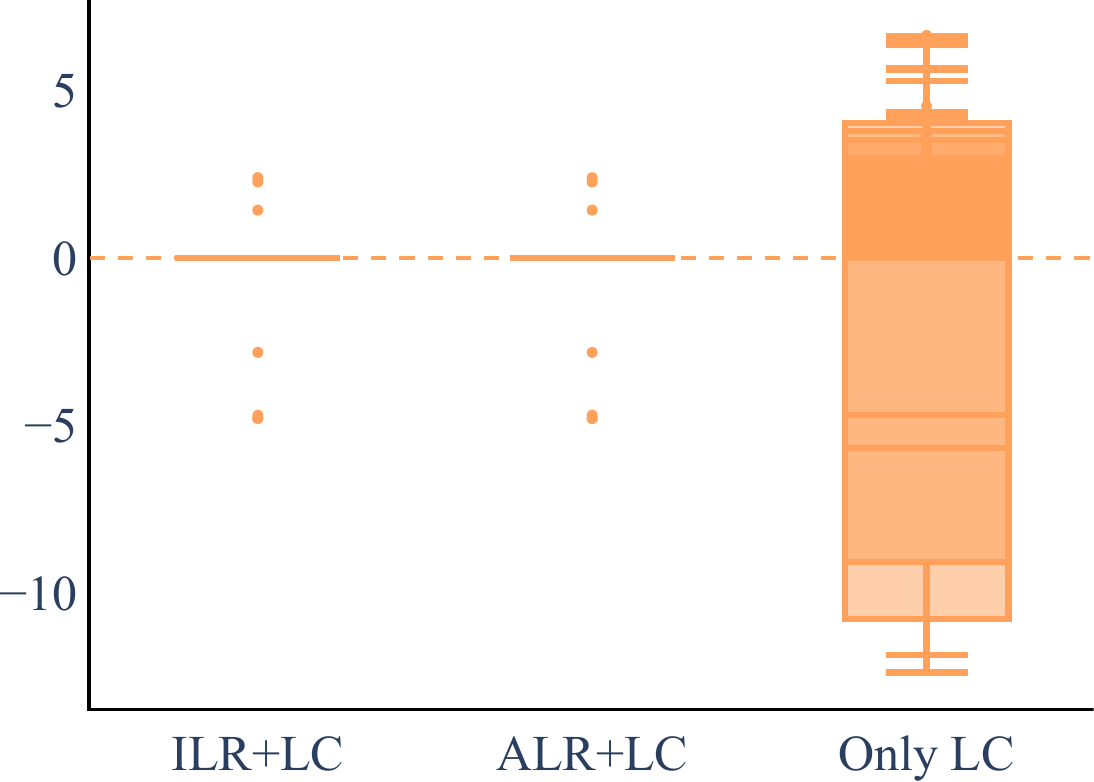}
  \caption{\textbf{Setting B with $p=30$, $q=10$: } The boxplots show the $\hat{\beta}$ values for the $20$ runs for each of the $8$ non-zero $\beta$ coefficients (dashed lines, left) and the $22$ zero $\beta$ coefficients (dashed line, right). The naive regression Only LC is not able to recover the $\beta$ values at all. \ilrlc{} and ALR+LC are better suited to recover the causal parameters when confounding is present. Despite the misspecified first stage, they are able to recover the causal $\beta$ values on average.}
\label{appfig:negbinom_p30_Beta}
\end{figure}

Moreover, for $p=30$, \ilrilr{} already is unfit to capture the causal effect due to missing regularization. Due to its high OOS MSE value, we omitted \ilrilr{} in \cref{appfig:negbinom_p30_Beta} for better readability. 
2SLS, which ignores the compositionality of $X$ altogether, is able to converge, but does not produce reasonable estimates.

\begin{figure}
  \centering
  \includegraphics[width=.45\textwidth]{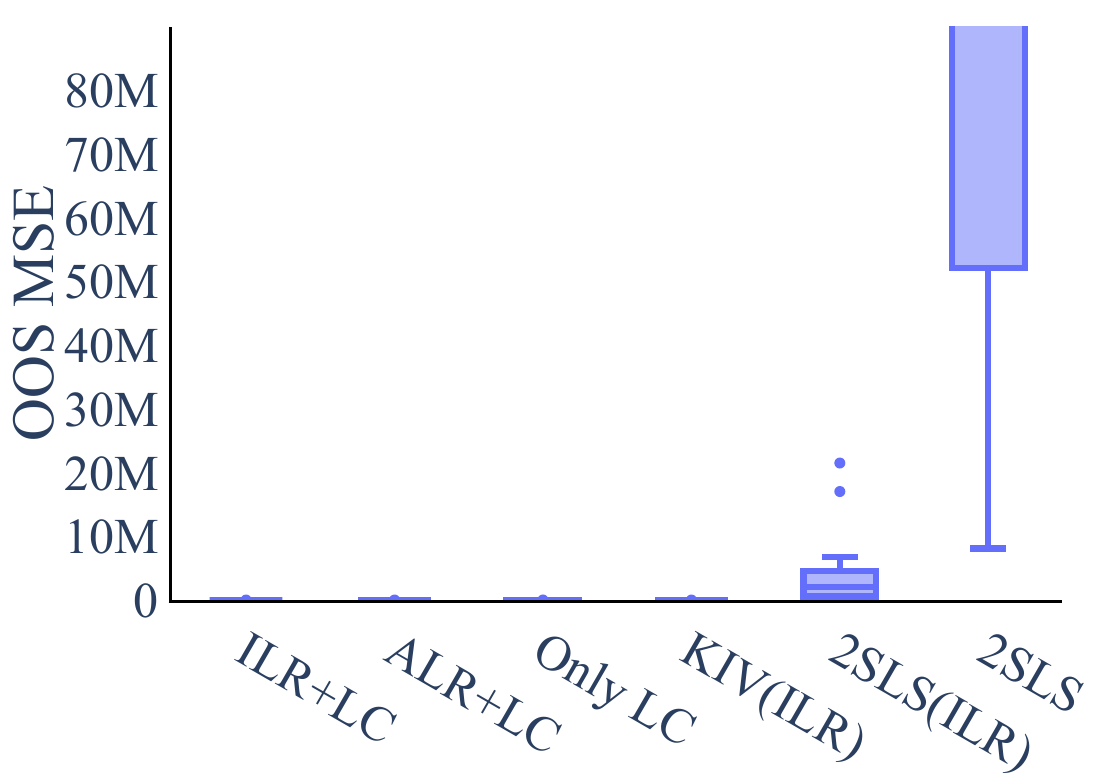}
  \hfill
  \includegraphics[width=.45\textwidth]{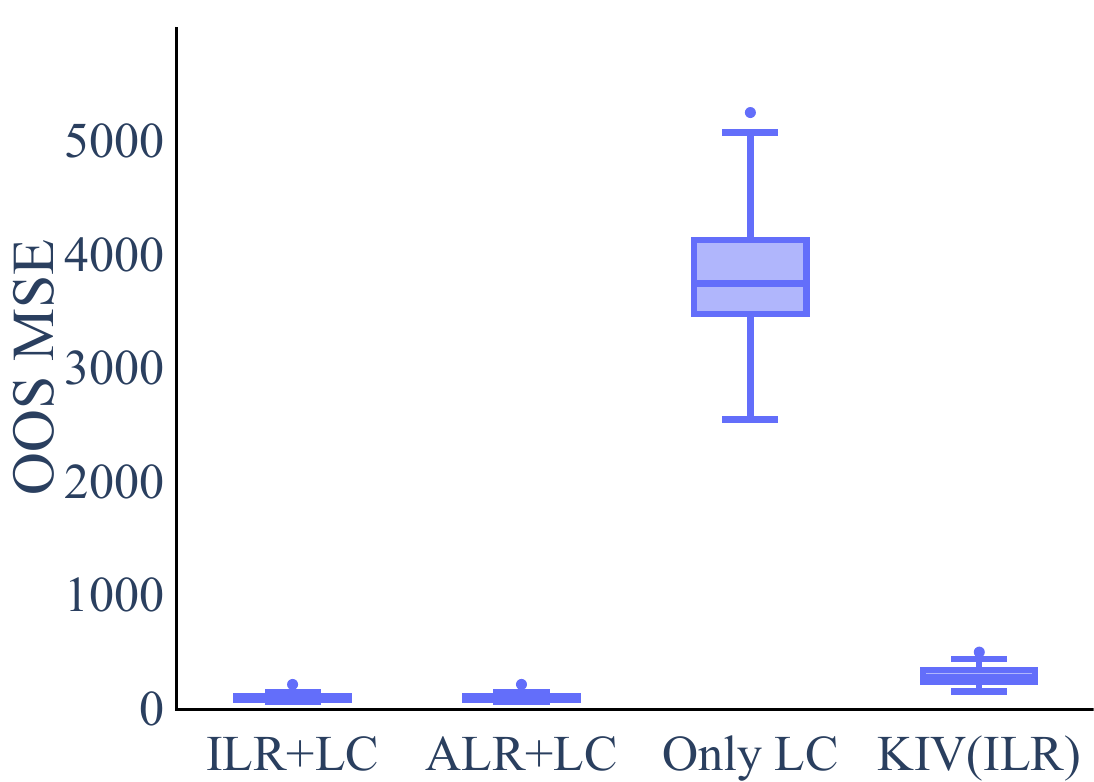}
  \caption{\textbf{Setting B with $p=30$, $q=10$: } The boxplots show the OOS MSE of $20$ runs. \ilrilr{} is only reasonable in low-dimensions (left). When we adjust the y-scale (right), we see that the remaining two-stage approaches outperform the naive regression.}
\label{appfig:negbinom_p30_MSE}
\end{figure}

For \kivilr{}, the difficulty of tuning the method in higher dimensions remains an issue (see \cref{appfig:negbinom_p30_MSE}). 
The remaining two-stage approaches, however, benefit substantially from the instrumentation of $X$ by $Z$. 
They outperform the naive regression both on OOS MSE (see \cref{appfig:negbinom_p30_MSE}), as well as on the recovery of the true $\beta$ values (see \cref{appfig:negbinom_p30_Beta}). 
While the naive regression not only fails to recover the true $\beta$ values, it also produces quite volatile estimates (see \cref{appfig:negbinom_p30_Beta}).

\subsubsection*{Setting B with $p=250, q=10$}

Microbiome compositional data is typically high-dimensional and comprises many zero values. Moreover, it is often assumed that only a few microbial compositions (and hence $\beta$ parameters) influence an outcome of interest $Y$. 
Thus, in the following, we will emulate such a scenario and assume a sparse $\beta$ as ground truth and additionally---as ZINegBinom can incorporate sparsity also on $X$---run the models on relatively sparse $X$ (see \cref{supp:data_generation}). 

Note that for higher-dimensional approaches, we omit results for DIR+LC due to computational issues stemming from the maximum likelihood estimation of the $\alpha_0$ and $\alpha$ parameters in the first stage. 

Both high-dimensional scenarios generally agree in their outcomes; for $p=250$ the shortcomings of the different approaches only get more enhanced.

\begin{figure}
  \centering
  \includegraphics[width=.45\textwidth]{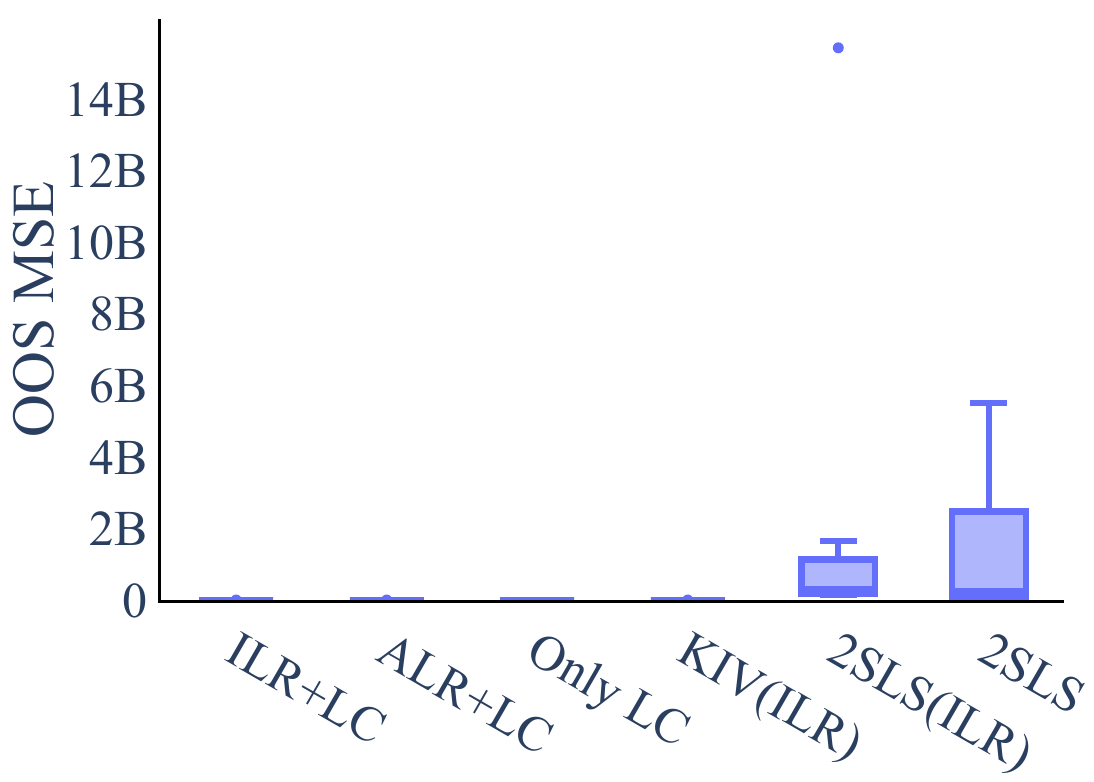}
  \hfill
  \includegraphics[width=.45\textwidth]{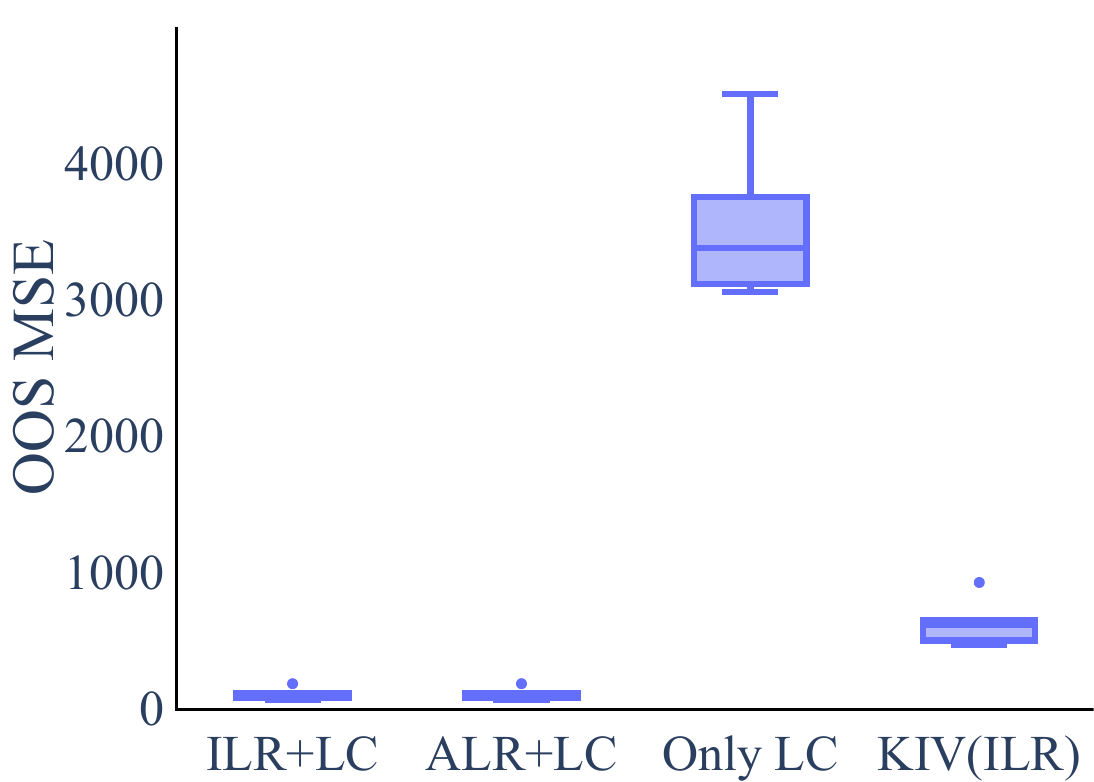}
  \caption{\textbf{Setting B with $p=250$, $q=10$: }The boxplots show the OOS MSE of $20$ runs. \ilrilr{} is only reasonable in low-dimensions and \kivilr{} is difficult to tune in higher dimensions (left). When we adjust the y-scale (right), we see that \ilrlc{} and ALR+LC outperform the naive regression.}
\label{appfig:negbinom_p250_MSE}
\end{figure}

\begin{figure}
  \centering
  \includegraphics[width=.45\textwidth]{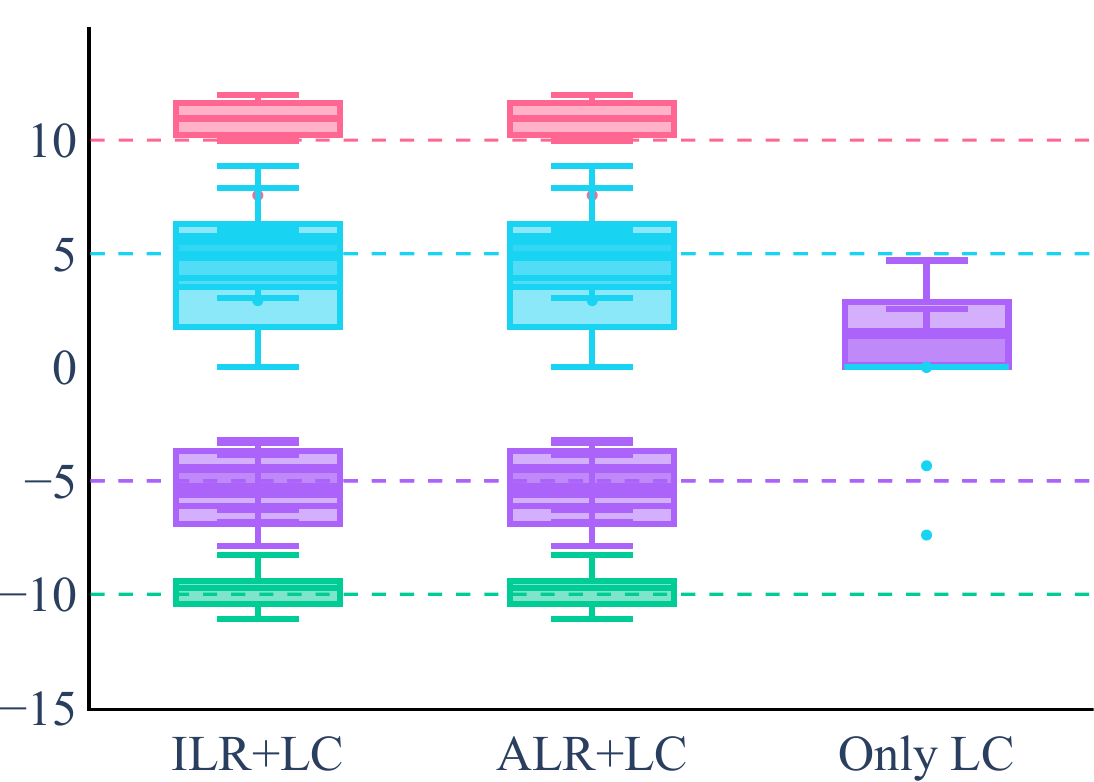}
  \hfill
  \includegraphics[width=.45\textwidth]{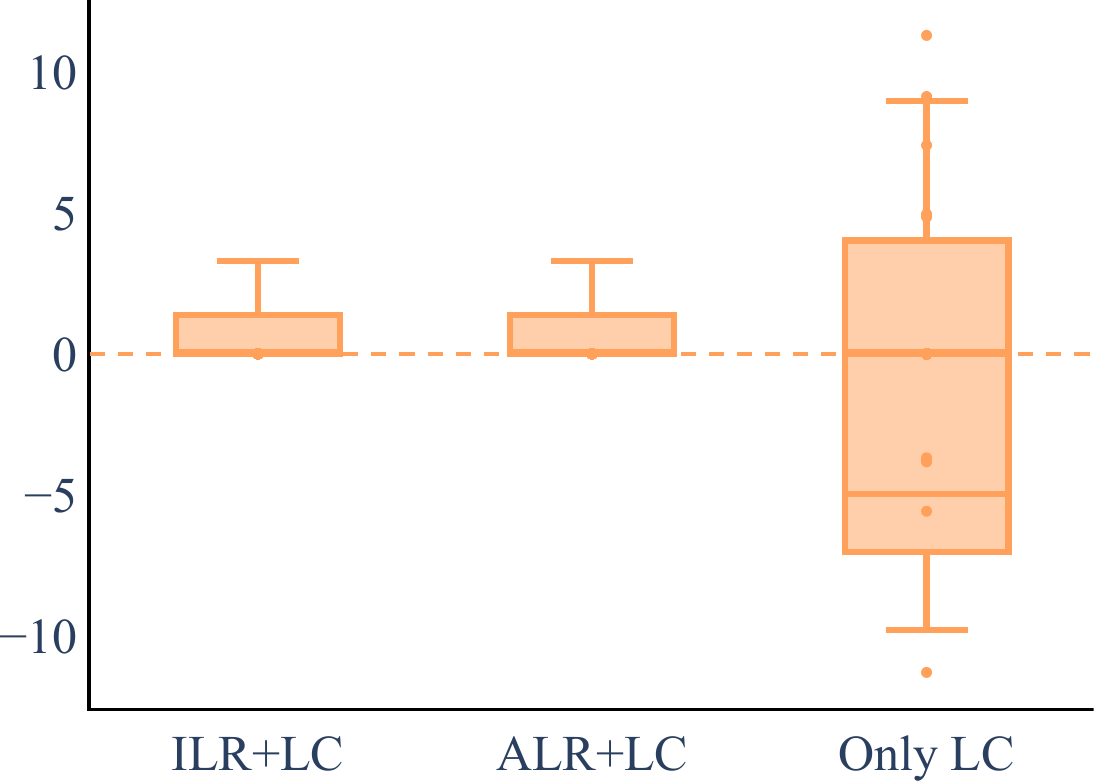}
  \caption{\textbf{Setting B with $p=250$, $q=10$: } The boxplots show the $\hat{\beta}$ values for the $20$ runs for each of the $8$ non-zero $\beta$ coefficients (dashed lines, left) and the $242$ zero $\beta$ coefficients (dashed line, right). Only LC is not able to recover the $\beta$ values at all. \ilrlc{} and ALR+LC are better suited to recover the causal parameters even when confounding is present. Despite the misspecified first stage, they are able to recover the causal $\beta$ values on average.}
\label{appfig:negbinom_p250_Beta}
\end{figure}

While 2SLS, which ignores the compositionality of $X$ altogether, is also able to converge for $p=250$, it does not produce reasonable estimates.
Further, the regularized two-stage methods still perform reasonably well, while \ilrilr{} and \kivilr{} cannot match that performance (see \cref{appfig:negbinom_p250_MSE}) due to the lack of sensible regularization. The naive approach can capture neither the causal effect nor the causal $\beta$ values (see \cref{appfig:negbinom_p250_MSE,appfig:negbinom_p250_Beta}). 

\subsection*{Further Settings for Robustness Estimation}

We will analyze the results form our ``robustness'' scenarios including a weak instrument setting and a setting with a nonlinear functional relationship in the second stage.

\subsubsection*{Weak Instrument}

\paragraph{Setting A with $p=3, q=2$ and weak instruments}
In a strong/valid instrument setting, two-stage methods have a clear advantage. To test the limitations of our methods, we now analyze an equivalent setting with a comparatively weak instrument.
In this setting, confounding is still noticeable (see \cref{appfig:linear_p3_xy_weak})
but the first stage F-statistic is much lower, indicating that we may suffer from weak instrument bias. 

\begin{figure}
  \centering
  \includegraphics[width=.45\textwidth]{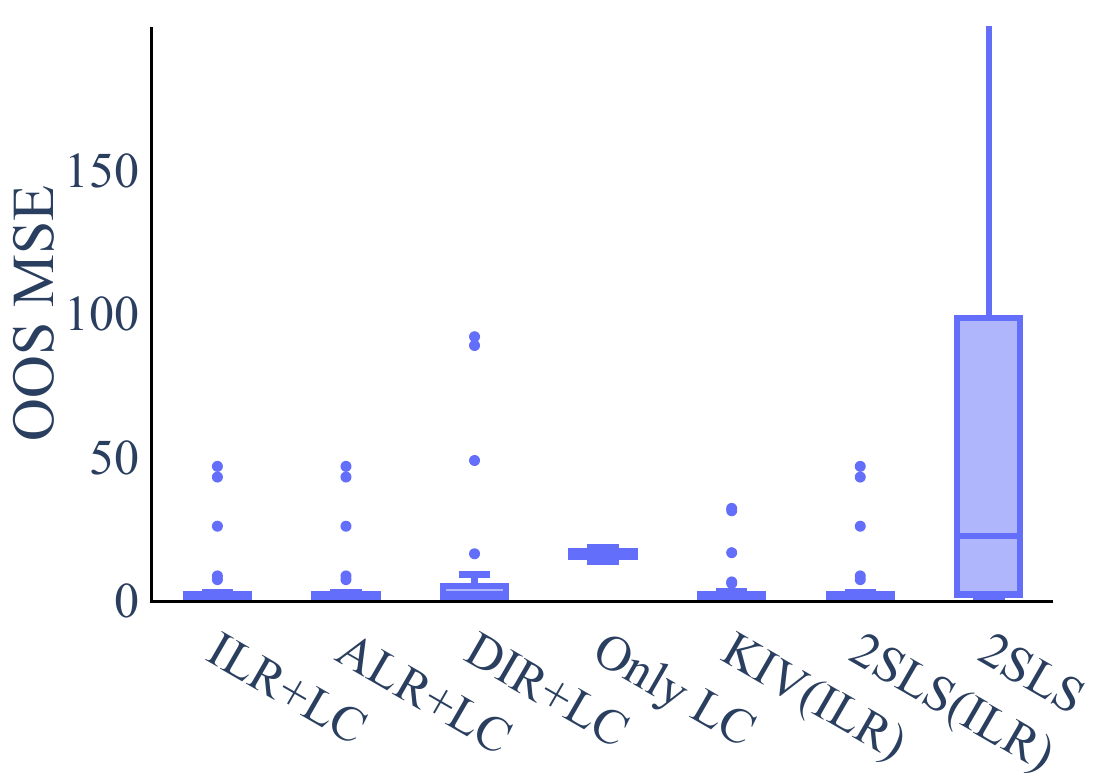}
  \hfill
  \includegraphics[width=.45\textwidth]{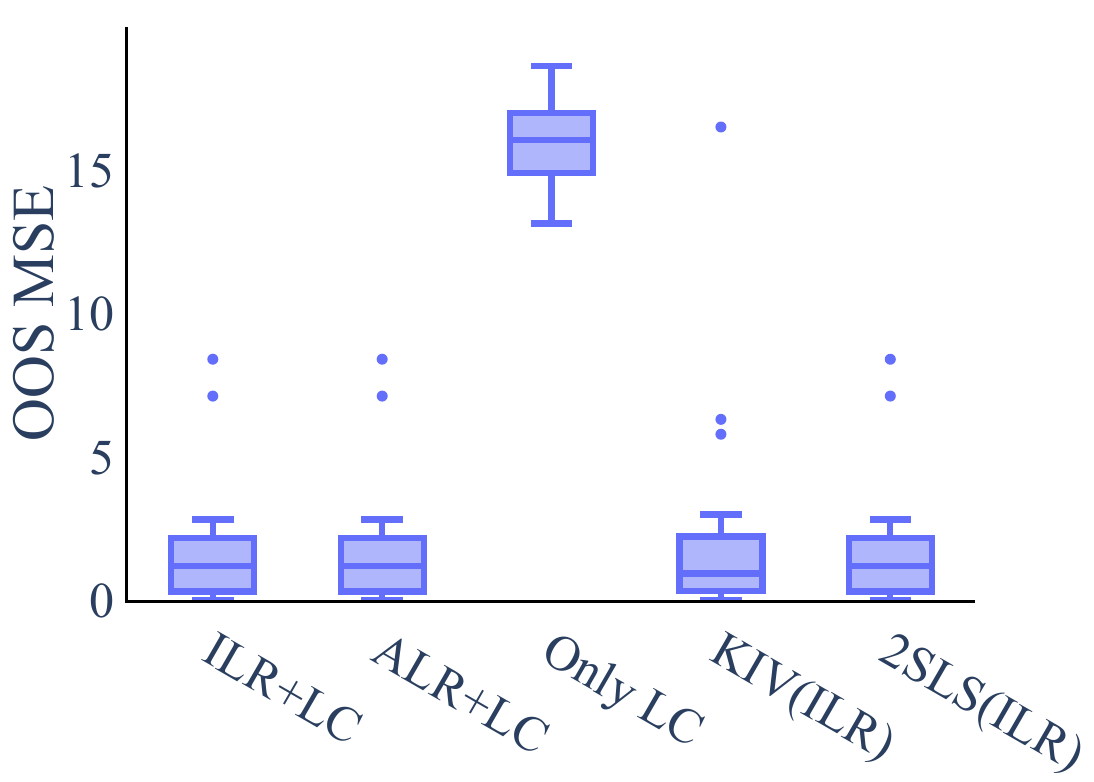}
  \caption{\textbf{Setting A with $p=3$, $q=2$ and weak instruments: } The boxplots show the OOS MSE of $50$ runs. The DIR+LC performs considerably worse than all the other methods. The problem might stem from the ``forbidden regression'' issue coming from two misspecified stages. On the right hand side we adjusted the y-scale. The graph shows that the other wellspecified methods still outperform the naive regression in terms of OOS MSE in the weak instrument setting. We observe a higher variance in performance than in the stronger instrument setting.}
\label{appfig:linear_p3_MSE_weak}
\end{figure}

\begin{figure}
  \centering
  \includegraphics[width=.45\textwidth]{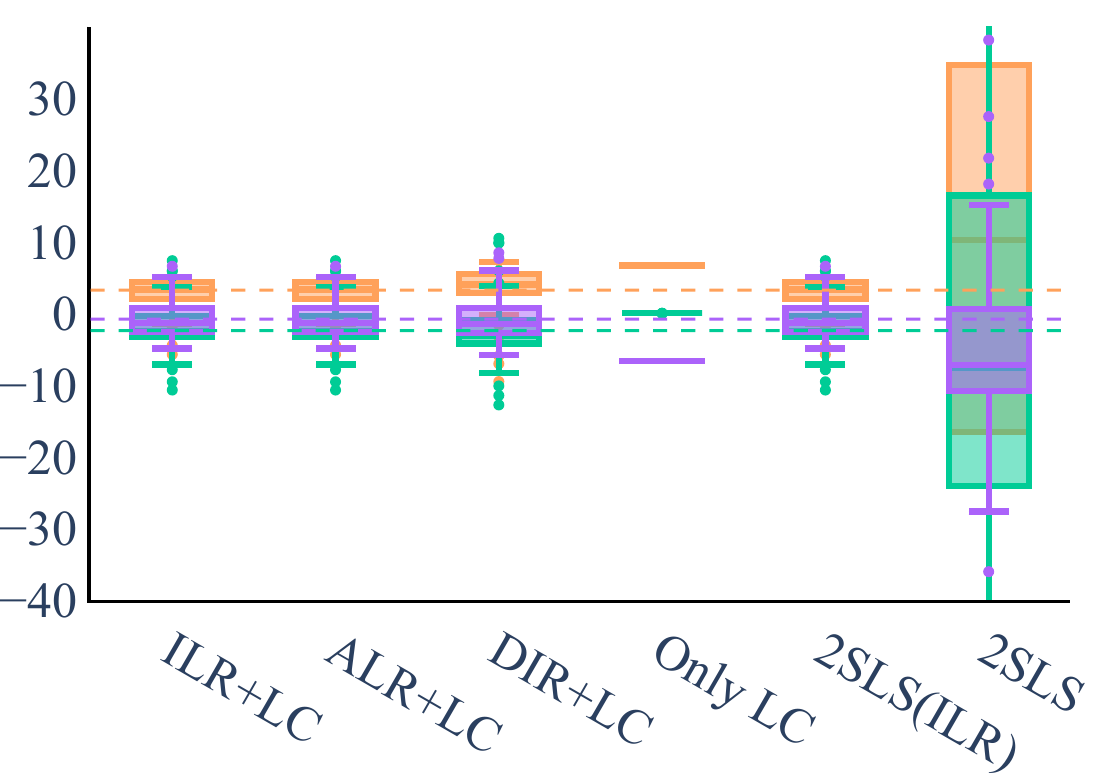}
  \hfill
  \includegraphics[width=.45\textwidth]{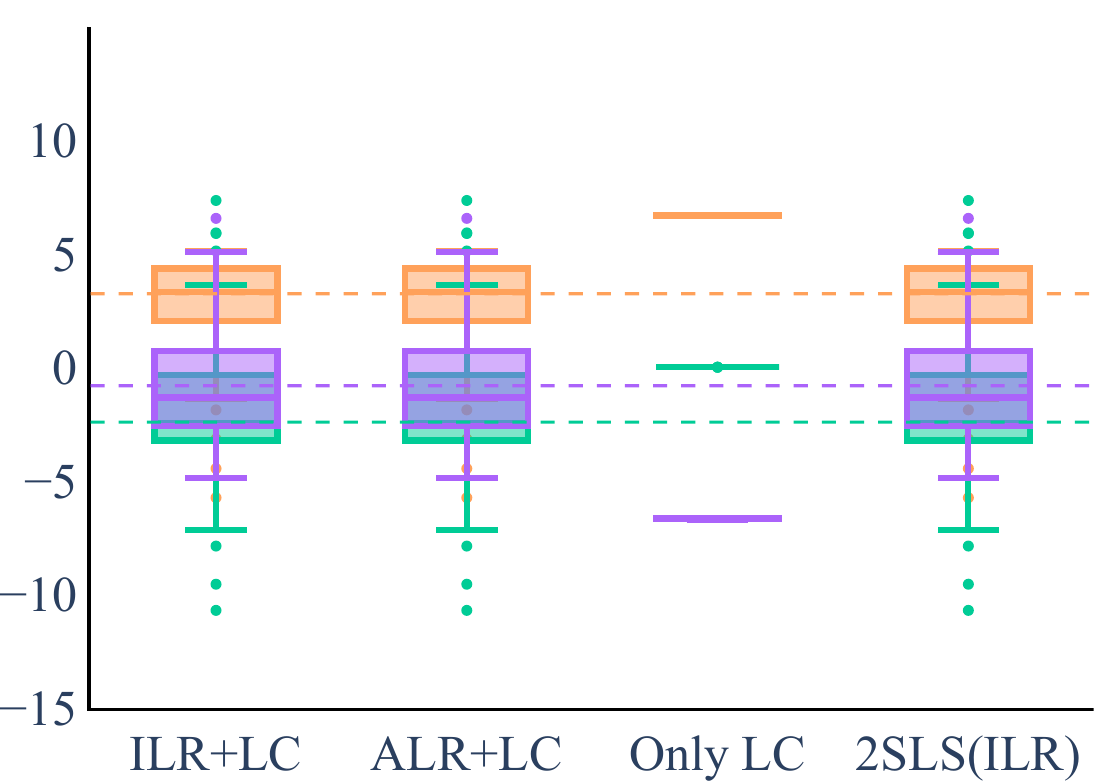}
  \caption{\textbf{Setting A with $p=3$, $q=2$ and weak instruments: } The boxplots show the $\hat{\beta}$ values for the $50$ runs for each of the $3$ $\beta$ coefficients (dashed lines). In the weaker instrument setting, we are able to recover the true $\beta$ values for the wellspecified two-stage methods, even though the variance of the $\hat{\beta}$ is higher while the naive regression is still subject to confounding (right, zoomed in plot). Only the DIR+LC is not able to recover the causal effect and seems biased toward the naive regression (left). This might be due to two misspecified stages.}
\label{appfig:linear_p3_Beta_weak}
\end{figure}

The two-stage methods have a higher variation in their estimates, both for OOS MSE and $\hat{\beta}$ (see \cref{appfig:linear_p3_MSE_weak,appfig:linear_p3_Beta_weak}), whereas the naive regression does not change at all (since only the first stage data generation has changed).
Nevertheless, the wellspecified two-stage methods (ALR+LC, ILR+LC, \ilrilr{}) still recover the causal effects better than the naive regression.
Only the DIR+LC regression runs into problems due to two misspecified stages.
We thus conclude that the ``forbidden regression'' is not necessarily detrimental to cause-effect estimation when the instrument is strong, but can indeed result in unreliable results for weaker instruments.

\subsubsection*{Non-linear Second Stage}

\paragraph{Setting A with $p=3, q=2$ and a non-linear function form of $f$}

The two-stage methods perform well if they are in a wellspecified setting. With the DIR+LC method, however, it becomes obvious that misspecification can become problematic. Furthermore, wellspecification is typically impossible to ascertain in practice and most real-world examples are likely not perfectly linear. 
Thus, we add a polynomial $X$ dependency term in the second stage to evaluate ALR+LC, ILR+LC and \ilrilr{} on a partly misspecified setting.

\begin{figure}
  \centering
  \includegraphics[width=.45\textwidth]{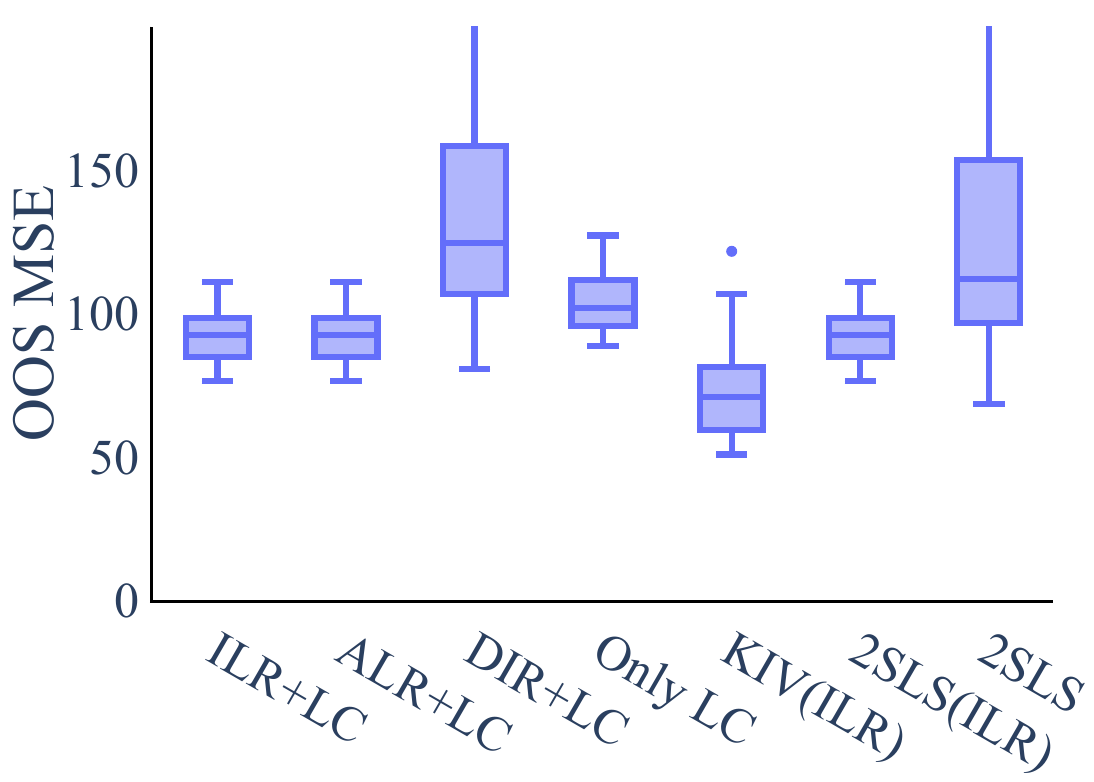}
  \hfill
  \includegraphics[width=.45\textwidth]{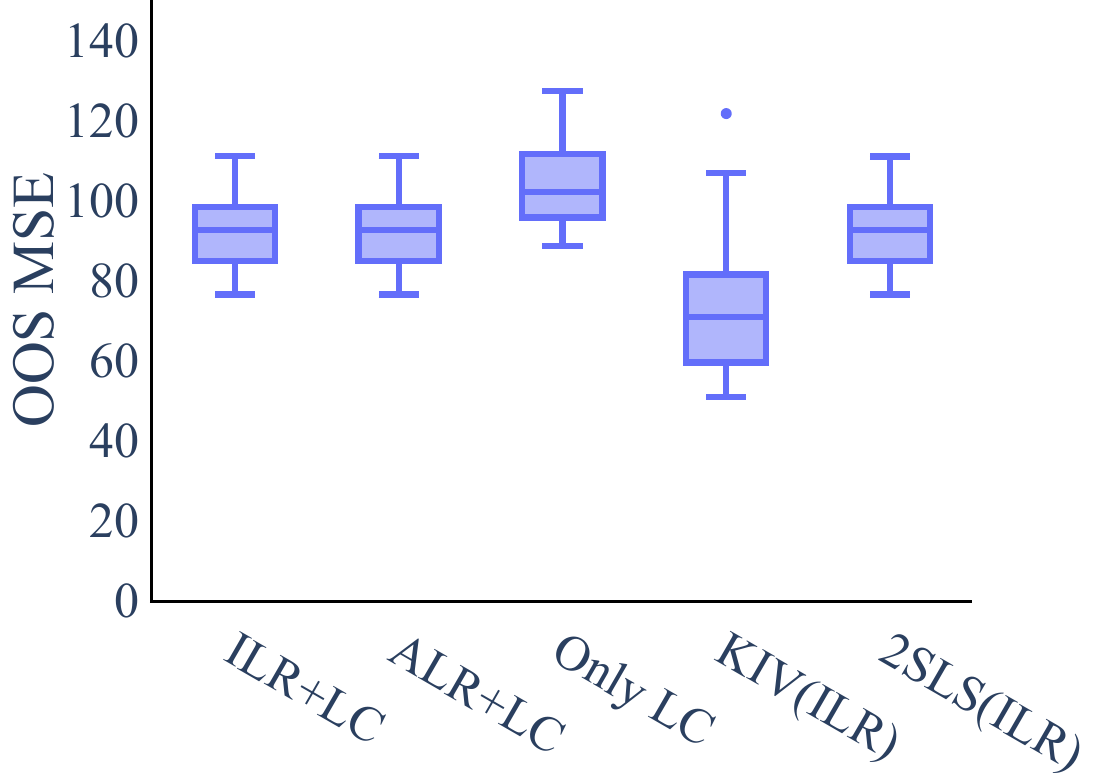}
  \caption{\textbf{Setting A with $p=3$, $q=2$ and a non-linear function form of $f$: } The boxplots show the OOS MSE of $50$ runs. DIR+LC and 2SLS perform worst (left). The other two-stage approaches are able to recover the causal effect better than the naive regression. The correctly specified first stage helps in filtering out the confounding effect in the two-stage methods.}
\label{appfig:nonlinear_p3_MSE}
\end{figure}

Note that we can only look at the OOS MSE as the $\beta$ values do not carry any causal interpretation (see \cref{appfig:nonlinear_p3_MSE}).
The DIR+LC still suffers from two misspecified stages and performs worst. 
When only the second stage is misspecified, ALR+LC, ILR+LC and \ilrilr{} still outperform the naive regressions. However, we are not able to capture the true causal effect because of the misspecification in the second stage. The overall error thus grows in all methods.

\subsubsection*{Scarce Data Example $p \gg n$}

\paragraph{Setting A with $p=250, q=10$ and $n=100$}

\begin{figure}
  \centering
  \includegraphics[width=.45\textwidth]{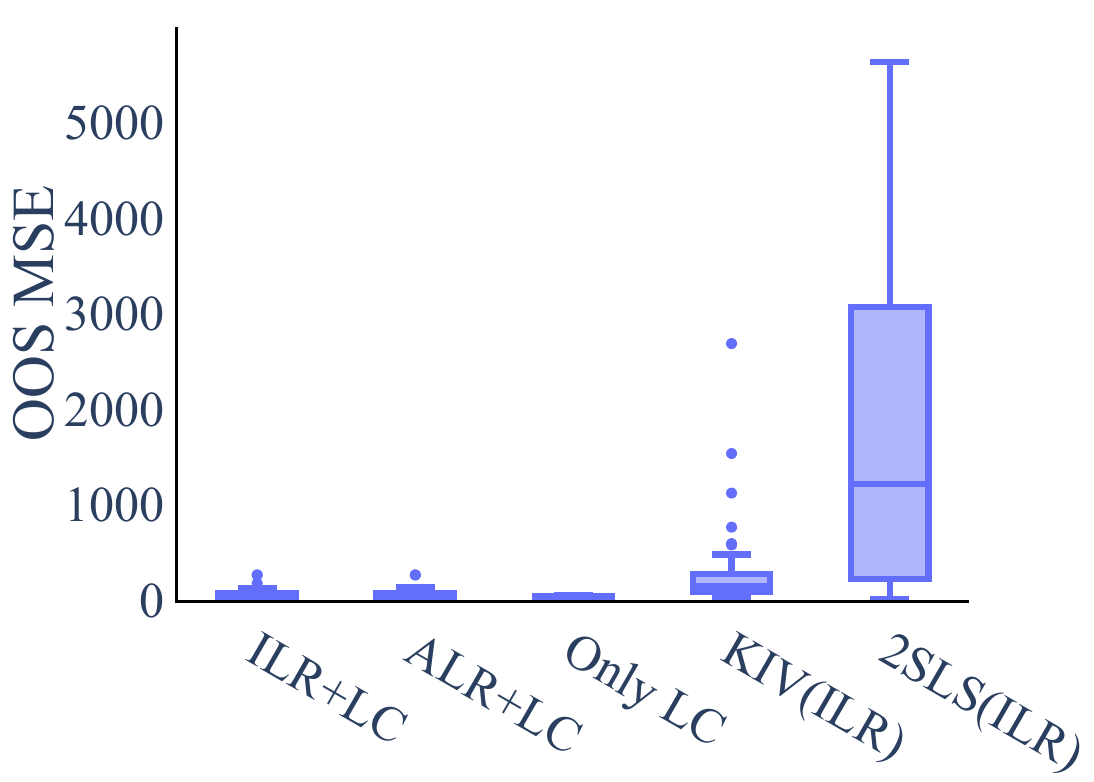}
  \hfill
  \includegraphics[width=.45\textwidth]{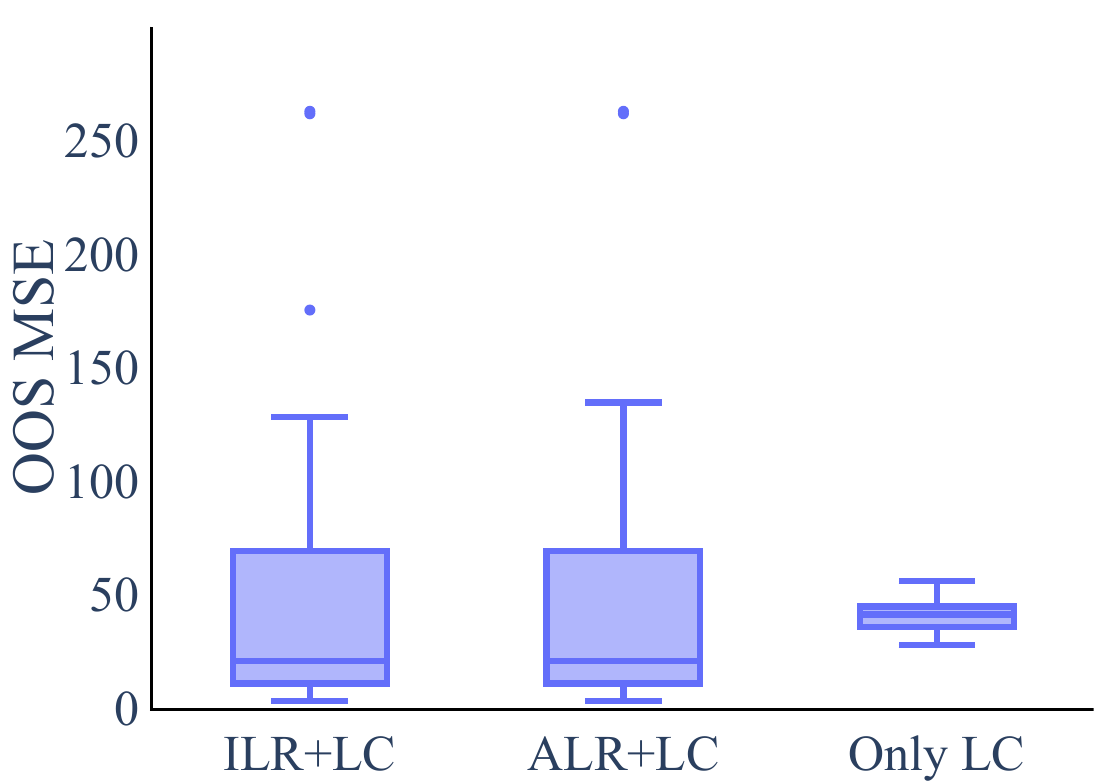}
  \caption{\textbf{Setting A with $n=100$, $p=250$, $q=10$: } The boxplots show the OOS MSE of $50$ runs. \ilrilr{} and \kivilr{} are volatile and lack sensible regularization (left). This problem is more pressing as the dimensionality grows and the number of data samples decreases. When we adjust the y-scale, we see that also Only LC (right) performs worse compared to the regularized two-stage approaches. Note however, that the width of the confidence intervals of the two-stage approaches has increased compared to the larger dataset.}
\label{appfig:linear_p250_MSE_n100}
\end{figure}

Also for $n=100$, the same problem of the lack of regularization persist. \ilrilr{} and \kivilr{})do not perform well at all (\cref{appfig:linear_p250_MSE_n100}).
For readability we thus omitted \ilrilr{} from the $\beta$ plots. 

\begin{figure}
  \centering
  \includegraphics[width=.45\textwidth]{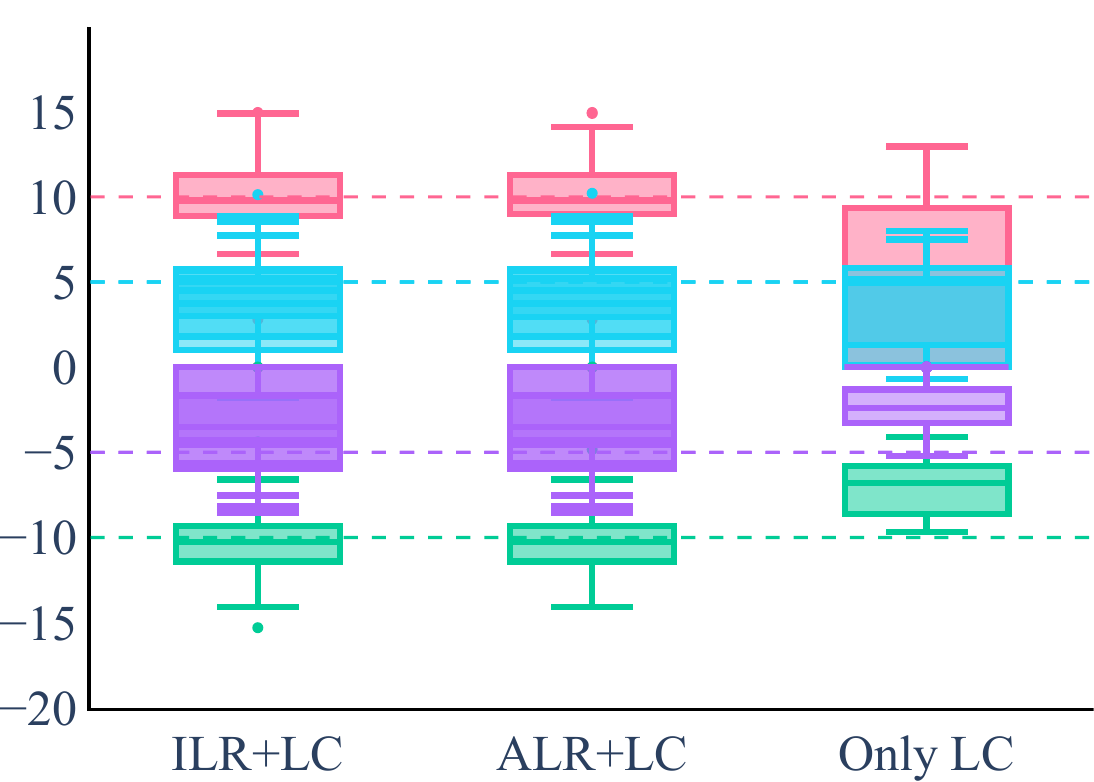}
  \hfill
  \includegraphics[width=.45\textwidth]{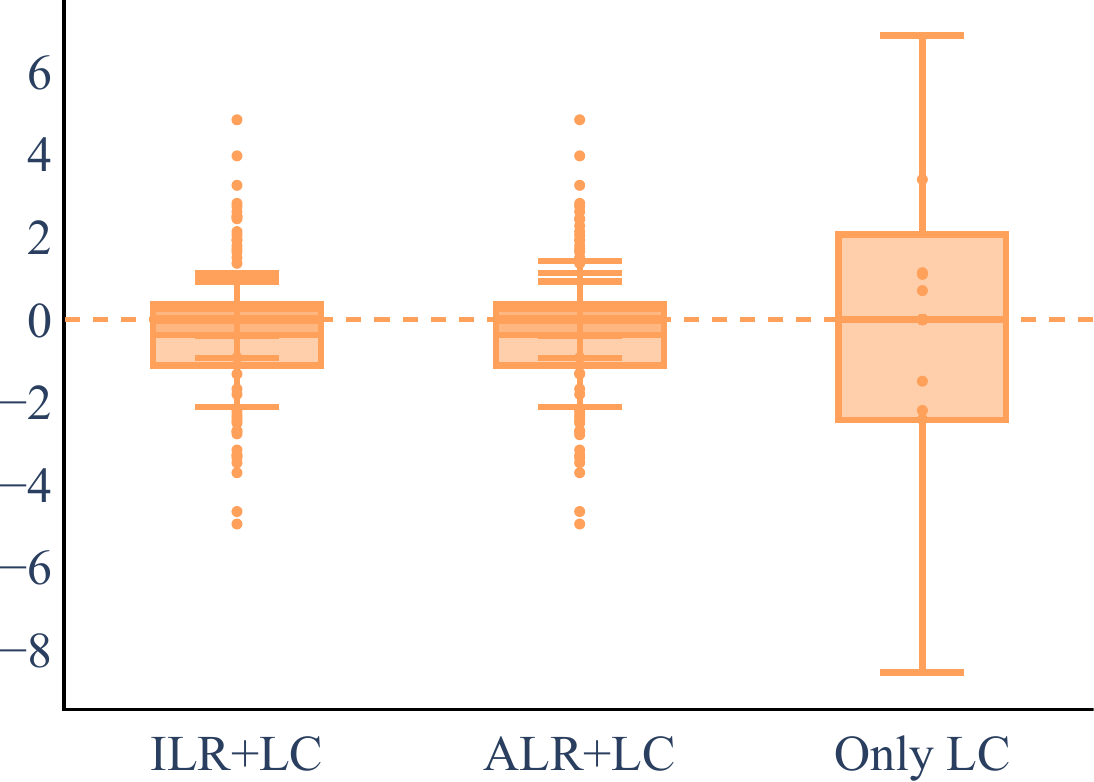}
  \caption{\textbf{Setting A with $n=100$, $p=250$, $q=10$: } The boxplots show the $\hat{\beta}$ values for the $50$ runs for each of the $8$ non-zero $\beta$ coefficients (dashed lines, left) and the $242$ zero $\beta$ coefficients (dashed line, right). The two-stage methods are able to recover the causal effect on average, whereas the naive regression method is not able to recover the true support. \ilrilr{} does not produce sensible estimates due to the missing regularization and is omitted for better readability. Note however, that the width of the confidence intervals of the two-stage approaches has increased compared to the larger dataset.}
\label{appfig:linear_p250_Beta_n100}
\end{figure}

As against the large dataset example, the two-stage methods naturally show much larger confidence interval around their estimates, whereas the naive regression does not suffer at the same scale. 
However the naive regression has troubles to recover the full support (see \cref{appfig:linear_p250_Beta_n100}).
Thus, even with a much larger uncertainty, the regularized two-stage approaches are able to recover the true causal $\beta$s. 

\end{document}